\documentclass[letterpaper]{article} % DO NOT CHANGE THIS
\usepackage{aaai23}  % DO NOT CHANGE THIS
\usepackage{times}  % DO NOT CHANGE THIS
\usepackage{helvet}  % DO NOT CHANGE THIS
\usepackage{courier}  % DO NOT CHANGE THIS
\usepackage[hyphens]{url}  % DO NOT CHANGE THIS
\usepackage{graphicx} % DO NOT CHANGE THIS
\usepackage{natbib}  % DO NOT CHANGE THIS AND DO NOT ADD ANY OPTIONS TO IT
\usepackage{caption} % DO NOT CHANGE THIS AND DO NOT ADD ANY OPTIONS TO IT
\usepackage{algorithm}
\floatname{algorithm}{Procedure}
\usepackage{algorithmic}

\usepackage{newfloat}
\usepackage{listings}
\DeclareCaptionStyle{ruled}{labelfont=normalfont,labelsep=colon,strut=off} % DO NOT CHANGE THIS
\floatstyle{ruled}
\newfloat{listing}{tb}{lst}{}
\floatname{listing}{Listing}
\title{On Error and Compression Rates for Prototype Rules}
\author {
    Omer Kerem,\textsuperscript{\rm 1}
    Roi Weiss\textsuperscript{\rm 2}
}
\affiliations {
    \textsuperscript{\rm 1} Ben-Gurion University of the Negev\\
    \textsuperscript{\rm 2} Ariel University\\
    omerkere at post dot bgu.ac.il, roiw at ariel dot ac.il
}

\usepackage{bibentry}
\usepackage{mathrsfs}
\usepackage{amssymb}
\usepackage{amsmath,amsthm}
\usepackage{amsfonts}
\usepackage{tikz}
\usepackage{bm}
\usepackage{dsfont}

\usepackage{makecell}

\usepackage{tikz}
\usepackage{forest}
\usetikzlibrary{intersections}
\usepackage{tikz}
\usetikzlibrary{arrows.meta}
\usepgflibrary{decorations.markings}
\usetikzlibrary{through}
\usetikzlibrary{decorations.pathmorphing}
\usetikzlibrary{backgrounds}
\usetikzlibrary{decorations.pathreplacing}

\usetikzlibrary{calc}

\newtheorem{theorem}{Theorem}

\newtheorem{lemma}[theorem]{Lemma}
\newtheorem{definition}{Definition}
\newtheorem{remark}{Remark}

\newtheorem{claim}[theorem]{Claim}

\newcommand{\ul}{m}
\def\N{\mathbb N}

\def\D{{\cal D}}
\def\Xul{{\bX'_\ul}}
\def\bXg{{\bX}}
\def\Xg{{X}}
\def\bYg{{\bY}}
\def\Yg{{Y}}
\def\Dul{\Xul}
\def\Xl{{\bX_n}}

\def\Dl{{\mathcal{D}_n}}

\def\R{{\mathbb R}}

\def\C{C}

\def\X{{\cal X}}
\def\bX{{\bf X}}
\def\bG{{\bf G}}

\def\PROB {{\mathbb P}}
\def\EXP {{\mathbb E}}
\def\Var{\mathbb{V}}

\def \p {^{\prime}}

\def \ve {\varepsilon}
\def\argmin{\mathop{\rm arg\, min}}
\def\argmax{\mathop{\rm arg\, max}}
\newcommand{\UB}[1]{{\rm UB(#1)}}
\newcommand{\supp}{\mathcal{S}}
\newcommand{\dlb}{\nu_0}
\newcommand{\ms}{L}

\newcommand{\Vor}{\mathcal V}
\newcommand{\V}{\Vor}

\newcommand{\nuc}{q}

\newcommand{\Y}{\mathcal Y}
\newcommand{\bY}{{\bf Y}}
\newcommand{\tXg}{\widetilde \Xg}

\newcommand{\tbXg}{\widetilde \bXg}

\newcommand{\setminust}{\!\setminus\!}

\newcommand{\tbX}{\widetilde \bX}

\newcommand{\tbY}{\widetilde \bY}

\newcommand{\NNm}{\mathcal N}

\newcommand{\tD}{\widetilde \D}
\newcommand{\Dg}{\D}

\newcommand{\tX}{\widetilde X}
\newcommand{\tY}{\widetilde Y}

\newcommand{\diam}{\mathrm{diam}}

\newcommand{\CN}{N}
\newcommand{\Seg}{\text{Seg}}
\newcommand{\boundary}[1]{{\mathcal B}_{#1}}
\newcommand{\clBound}[1]{\bar{\boundary{}}_{#1}}
\newcommand{\Bboundary}[1]{\partial\boundary{#1}}

\newcommand{\dtb}{\delta}
\newcommand{\reg}{\eta}

\newcommand{\gm}{\xi}
\newcommand{\cf}[1]{h_{#1}}
\newcommand{\tcf}[1]{\tilde{h}_{#1}}
\newcommand{\eqdef}{\overset{\text{def}}=}

\renewcommand{\C}{\textbf{C}}
\newcommand{\Cp}{\C_+}
\newcommand{\interior}{\text{interior}}
\newcommand{\Cl}{\text{Cl}}
\newcommand{\relint}{\text{relint}}
\newcommand{\bXn}{s}

\renewcommand{\vec}{\boldsymbol}

\newcommand{\optinet}{\textup{OptiNet}}
\newcommand{\protonn}{\textup{Proto-NN}}
\newcommand{\hybrid}{\textup{Proto-$k$-NN}}
\newcommand{\protoknnn}{\textup{Proto-$k_n$-NN}}
\newcommand{\compalg}{\textup{ProtoComp}}
\newcommand{\compheu}{\textup{ProtoCompApprox}}

\newcommand{\dist}{\nu}

\newcommand{\tm}{\widetilde m}
\newcommand{\goptinet}{g^{\optinet{}}}

\newcommand{\Xinn}[1]{X^{(#1)}}
\newcommand{\Yinn}[1]{Y^{(#1)}}
\newcommand{\Ball}[2]{B_{#2}(#1)}
\newcommand{\IND}[1]{\mathbb I_{\{#1\}}}
\newcommand{\surf}{\mathcal M}

\newcommand{\barBall}[2]{{\bar B}_{#2}(#1)}

\usepackage{todonotes}
\begin{document}

\maketitle

\begin{abstract}
We study the close interplay between error and compression in the non-parametric multiclass classification setting in terms of prototype learning rules. We focus in particular on a recently proposed compression-based learning rule termed \optinet{} \citep{kontorovichsabatourner16,KoSaWe17,HaKoSaWe20}. Beyond its computational merits, this rule has been recently shown to be universally consistent in any metric instance space that admits a universally consistent rule---the first learning algorithm known to enjoy this property. However, its error and compression rates have been left open. Here we derive such rates in the case where instances reside in Euclidean space under commonly posed smoothness and tail conditions on the data distribution. We first show that \optinet{} achieves non-trivial compression rates while enjoying near minimax-optimal error rates. We then proceed to study a novel general compression scheme for further compressing prototype rules that locally adapts to the noise level without sacrificing accuracy. Applying it to \optinet{}, we show that under a geometric margin condition, further gain in the compression rate is achieved. Experimental results comparing the performance of the various methods are presented.
\end{abstract}

\section{Introduction}
%\chapter{INTRODUCTION}
\label{chapter:intro}
%Following the pioneering works by von Luxburg and Bousquet \cite{von2004distance},  Chaudhuri and Dasgupta \citep{ChDa14}, Kpotufe \citep{kpo09} 
%%Kontorovitch and Gottlieb \citep{?} 
%on learning in metric spaces, in recent years we have been
%developing an alternative family of
%%AK
%%NN rules
%learning algorithms
%based on \emph{regularized
%%and compression-based 
%versions of the 1-NN rule}. As established in our 
%%early 
%works \cite{kontorovich2014bayes,kontorovich2014maximum,KoSaWe17,HaKoSaWe20,GyWe21} and
%%AK
%%that
%those
%of others \cite{GoKoNi14,kontorovichsabatourner16,gottlieb2016adaptive,xue2018achieving}, 
%%[Kontorovich \& Weiss, 2014] and [Kontorovich \& Weiss, 2015], 
%regularized 1-NN algorithms enjoy many statistical and computational advantages over the traditional $k$-NN rule. Salient among those are 
%%tight data-dependent generalization bounds and 
%near-optimal sample compression guarantees implying considerable runtime and memory savings \cite{GoKoNi14}.
%
%

%Popularized by Occam's razor, 
%Popularized by Occam's razor rule of thumb 
%\citep{ariew1976ockham},
%As popularized by Occam's razor rule of thumb \citep{ariew1976ockham}, 
The interplay between learning and compression has long been recognized,
%It has been 
popularized by Occam's razor rule of thumb \citep{ariew1976ockham},
%It has been 
and rigorously studied
%in several frameworks, including
%in the statistical learning framework 
in several frameworks,
%in the name 
%in the form of 
%of 
including in information theory in terms of the minimum description length principal and Kolmogorov complexity \citep{cover1999elements,li2008introduction},
and more recently in terms of sample compression schemes
%its modern
%the statistical learning framework 
%\citep{cover1999elements}, and up to its 
%the PAC formulation 
%framework
in the PAC statistical learning framework
%where compression serves a dual purpose: it reduces the computational complexity, and further acts as a regularizer so as to avoid statistical overfitting. 
%PAC 
\citep{warmuth86,floyd1995sample,graepel2005pac,GoKoNi14,david2016supervised,hanneke2019sharp,hanneke2019sample,bousquet2020proper,hanneke2021stable,alon2021theory}.
%Indeed, in these works, compression serves a dual purpose: it reduces the computational complexity; and further acts as a regularizer so as to avoid statistical overfitting. 

Consider for example the well-known support vector machine (SVM) \citep{vapnik2013nature}.
Given a binary labeled dataset, 
%it is sufficient to
%AK
%keep
%needs to
SVM retains
only the samples constituting the support vectors.
%AK
%In the case where opposingly-labeled data can be completely separated by a hyperplane,
In the {\em linearly realizable} case---%
%---i.e., 
where a hyperplane discriminator can achieve zero population error---%
the number of support vectors can be taken to be at most $d$, leading to ultrafast optimal error and compression rates of order $O(d/n)$, where $d$ is the dimension of the instance space $\R^d$ and $n$ is the number of independent labeled samples in the dataset.
More generally in the parametric setting, 
%it has been recently shown by that 
similar optimal rates are achieved
when one 
%has  
can construct
%what is known as 
a \emph{stable} compression scheme of size $d$ that is guaranteed to 
%almost surely 
obtain zero sample error for all $n$ while retaining only up to $d$ samples \citep{bousquet2020proper}.
%When the compression scheme is not stable, the error rates incur an additional factor logarithmic in $n$. 
%\todo{What is known for the agnostic parametric setting?}%
%\todo{Mention extension to margin \citep{hanneke2021stable}? \citet{crammer2002margin}?}
%Nevertheless, 

In the more general non-parametric 
%supervised 
%learning 
setting,
however,
it is well known that for \emph{any} learning rule,
%however, 
%classification rule 
the rate at which its error converges to the optimal one 
%excess error rate 
can be arbitrarily slow without 
further assumptions on the data distribution
%smoothness and tail conditions on the data distribution \citep{?}.
\citep{DeGyLu96}.
%arbitrariliy slow -rates
%However, 
The same holds for the achievable sample compression rates of accurate rules.
%Similarly, compression rates for accurate rules can be arbitrary slow.
% without any assumptions on the data distribution.
%Characterizations of
%interplay between 
Results on
the jointly
%interplay between 
%the 
achievable error and compression rates
% and the corresponding error rates
%, and their relation to the achievable error rates, 
in the non-parametric setting in terms of the properties of the data distribution and the instance space
%, and their relation to the corresponding error  rates,
are 
scarce
%, and the compression-error interplay in non-parametric statistics is
and still not well understood.
%In this paper 
Here we aim at narrowing this gap.
%However, similarly to the case of error rates, compression rates for accurate rules can be arbitrary slow without any assumptions on the data distribution.

\paragraph{Prototype learning rules.} A large family of non-parametric learning rules suitable for studying the error-compression interplay 
%between sample compression and error 
are prototype learning rules \cite[Chapter 19]{DeGyLu96}.
Such rules compress the data into a small number of prototypes and pair each prototype with a suitable label as computed from the dataset. 
A new instance is then labeled according to the label paired to its nearest 
%neighbor (NN) 
prototype 
%of the instance
in the 
%newly constructed 
compressed dataset.
The various prototype rules then differ in
%The choice of the prototype set and 
the way the prototypes and the paired labels are computed. 

%in the parametric setting, the interplay between the VC dimension of the hypothesis class under consideration and the accompanied uniform error and compression rates were studied in the PAC framework, leading to some beautiful results \citet{graepel2005pac,hanneke2019sharp,hanneke2019sample,hanneke2021stable,alon2021theory}.
%
%More generally in the parametric setting, the interplay between the VC dimension of the hypothesis class under consideration and the accompanied uniform error and compression rates were studied in the PAC framework, leading to some beautiful results \citet{graepel2005pac,hanneke2019sharp,hanneke2019sample,hanneke2021stable,alon2021theory}.
%However, in the non-parametric setting considered in this proposal, it is less clear what compression rates are achievable and how they depend on the data distribution.
%Prototype rules have been studied by .

%and in the paired labels are chosen differ among different prototype rules.
The simplest prototype rule is the 
%plain-vanilla 
1-nearest-neighbor rule (1-NN)  whose prototype-label set is simply the whole dataset.
This rule, however, is known to be inconsistent in general---%
%that is, 
its error does not necessarily converge to the optimal one as the sample size grows. 
%\citep{DeGyLu96}.
%While 
It is also known that the $k$-NN rule, which labels a new instance 
according to the label having the most counts among its $k$ nearest neighbors in the dataset, is consistent for \emph{any} data distribution, provided
%, with number of neighbors 
$k\to\infty$ such that $k/n\to0$.
In other words, 
%is known to be 
%universal 
%is consistent for \emph{any} data distribution---that is, 
$k$-NN is {universally consistent} in $\R^d$
for any 
$d>0$
%the Euclidean space 
\citep{DeGyLu96}.
%
%Another important property of a consistent classification rule is the rate at which the excess error
%% probability 
%%$\EXP\{L(g_{n})\} - L^*$ 
%convergences to 0 as $n\to\infty$.
%For \emph{any} 
%classification rule,
%%$g_n$, 
%this rate can be arbitrarily slow without 
%further assumptions on the data distribution

Beyond universal consistency, for several large families of data distributions, $k$-NN with a properly tuned 
%number of neighbors 
$k$ achieves \emph{minimax optimal} error rates---namely, the rate at which its error converges to the optimal one is also a
lower bound for \emph{any} other learning rule over the family of distributions under consideration.
In particular, under the commonly posed $\beta$-H{\"o}lder smoothness condition on the 
labels' conditional probabilities,
%regression functions, 
the strong density condition on the instance marginal distribution,
%a tail condition in the form of the strong density condition, 
and the $\alpha$-Tsybakov margin condition that bounds the total mass of points having large noise, the $k$-NN rule with the choice $k\approx n^{{2\beta}/{(2\beta+d)}}$ achieves the minimax optimal error rate of order $n^{-{\beta(1+\alpha)}/{(2\beta+d)}}$ \citep{AuTs05,ChDa14,GaKlMa16}.

While the $k$-NN rule does not attempt to compress the dataset, it has been recently shown by \citet{xue2018achieving} 
%(in the context of distributed computation) 
and \citet{GyWe21} that a natural prototype version of $k$-NN still enjoys the same optimal error rates as $k$-NN while retaining only $m=O(n/k)$ prototypes from the dataset (see also \citet{biau2010layered}
%,BiCeGu10
for results in the same spirit).
This rule, termed \hybrid{},
% in \citet{GyWe21}, 
randomly draws $m\approx n/k$ prototypes from the dataset and pairs each prototype with the label having the most counts among its $k$ nearest neighbors in the original dataset.
Notably, with 
$k\approx n^{{2\beta}/{(2\beta+d)}}$
%$k$ 
as above, the compression rate satisfies $m/n\approx n^{-{2\beta}/{(2\beta+d)}}\xrightarrow[]{n\to\infty}0$
%, the compression rate achieved is $m/n\approx n^{-{2\beta}/{(2\beta+2)}}$ 
while still enjoying the minimax error rate of order $n^{-{\beta(1+\alpha)}/{(2\beta+d)}}$.
%notably, $m/n\to 0$ as $n\to0$.
%Notably, the compression rate achieved satisfies $m/n\approxn^{-{2\beta}/{(2\beta+2)}}\xrightarrow[]{n\to\infty}0$.
%With $k$ as above, the compression rate achieved is $m/n\approx n^{-{2\beta}/{(2\beta+2)}}$ while still enjoying the minimax optimal error rate of order $n^{-{\beta(1+\alpha)}/{(2\beta+d)}}$;
%%notably, $m/n\to 0$ as $n\to0$.

%Evidently, 
%compression serves a dual purpose: it reduces the computational complexity, and further acts as a regularizer so as to avoid statistical overfitting. 

Another simple prototype rule is \protonn{} \citep{GyWe21}.
Similarly to \hybrid{}, this rule randomly draws $m\ll n$ prototypes from the dataset, but pairs each prototype with the label having the most counts among the samples from the dataset that fell into its Voronoi cell as determined by the other prototypes.
While at first glance it may seem that \protonn{} and \hybrid{} should behave similarly, 
it has been recently established by \citet{GyWe21} that,
in contrast to $k$-NN and \hybrid{}, \protonn{} 
%has been recently shown to be 
is universally consistent in \emph{any} metric space admitting a universally consistent learning rule, including many important 
%separable 
infinite-dimensional metric spaces for which $k$-NN and \hybrid{} fail to be consistent \citep{cerou2006nearest,GyWe21}.
%separable metric space---it is consistent converges to the optimal error 
%for any distribution in any separable metric space 
%Its error and compression rates were left open though.
%
Chronologically, \protonn{} was derived as a simplification of \optinet{}, a prototype rule that was first introduced in \citet{kontorovichsabatourner16}, further studied in \citet{KoSaWe17},
%\citep{kontorovichsabatourner16,KoSaWe17,HaKoSaWe20} 
and eventually shown in \citet{HaKoSaWe20} to be the first algorithm known to be universally consistent in any metric space that admits such a rule.
% a universally consistent rule.
%Similarly to \protonn{}, 
However, the error and compression rates for the Voronoi partition-based rules \protonn{} and \optinet{} were left open.

\paragraph{Main contributions.}
In this paper we continue the study of error and compression rates for prototype rules and focus on
%\variant{}, a close variant of 
\optinet{}
% that is more suitable for deriving error and compression rates 
for the case where instances reside in the familiar Euclidean space.
%\variant{} 
\optinet{}
selects its prototypes by computing a $\gamma$-net over the dataset for an appropriately tuned margin $\gamma>0$ and, similarly to \protonn{}, pairs each prototype with the label having the most counts among the samples that fell into its Voronoi cell.

%This prototype rule, termed \optinet{}, is the main subject of study in this paper as
%, which further motivated its study.
%Here 
%we take first steps in establishing its error and compression rates in the Euclidean instance space.

%$k$ nearest neighbors in the original dataset.

%In this paper
%We study jointly achievable error and compression rates for a close variant of
% another recently proposed prototype learning rule, termed 
%\optinet{} \citep{kontorovichsabatourner16,KoSaWe17,HaKoSaWe20}.
%In its original form, this rule first computes a $\gamma$-net of the dataset 
As our first main contribution, we establish 
both theoretically
%(Theorem \ref{thm}) 
and empirically that 
\optinet{}
%\variant{} 
achieves minimax optimal error rates under the aforementioned smoothness, margin, and tail conditions,
% on the data distribution, 
while enjoying compression rates 
similar
to those obtained 
%that are faster by a logarithmic factor as compared to those derived 
for \hybrid{} in \citet{xue2018achieving} and \cite{GyWe21}, and in some cases even faster rates.
%\todo{Need to verify this}
In fact, as established by \citet{GoKoNi14,chitnis2022refined},
% in some cases
%\variant{} 
\optinet{}
achieves \emph{near-optimal} compression rates in the sense that further compressing the dataset 
while remaining consistent on the dataset
%without altering the classifier on it
% the original dataset 
 is an NP-hard problem.
%\todo{It seems that the  geometric margin breaks the NP-hardness. Need to ask Lee-Ad or Aryeh}
%We still leave the rates achieved by \protonn{} open.

%The prototypes of \optinet{} are computed by a $\gamma$-net of the instances for an appropritaely tuned $\gamma>0$.
%Our variant of \optinet{} first computes a $\gamma$-net of the instances in the dataset,

%\paragraph{Related work.}
%One of the main motivations for studying \optinet{} is 
%%As a side remark, we note 
%that, in contrast to $k$-NN and \hybrid{}, \optinet{} has been recently shown to be universally consistent in any metric space that admits a universally consistent rule
%%separable metric space---it is consistent converges to the optimal error 
%%for any distribution in any separable metric space 
%\citep{HaKoSaWe20}.
%%, which further motivated its study.
%It error and compression rates were left open though.
%Here we take a first step in establishing its rates in the familiar Euclidean space.
%
%Beyond of the above main motivation,

%We then proceed to study a novel general compression scheme for further compressing prototype rules. 
Next, notably, the compression rate $m/n\approx n^{-{2\beta}/{(2\beta+d)}}$ derived 
%in \citet{xue2018achieving} and \citet{GyWe21} 
for \hybrid{}, as well as those derived for 
\optinet{}
%\variant{} 
in this paper, are insensitive to the Tsybakov noise parameter $\alpha$ that restricts the mass of points having high noise level.
This stems from the fact that these rules do not attempt to focus their resources on the decision boundary where classification is harder.
%Intuitively, classification is harder near the decision boundary and, with limited resources, a classifier should put more effort near it as compared to regions where the correct label is stable.
%In the Euclidean space, 
%The latter 
This approach was made practical by several adaptive learning algorithms, such as decision trees \citep{scott2006minimax,blanchard2007optimal}, random forests \citep{lin2006random,biau2008consistency,biau2010layered}, as well as other hierarchical tree-based
\citep{kpotufe2012tree,binev2014classification} 
and compression-based \citep{kusner2014stochastic}
algorithms.
However, as far as we know, no compression rates have been established for any of those algorithms.
%\todo{verify}

As our second main contribution, we study \compalg, a new general and simple non-lossy compression scheme for further compressing prototype rules by removing spurious prototypes that are far from the decision boundary. 
Applying it to 
\optinet{},
%\variant, 
we show both theoretically and empirically that under an additional 
%mild 
geometric 
%noise
margin
condition, further gain in the compression rate is achieved
%by adapting to the decision boundary, 
without sacrificing accuracy.

\section{Problem setup}
%Let $(\X,\rho)$ be a separable metric space, equipped with its Borel $\sigma$-field.
%In the following, we take 
Our instance space is $\X=\R^d$ equipped with 
%the Borel $\sigma$-field as induced by 
the Euclidean metric $\rho(x,y) = {\lVert x - y\rVert}_2$, $x,y\in\X$.
%\todo{Omer: Add reference if the paper is split.}
Assume that the feature element $X$ takes values in $\X$ and let its label
$Y$ take values in  $\Y = \{1,\dots ,M\}$.
If $g:\X\to\Y$ is an arbitrary measurable decision function
then its error probability is 
%denoted by
\[
L(g)=\PROB\{g(X)\ne Y\}.
\]
Denote by $\dist$ the 
%unknown 
probability distribution of $(X,Y)$ and let $\mu$ be the marginal distribution of $X$ and
\[
P_j(x)=\PROB\{Y=j\mid X=x\},\qquad
j\in\Y
%=1,\dots ,M
.
\]
Then the Bayes decision
$g^*(x) = \argmax_{j\in\Y}  P_j(x)$
% $g^*$,
%\begin{align*}
%%\label{eq:Bayes-decision}
%g^*(x) = \argmax_{j\in\Y}  P_j(x),
%\end{align*}
minimizes the error probability over all measurable classifiers.
Its error, also known as the Bayes-optimal error, is denoted by
\(
L^*=\PROB\{g^*(X)\ne Y\}.
\)

In the standard model of pattern recognition, 
%$\dist$ and
$g^*$ and $L^*$ are unknown and
a learner is 
given instead a labeled 
dataset
consisting of $n$
independent samples of $(X,Y)$,
%copies
%samples 
%$\D_n=\{(X_1,Y_1), \dots, (X_n,Y_n)\}$ 
%$\D_n=\{(X_i,Y_i)\}_{i=1}^n$ 
\[
\Dl=\{(X_1,Y_1), \dots, (X_n,Y_n)\}
= (\bX_n,\bY_n).
\]
Based on $\D_n$, one constructs a classifier
% rule 
 $g_n:\X\to\Y$.
%Based on $\D_n$, one can estimate the regression functions $P_j$ by some $P_{n,j}$,
%%, $j\in\Y$, 
%and the plug-in classification rule $g_n$ derived from $P_{n,j}$ is
%\(
%g_n(x) = \argmax_{j\in\Y} P_{n,j}(x) .
%\)
%The classifier 
The rule $g_n$ is weakly consistent for a distribution $\dist$ if
%$\EXP\{L(g_{n})\} \xrightarrow[n\to\infty]{} L^*.$
%\begin{align*}
%\(
$\lim_{n\to\infty} \EXP\{L(g_{n})\} = L^*$.
%\)
%\end{align*}
It is strongly consistent for $\dist$ if
%\begin{align*}
%\(
%\PROB\{\lim_{n\to\infty} L(g_{n}) = L^* \} = 1.
%\)
%\end{align*}
%$$L(g_{n}) \xrightarrow[n\to\infty]{} L^*\quad\text{almost surely}.$$
%$L(g_{n}) \xrightarrow[n\to\infty]{} L^*$
$\lim_{n\to\infty} L(g_{n}) = L^*$
almost surely.
% (a.s.).
The rule 
$g_n$ is \emph{universally consistent} 
in the space $(\X,\rho)$
%the space 
if it is consistent for \emph{any} distribution 
%$\dist$ 
over the 
%Borel $\sigma$-field of the 
product space $\X\times\Y$ equipped with the Borel $\sigma$-field.

\paragraph{Notation.}
We use standard $O$-notation and use $\widetilde O$ 
to hide some logarithmic factors.
In the following, constants such as $C,c$ etc.\ may change from line to line even in the same equation, and in general may depend on the dimension $d$ and other parameters.
The characteristic function $\IND{\cdot}$ is 1 if its argument is true and 0 otherwise.
$\Ball{x}{\gamma}= \{x'\in\X : \rho(x,x')<\gamma\}$
is the open ball around $x$ with radius $\gamma\geq0$.
Table \ref{table:notation}
%\todo{}
in the supplementary material summarizes the main notation used below.
All proofs are deferred to the supplementary material.

%In the standard model of pattern recognition, 
%\section{A prototype learning rule}
\subsection{Prototype learning rules}
%\section{PROTOTYPE LEARNING RULES}
%\paragraph{Prototype rules.}
For simplicity, assume that in addition to $\D_n=(\bX_n,\bY_n)$, we are also given $\ul\ll n$ unlabeled samples \(
\Dul=\left\{X'_1, \dots, X'_\ul\right\}
\)
which are
independent and identical samples of $X$.
%copies
%and additional $n$ training labeled samples which are independent of $\Dul$ and independent identical copies of $(X,Y)$,
%\[
%\Dl=\{(X_1,Y_1), \dots, (X_n,Y_n)\}.
%\]
%%We denote the instances obtained from the labeled set $\Dl$ by $\bX_n = \{X_1,\dots,X_n\}$.
%Based on the samples, one can estimate the regression
%functions $P_j$ by some
%$P_{n,\ul,j}$, 
%%$\tilde P_{j}$, 
%$j=1,\dots ,M$, and the corresponding
%plug-in classification rule $g_{n,m}$ derived from 
%$P_{n,\ul,j}$
%%$\tilde P_{j}$ 
%is
%% defined by
%\[
%g_{n,m}(x) = \argmax_{j\in\Y} 
%P_{n,\ul,j}(x)
%%\tilde  P_{j}(x)
%.
%\]
%%\todo{Note that we don't use $g_{n,m}$ in the sequel.}
%%\todo{We actually didn't use what was here, so I removed it.}
%The classifier $g_{n,m}$ is strongly Bayes consistent if
%\[
%\P\{\lim_{n,m\to \infty} \}
%\]
%A large family of non-parametric methods consists of prototype rules.
%Given a distance function $\rho$, for
% a sequence
%\textcolor{violet}{[maybe ``sequence" or ``enumerated set" instead]}
%(Sampling $\bX'_m$ randomly from $\bX_n$ slightly complicates the analysis without much gain in insight.)
All prototype rules considered in this paper select their set of prototypes as a subset  $\tbX=\{\tX_1, \dots,\tX_{\tm}\}\subseteq \bX_m'$ of a possibly data-dependent size $\tm = |\tbX|\leq m$.
%(Selecting the prototypes randomly from $\bX_n$ slightly complicates the analysis without altering much our main conclusions.)
%
%Given 
%a set of prototypes $\bX=\{x_1,\ldots,x_{m}\}\subseteq\X$, 
For $i\in\{1,\dots,\tm\}$ and $x\in\X$ let
%$X^{(i)}(x;\tbX)$
$\Xinn{i}(x;\tbX)$
be the $i$th
%$X_{(1)}(x,\bX)= \argmin_{x'\in\bX} \rho(x,x')$ the 
nearest neighbor of $x$ in $\tbX$, 
%$ \Xinn{1}(x;\tbX) \leq \dots \leq \Xinn{\tm}(x;\tbX),$
breaking ties 
towards
%according to 
the prototype 
%in $\bX$ 
with the smaller index in $\bX_m'$.
The prototypes in $\tbX$ induce a {Voronoi partition} of $\X$, denoted 
$$\V(\tbX) = \{V_1(\tbX),\dots, V_{\tm}(\tbX)\},$$
where the Voronoi cell numbered $\ell\in\{1,\dots,\tm\}$ is
%$V_\ell(\bX) = \{x \in \X :x_\ell =  X_{(1)}(x,\bX)\}$, $\ell\in\{1,\dots,m\}$;
\begin{align*}
V_\ell(\tbX) = \{x \in \X :\tX_\ell =  \Xinn{1}(x;\tbX)\} 
%\,\, \ell\in\{1,\dots,\tm\}
.
\end{align*}
%breaking ties lexicographically.
%$V(x;\tbX)$ or $V_x(\tbX)$ denote the cell in which $x$ resides.
%
To obtain a 1-NN rule, each prototype $\tX_\ell\in\tbX$ is paired with the label 
$\tY_\ell = \argmax_{j\in\Y} P_{n,\ell,j}$
where $P_{n,\ell,j}$ is the score for label $j\in\Y$ estimated from the data, aiming at relatively estimating the probabilities $P_j$ at a local neighborhood of $\tX_\ell$.
%as determined by some local estimators $P_{n,\ell,j}$ of the regression functions $P_j$, 
This results in a compressed labeled set
% of prototype-label pairs 
$$\tD=(\tbX,\tbY).$$
%\footnote{
%%The notation $\tD$ represents a compressed labeled data set, where, its concrete definition varies and is defined before used. %in the relevant places.
%$\tD$ is a notation for a general compressed sample, where, its concrete definition varies and is defined before used. %in the relevant places.
%}
Denoting $\Xinn{i}(x;\tD) = \Xinn{i}(x;\tbX)$,
%$=\argmin_{x'\in\bX} \rho(x,x')$ 
and $\Yinn{i}(x;\tD)$ as the label paired to $\Xinn{i}(x;\tbX)$ in $\tD$,
the corresponding prototype rule is
$$g_n(x) = \Yinn{1}(x;\tD), \qquad x\in\X.$$
Its expected error 
%probability
%, averaged over the entire dataset, 
is
%$ \EXP\{\PROB\{g_n(X)\neq Y | \bX'_m,\bX_n,\bY_n\}\}$
$\PROB\{g_n(X)\neq Y \}$
and 
its compression rate is 
%$\EXP\{\frac{|\tD|}{|\Dl|}\} = \frac{\EXP\{|\tD|\}}{n}$.
$$\EXP\{|\tD|/|\Dl|\} = \EXP\{|\tD|\}/n.$$
%
%The label $Y_\ell$ is determined by some local estimation $P_{n,\ell,j}$ of the regression functions $P_j$.

For both \protonn{} and \hybrid{} the prototype set is taken as $\tbX=\Xul$.
%With the convention that $0/0=0$, 
The local relative estimators $P_{n,\ell,j}$ of $P_j$ computed by \textbf{\protonn{}} simply count the number of times each label $j\in\Y$ has been observed among the samples in $\D_n$ that fell into the cell $V_\ell(\tbX)$,
% corresponding to prototype $\tX_\ell$,
\begin{align*}
%\label{eq:proto_est}
%\frac{
\sum_{i=1}^n \IND{Y_i=j,X_i\in V_{\ell}(\tbX)},
%}{\sum_{i=1}^n \IND_{\{X_i\in V_{\ell}(\tbX)\}}},
%\quad 
%\ell\in\{1,\dots,m\}, 
\qquad
j\in\Y.
\end{align*}
%such that $0/0=0$ by definition.
%while those
%$P_{n,\ell,j}$ 
For \textbf{\hybrid{}}, these are counted among the $k$ nearest neighbors of $\tX_\ell$ in $\D_n$, 
\begin{align*}
%\label{eq:proto_est}
%\frac{1}{k}
\sum_{i=1}^k \IND{\Yinn{i}(\widetilde X_{\ell};\D_n)=j},
\qquad 
j\in\Y.
%\ell\in\{1,\dots,m\}.
\end{align*}
The construction time of \protonn{} and \hybrid{} is $O(m n)$ and 
%its query time is $O(m)$.
a query takes $O(m)$ time.

In this paper we focus on 
%a variant of 
%the 
\textbf{\optinet{}}	
%algorithm 
%which selects its prototypes differently
\citep{kontorovichsabatourner16,KoSaWe17,HaKoSaWe20}.
%which we call \textbf{\variant{}}.
For margin $\gamma = \gamma(n) >0$ 
%and $\ul = \ul(n)\in \N$ 
to be chosen below, 
\optinet{}
%\variant{} 
first constructs 
%Let $\bX_n (\gamma)$ be 
a $\gamma$-net 
%$\Xg(\gamma)$ 
of the unlabeled samples $\Xul$,
namely,
%a $\gamma$-{net} of $\Dul$ is 
%corresponding to
any \emph{maximal} set $\bXg(\gamma)\subseteq \Xul$ in which all interpoint distances are at least $\gamma$. 
The $\gamma$-net obtained constitutes the prototype set $\tbX$ of 
\optinet{},
%\variant{},
%is 
%denoted by
\begin{align*}
%\label{def:Xg(gamma)}
\tbX = \bXg(\gamma)=\{\Xg_1(\gamma),\dots, \Xg_{m(\gamma)}(\gamma)\}
%\subseteq \Xul
\end{align*}
where $m(\gamma)=|\bXg(\gamma)|\leq m$ denotes the data-dependent size of the $\gamma$-net.
%$\Xul (\gamma)$  
%of the unlabeled samples $\Dul$.
%$\bX_n :=\{X_1,\dots ,X_n\}$ 
%with size $M_n(\gamma)$.
This net induces
%Introduce 
a 
%data-driven 
Voronoi partition $\V(\bXg(\gamma))=\{V_1(\gamma),\dots,V_{m(\gamma)}(\gamma)\}$
 of $\X$.
%,
%$$\Vor_{\ul,\gamma}:= \{V_1,\dots, V_{\ns}\},$$
%$\Vor$
%$\P_{n,\gamma}$ 
% := \{V_1(\Xul),\dots, V_{m}(\Xul)\}$,
% such that $\P_{n,\gamma}$ is a Voronoi partition, 
%where the 
%whose nuclei set is
%$\Xul(\gamma):=\{X_1(\gamma),\dots, X_{\ns}(\gamma)\}\subseteq \Xul$
%$\Xg(\gamma)$ 
%and 
%\begin{align*}
%V_i 
%%= V_j(\Xul) 
%:= \left\{x \in \X : X_i(\gamma) = 
%%\argmin_{1\le j\le \ns} 
%X_{(1)}(x,\Xg(\gamma)) \right\}.
%%V_i = V_i(\Xul) := \left\{x \in \X : i = \argmin_{1\le j\le \ns} \rho(x,X_j(\gamma)) \right\}
%\end{align*}
%\textcolor{violet}{\sout{breaking ties lexicographically.}}
%\todo{deal with tie-breaking}
%where each Voronoi cell is
%\begin{align*}
%V_i(\Xul) := \{x \in \X : i = \argmin_{1\le j\le m} \rho(x,x_j) \},
%\end{align*}
%breaking ties lexicographically.
%where $\argmin_{1\le j\le \ns} \rho(x,X_j(\gamma))$ is the set of all nearest neighbors of $x$ in $\Xg$.
%The prototype 
%nearest neighbor 
%classification rule that we consider here 
Similarly to \protonn{},
\optinet{}
%\variant{} 
estimates $P_j$ 
%(relatively) 
%based on 
%%the labeled sample 
%$\Dl$ as follows.
%Define the 
%%$\UB{\gamma}$ is the
%$\gamma$-envelope around 
%%the unlabeled samples 
%$\Xul$ by
%% defined by
%%\todo{Take close ball?}
%%\todo{Need to change it to $\bigcup_{X' \in \Xg(\gamma)} S_{X', 2\gamma}$?}
%\begin{align}
%\label{eq:UB}
%\UB{\gamma} 
%%= \UB{\gamma}(\Xul) 
%=
%%\bigcup_{X'_i \in \Xul} 
%\bigcup_{i = 1}^m 
%\Ball{X'_i}{\gamma}
%%S_{X'_i, \gamma}
%\end{align}
%where 
%$\Ball{x}{\gamma}= \{x'\in\X : \rho(x,x')<\gamma\}$
%is the open ball around $x$ with radius $\gamma$.
%%,$\Ball{x}{\gamma}= \{x'\in\X : \rho(x,x')<\gamma\}.$
%%denotes the open sphere centered at $x$ with radius $\gamma$.
%%
%%\optinet{} estimates $P_j$ 
%%with $P_{n,\ell,j}$
%% of 
%For cell
%%label $j\in\Y$ and 
%%prototype $\Xg_\ell(\gamma)\in\bXg(\gamma)$ with 
%$\ell\in\{1,\dots,m(\gamma)\}$, $P_j$ is estimated 
by counting the labels from $\Dl$ that fell in the Voronoi cell $V_{\ell}(\gamma)$,
%as restricted to $\UB{\gamma}$,
%that is,
\begin{align}
\label{gest}
%\label{eq:proto_est}
P_{n,\ell,j}=
%\frac{
\sum_{i=1}^n \IND{Y_i=j,X_i\in V_{\ell}(\gamma) 
%\cap \UB{\gamma}
}
%}{\sum_{i=1}^n \IND_{\{X_i\in V_{\ell}(\gamma) \cap \UB{\gamma}\}}}
.
%\qquad x\in V_\ell,
%\quad 
%\ell\in\{1,\dots,m\}, 
%\quad
%j\in\Y,
%\mbox{ if } x\in V(x,\bX(\gamma)),
\end{align}
%========================
%If $V_x$ is the cell of the partition $\Vor_{\ul,\gamma}$ for which $x\in V_x$, then
%$P_j(x)$ is estimated by
%the piecewise constant function
%\begin{align}
%\label{gest}
%P_{n,j}(x) :=\frac{\frac{1}{n}\sum_{X_i\in \Xl} \IND_{\{Y_i=j,X_i\in V_x\cap \UB{\gamma}\}}}{\mu(V_x \cap \UB{\gamma}) }.
%\end{align}
The
prototype
$\Xg_\ell(\gamma)\in\bXg(\gamma)$
%As before, each prototype 
%$\tX_\ell(\gamma)\in\Xg(\gamma)$ 
is then paired with the label
% $Y_i(\gamma) \in \Y$ corresponding to the plurality vote over the samples from $\D_n$ that fell into 
%$V_i \cap \UB{\gamma}$, 
%namely,
$\Yg_\ell(\gamma) = \argmax_{j\in\Y} P_{n,\ell,j},$
%\begin{align*}
%\tY_\ell(\gamma) := \argmax_{j\in\Y} P_{n,\ell,j},
%%(X_i(\gamma)),
%\end{align*}
%breaking ties towards smaller labels,
leading to a labeled set 
$$\Dg(\gamma) = (\bXg(\gamma),\bYg(\gamma))$$
 of size $m(\gamma)=|\Dg(\gamma)|$.
%\begin{align*}
%\D(\gamma) &:= (\bX(\gamma),\bY(\gamma)) 
%\\
%& = \{(X_1(\gamma),Y_1(\gamma)),\dots,(X_l(\gamma),Y_l(\gamma)) \}.
%\end{align*}
The prototype 
%nearest-neighbor 
classification rule is then
\begin{equation}
\label{gdec}
%g_{n}(x) = 
\goptinet_{n,\ul,\gamma}(x) = 
\Yinn{1}(x; \Dg(\gamma))
%\argmax_{j\in\Y}  P_{n,j}(x)
,\qquad x\in\X.
\end{equation}
%\begin{equation}
%\label{gdec}
%g_{n}(x) = g_{n,\ul,\gamma}(x) := \argmax_{j\in\Y}  P_{n,j}(x).
%\end{equation}
%This classification rule is just the plurality vote within a restricted version of the cell.
%As will become clear below, 

The construction time of 
\optinet{}
%\variant{} 
is $O(n m)$ and a query time is of order $m(\gamma) \leq m$ which depends on the margin $\gamma$ chosen at construction.

\begin{remark}
The universal consistency of \optinet{} has been established in \cite{HaKoSaWe20} by performing a model selection procedure over $\gamma$, for example, by a validation procedure. For general metric spaces such a procedure is unavoidable (this is in contrast to \protonn{}). However, in finite dimension, one can show that any sequence $\gamma_n\to 0$ such that $n\gamma^d_n \to \infty$ ensures universal consistency. Below we set $\gamma_n$ explicitly to ensure minimax error rates.
%The restriction to $\UB{\gamma}$ in \eqref{gest} does not appear in \optinet{} and it is not needed to establish its universal consistency.
%% \citep{HaKoSaWe20}. 
%It is added to \variant{} to better control the bias of $P_{n,\ell,j}$ in estimating $P_j$ for Voronoi cells of unbounded diameter.
%Whether one can remove this restriction in what follows is left open.
\end{remark}

\section{Error and compression}
%\section{Error and Compression Rates}
%\section{ ERROR AND COMPRESSION}
%\section{Error and compression}
In this section we study rates of convergence of the excess error probability $\EXP\{L(\goptinet_{n,\ul,\gamma})\}-L^*$ and the compression ratio $\EXP\{|\Dg(\gamma)|\}/n$
%, where $g_{n}=\goptinet_{n,\ul,\gamma}$
% = g_{n,\ul,\gamma_{n,\ul}}$ 
for the 
%$\gamma$-net partition 
classifier in \eqref{gdec}.
To obtain non-trivial rates one needs to impose some conditions on 
%the distribution of
% the labeled pair 
$(X,Y)$ \citep{DeGy85}.
% either the marginal distribution $\mu$ of $X$ or the regression functions $P_j$.
% and $\gamma_n$ is specified below.
%
%As a first step, we take $\X=\R^d$.
%In addition to a margin condition, we also assume that
%the marginal distribution $\mu$ of $X$ has a density $f$ w.r.t.\ the Lebesgue measure $\lambda$ and that $f$ satisfies the Strong Density Condition (see definitions below).
%
We first assume that
the marginal distribution $\mu$ of $X$ has a density 
%$f$ 
with respect to the Lebesgue measure $\lambda$ that 
%$f$ 
satisfies the
minimal mass condition (\text{MMC}) \citep{AuTs05}.
%\todo{ref. Omer: To what paper? AuTs05?}
%For now we also simplify the modified Lipschitz condition to the standard one,
%\begin{equation}
%\label{Hold_standard}
%|P_j(x)-P_j(z)|\le
%C ||x-z||.
%%h(\mu(S_{x,\rho(x,z)})),
%\end{equation}

\begin{definition}
\label{def:MMA}
The distribution $\mu$ of $X$ with density $f$ satisfies the \textit{minimal mass condition}
%(MMA)
 if there exist $\kappa>0$ and $\gamma_0>0$ such that $\forall \gamma\leq\gamma_0$,
\begin{equation}
\label{MMA}
\PROB\{X\in 
%S_{x,\gamma}
\Ball{x}{\gamma}
\} \geq \kappa f(x)\gamma^d,
\qquad
\forall x\in\X.
\end{equation}
%where $S_{x,\gamma}$ denotes the sphere centered at $x$ and having the radius $\gamma$.
\end{definition}
%%\paragraph{Tail condition.} 
%\begin{definition}
%The distribution $\mu$ of $X$ with density $f$ satisfies the weak tail condition if there exist 
%$\chi>0$ and $\eta_0>0$ such that $\forall \eta\leq\eta_0$,
%%a decreasing function $\varphi:\R^+\to (0,1]$ and $\eta_0>0$ such that $\varphi(\eta)\downarrow0$ as $\eta\downarrow0$ and $\forall\eta<\eta_0$,
%\begin{align}
%\label{tail}
%%F(\eta) = 
%\PROB\{f(X) \leq \eta\} \leq
%\eta^\chi
%%\varphi(\eta)
%.
%\end{align}
%\end{definition}
%The limit $\chi\to0$ corresponds to the assumption the $f$ is bounded away from zero.
We also assume that the density $f$ is 
\textit{bounded away from zero} (\text{BAZ}), namely $\exists \dlb>0$ such that
\citep{AuTs05}
\begin{equation}
f(x) \geq \dlb,
\qquad \forall x\in\supp(\mu),
\label{BAFZ}
\end{equation}
where 
\begin{align*}
%$
\supp(\mu)=\{x\in\X : \mu(\Ball{x}{r})>0, \forall r>0\}
%$
\end{align*}
is the support of $\mu$.
%\begin{equation*}
%\supp(\mu) =\{x\in\X : \mu(\Ball{x}{r})>0, \forall r>0\}.
%%\label{supp}
%\end{equation*}
%
The BAZ condition 
%that $f$ is bounded away from zero (BAZ) 
together with the MMC are known to be equivalent to the \emph{strong density condition}\label{SDC} (\textbf{SDC}) \citep{GaKlMa16}, so from now on we refer to the conjunction of MMC and BAZ as SDC.

%Lastly, 
We also make the standard assumption that the $P_j$s 
%label's conditional probabilities 
are H\"older continuous\label{def:Holder}, that is, there are $C>0$ and $0<\beta\leq 1$ such that for all $x,x'\in\X$,
%condition to the standard one,
\begin{equation*}
%\label{Hold_standard}
|P_j(x')-P_j(x)|\le
C \rho(x,x')^\beta.
% = C ||x-z||.
%h(\mu(S_{x,\rho(x,z)})),
\end{equation*}

Lastly,
for the two-class setup, the Tsybakov margin condition has been  investigated by
\citet{MaTs99}, \citet{Tsy04}, \citet{AuTs05}.
This condition allows for faster error rates than those achievable for density estimation and real-valued regression.
%Puchkin and Spokoiny 
\citet{xue2018achieving,PuSp20,GyWe21} generalized 
this
%the margin 
condition to multiclass.
%Let $\reg:\X\to \R^+$ be given by
%\begin{align}
%\label{eq:reg}
%\reg(x) := P_{(1)}(x)-P_{(2)}(x) \geq 0.
%\end{align}

\begin{definition}
\label{def:Tsybakov}
Let $P_{(1)}(x)\ge \dots \ge P_{(M)}(x)$ be the ordered values of the conditionals 
$P_{1}(x), \dots , P_{M}(x)$,
breaking ties lexicography,
and define the margin 
%$\reg:\X\to \R^+$ be given by
\begin{align}
\label{eq:reg}
\reg(x) = P_{(1)}(x)-P_{(2)}(x) \geq 0.
\end{align}
Then the Tsybakov margin condition  means  that
there are $\alpha>0$ and $c^*>0$ such that
% for all $0 < t \leq 1$,
\begin{equation}
\label{eq:Tsybakov}
\PROB\left\{\reg(X)\le t\right\}
\le c^* t^{\alpha},
\qquad 0 < t \leq 1.
\end{equation}
%where $\alpha>0$ and $c^*>0$.
\end{definition}

The SDC condition implies that the support $\supp(\mu)$ is bounded.
Larger $\beta$ means smoother  
%regression functions 
$P_j$s and larger $\alpha$ means less mass of points have high noise levels. 
For the two-class problem, \citet{AuTs05} showed that under the SDC and the margin condition with $\alpha\beta\leq d$, the minimax optimal error rate of convergence for the class of $\beta$-H\"older-continuous $P_j$s is of order
\begin{align}
\label{eq:minimax_rate}
%\widetilde O\Big(
n^{-\frac{\beta(1+\alpha)}{2\beta+d}}
%\Big)
;
\end{align}
i.e., this rate is a \emph{lower bound} for {\it any} classifier.
%They are compatible when $\alpha\beta\leq d$ \citep{?}.

\begin{theorem}
\label{thm}
%Let 
%$\X=\R^d$ and 
%$\gamma>0$ and 
Assume the marginal distribution $\mu$ of $X$ has a density 
%$f$ 
that satisfies the 
strong density condition with $\gamma_0>0$
%{minimal mass condition} and is bounded away from zero 
\textup{(\textbf{SDC})}.
%the tail condition.
%$f$ is bounded away from zero.
%the distribution function $H_x(\cdot)$ is continuous for each $x$.
If the Tsybakov margin condition is satisfied with $\alpha>0$ and the
%generalized
H\"older continuity condition is met with
%parameters 
$0<\beta\leq 1$,
% and $C>0$,
then for any $0<\gamma\leq\gamma_0$,
\begin{align}
&\EXP\{L(\goptinet_{n,\ul,\gamma})\}-L^*=
\label{eq:err_rate}
	\\
&\quad
%= 
 O\Big(\big(n\gamma^{d}\big)^{-\frac{1+\alpha}{2}}\Big)
% \\
% &
+ O\Big( \gamma^{\beta(1+\alpha)} \Big)	
\nonumber
%\\&\qquad
%\\& \quad
+ \exp\!\Big(\!-\Omega(\ul \gamma^d) \Big).
\end{align}
%\todo{Omer: Is the indentation appropriate?}
\end{theorem}

The first two terms in \eqref{eq:err_rate}
%the rates of Theorem \ref{thm} 
are the variance and bias terms respectively. The last term stems from the fact that the partition defining the classifier is completely data driven.
Setting 
in Theorem \ref{thm} 
%$\ul = n^{\frac{d}{d+2}}\log n$ and $\gamma = n^{-1/(d+2)}$
%$\ul$ and $\gamma$ 
%as
\begin{align}
\gamma &= \gamma_n  = n^{-\frac{1}{2\beta+d}},
%\qquad\text{and}\qquad
\label{eq:gamma_n and m_n}
\\
\nonumber 
\ul 
&
=  \ul_n 
=
\frac{\log\!\big(\gamma_n^{-\beta(1+\alpha)}\big)}
{\gamma_n^{d}}
%\approx 
= O\left(
n^{\frac{d}{2\beta+d}}
\log n
\right),
\end{align}
%%\begin{align*}
%%J_{n,2,j,\ell} \leq c^* n^{-\frac{\al\dfrac{•}{•}pha+1}{d+2}} + e^{-cn^{\frac{2}{2+d}}}.
%%\end{align*}
%in Theorem \ref{thm} 
yields, up to a logarithmic factor, the optimal minimax error rate
\eqref{eq:minimax_rate}.
%% of order
%\citep{AuTs05}
%\[
%%\widetilde O\Big(
%n^{-\frac{\beta(1+\alpha)}{2\beta+d}}
%%\Big)
%.
%\]
The compression rate satisfies
%is bounded by
\begin{align*}
%\label{eq:comp_rate}
\frac{|\D(\gamma_n)|}{n}\leq\frac{m_n}{n} = \widetilde O\Big(n^{-\frac{2\beta}{2\beta+d}}\Big) 
\xrightarrow[n\to\infty]{} 0.
%\qquad \text{as } n\to\infty.
\end{align*}
The right hand side of this upper bound is the compression rate obtained for \hybrid{} in \citet{xue2018achieving}.
For 
\optinet{}
%\variant{} 
we obtain a faster bound by a logarithmic factor.
%For a tighter bound, 
For $S\subseteq\X$ let $\CN_\gamma(S)\in\mathbb N$ be the maximal cardinality of a $\gamma$-net over $S$.
Then, under the SDC, and more generally when the support $\supp(\mu)$ is bounded, there is a constant $C=C(\mu)$ such that the size of any $\gamma_n$-net of $\supp(\mu)$ 
satisfies
%is bounded 
%One can show that
\citep{KL04}
%such that
\begin{align}
\label{eq:cn_bound}
\frac{|\D(\gamma_n)|}{n} \leq \frac{\CN_{\gamma_n}(\supp(\mu))}{n} \leq \frac{1}{n}\left(
%\frac{2\diam(S)}
\frac{C}
{\gamma_n}\right)^d = O\Big(n^{-\frac{2\beta}{2\beta+d}}\Big).
\end{align}
In particular, $C\leq 2\,{\diam(S)} = 2\sup_{x,x'\in S} \rho(x,x')$.
Table \ref{table:rates}
%\todo{}
%\ref{table:notation}
in the supplementary material
summarizes the error and compression rates available for the various prototype rules.
% is brought in Table 
%\todo{Fix the place of this sentence}

%We currently do not know whether this faster rate holds also for \hybrid{}.
%\todo{Verify}
%Up to logarithmic factors, the upper bound in \eqref{eq:comp_rate} on the compression rate of \optinet{} is the compression rate obtained for \hybrid{}.
%However, that of \optinet{} can be significantly faster in practice.
%\todo{When?}

%, for example when the distribution is such that large mass is concentrated ???

%\textcolor{blue}{Say what the compression ratio is for the standard OptiNet.}

%\textcolor{blue}{Move proofs to new section}

%\section{Further Compression}
%\section{ FURTHER COMPRESSION}
\section{Further compression}
\label{sec:further_compression}
Evidently, the compression rate in \eqref{eq:cn_bound}
% that bounds from above 
%for \optinet{}, as well as that for \hybrid{} \citep{xue2018achieving,GyWe21},
%, and also  \hybrid{} 
%and bounds from above that of \optinet{} 
is insensitive to the Tsybakov margin parameter $\alpha$.
This is not surprising, since 
\optinet{},
%\variant{}, 
as well as 
%\todo{Omer: Verify with Roi this above sentence.}
\hybrid{} and \protonn{}, do not attempt to
adapt to the noise level
%be 
\emph{locally}.
% adaptive 
%
% but rather globally in a worst case manner.
%focus their resources on the decision boundary where classification is harder.
%Indeed, 
The Tsybakov condition \eqref{eq:Tsybakov} restricts the total mass of points $x$ having large noise
%The latter is
as 
manifested by a small margin $\eta(x)=P_{(1)}(x)-P_{(2)}(x)$ (see \eqref{eq:reg} and \eqref{eq:Tsybakov}).
So when $\eta$ is also smooth (such as having H{\"o}lder parameter $\beta=1$) 
one may expect
% that 
%under the Tsybakov margin condition 
%there will be 
large regions in the support in which the Bayes-optimal label $g^*(x)$ 
%in \eqref{eq:Bayes-decision} 
is unaltered.
%, forming (possibly overlapping) clusters of different optimal labels. 

One approach to leverage such conditions
would be to try and designate a prototype for each region in which $g^*$ is stable.
% should suffice to correctly classify all points in each region.
This approach is taken, for example, by the well-known $k$-means algorithm \citep{macqueen1967some} and by vector quantization algorithms, including the more recent deep learning ones such as Prototypical Networks \citep{snell2017prototypical}. 
%\todo{Omer: added reference}
%Its compression however is typically lossy (as opposed to accurate).
%In some sense, it is also the intuition behind the fast hierarchical compression heuristic proposed by \citet{GoKoNi14}.
% for exact compression.
%which no compression guarantees have been established so far.
However, when the decision boundary is not well behaved, this approach may lead to a significant degradation in accuracy. 
In such cases, an alternative non-parametric approach would be to focus around the decision boundary where $\eta$ is small.
%to form a ``blanket'' of prototypes around the decision boundary where $\eta$ is small. 
The adaptive algorithms mentioned in the Introduction 
%\ref{chapter:intro} 
follow this approach.
However, as far as we know, no compression rates have been established for any of those algorithms.
%\todo{Aryeh's paper on average margin?}
%So it seems that the compression rate above is ``suboptimal''. 
%under the  condition.
%
%Studying an appropriately combined compression-error optimality criterion is beyond the scope of this paper. 

Here we follow the non-parametric approach above
%take the above intuition 
and study a new compression scheme for general prototype rules that further compresses the prototype set 
by removing prototypes lying inside those regions where $g^*$ is stable, 
essentially 
forming a ``blanket'' of prototypes around the decision boundary.

To introduce our prototype compression scheme, which we term \textbf{\compalg{}}, consider a finite labeled set $\D' = (\bX',\bY')$ where the instances in $\bX'$ are distinct.
For any $X'\in \bX'$ we denote by $Y'(X')\in\bY'$ its corresponding label in $\D'$.
Simply put, a prototype-label pair $(X',Y')\in\D'$ is removed 
from $\D'$ 
if the labels of all its neighboring cells have the same label as $Y'$.
Formally, let $V_x(\bX')$ be the cell containing $x$ in the Voronoi partition induced by $\bX'$.
For $X'\in \bX'$ we 
% let $Y'(X')\in\bY'$ be its corresponding label in $\D'$ and 
define its set of neighbors in $\bX'$ by
% $\NNm(X',\bX')$,
%\textcolor{blue}{*Use interior of $V$*}
\begin{align}
\label{def:NNm}
&\NNm(X';\bX') =
%\\
%&\{Q'\in \bX' : \exists x\in V_{X'}(\bX')	\nonumber
%\text{ s.t. } Q' = X_{(1)}(x, \bX'\setminust\{X'\}) \}\nonumber
\\&\nonumber
\quad
\big\{Q'\in \bX' : \exists x\in V_{X'}(\bX')	
\text{\,\, s.t.\,\, } Q' = \Xinn{2}(x; \bX') \big\},
\end{align}
where $\Xinn{2}(x; \bX')$ denotes the second nearest neighbor of $x$ in $\bX'$.
Given an oracle to 
$\NNm$,
%$\NNm(\cdot;\cdot)$, 
consider the following iterative algorithm for removing spurious prototypes from $\D'$. %\footnote{\textcolor{violet}{In section \ref{sec:problem setup} we let $\tD$ the sample data that is compressed from $\D$ by $\variant{}$. In this section we let $\tD$ the sample data that is (further) compressed from $\D'$ by \compalg.}}
Initialize $\tD = \D'$ and iterate over the elements in $\tD = (\tbX,\tbY)$.
At any stage, if $\tD$ is a singleton, 
%return $\tD$,
exit the loop. 
Else, for
$(X',Y') \in \tD$, if 
$Y'(Q') = Y'$ 
for all $Q'\in \NNm(X';\tbX)$, set $\tD \leftarrow \tD \setminus \left\{\left(X',Y'\right)\right\}$.

By the definition of $\NNm$, it is clear that at any stage of the iterative algorithm,
%\todo{holds only $\lambda$-almost every $x$?}
%for all $\in\X$, 
$$\Yinn{1}(x; \tD)= \Yinn{1}(x; \D')
,\qquad \forall x\in\X
.
$$
Therefore, the classifier based on $\tD$ is identical
%\todo{up to tie breaking on a set of measure zero}
 to the one based on $\D'$ and they have the same error.
% as established in Theorem \ref{}.
%\todo{Omer: Not sure if it's necessary, but maybe change the notation from $\tD$ to $\widehat \D$ because in Theorem \ref{thm:comp_agree Lebesgue} we redefine $\tD$, and we don't explicitly claim that they are equal.}

%While given an oracle to $\NNm$, 
While the above iterative compression procedure can in principal be applied in any metric space,
computing $\NNm$ might be infeasible in some cases.
For example, when the metric space has no vector-space structure, determining the content and boundary of a Voronoi cell may require brute force computation.
%In Section \ref{sec:simulations} we propose and empirically study a simple natural approximation for $\NNm$ that uses the samples in $\D_n$ to efficiently approximate $\NNm(X',\bX')$ in any metric space.
%In addition,
%Unfortunately, 
In addition, analyzing its compression rate 
%of this algorithm 
is challenging since removing a prototype may change the Voronoi partition considerably.
Nevertheless, when the instance space is the Euclidean one, 
things become more tractable.
%$\X$  $(\R^d, \lVert\cdot\rVert_2)$ things  are better.
%\todo{Omer: Verify change.}
As we establish in the following theorem,
% and the discussion that follows, 
%\textcolor{violet}{\sout{under the additional assumptions}}
%\todo{What assumptions are needed exactly? Omer: done.}
in $(\R^d, \lVert\cdot\rVert_2)$ one can remove 
%if 
all spurious prototypes in $\D'$ 
%are 
%can be
%removed 
\emph{simultaneously}, 
resulting in an identical classifier for $\lambda$-almost all $x\in\X$ (where $\lambda$ is the Lebesgue measure).

%The proof is given in Section \ref{sec:comp_ident}.
%then, \textcolor{violet}{with probability one over the drawn of $\D'$, the label of $Y'_{(1)}(x,\cdot)$ remains the same for $\lambda$-almost all $x\in\X$.}

\begin{theorem}
%\todoi{Maybe can improve the wording of the Lemma.}
\label{thm:comp_agree Lebesgue}
Let $\X=\R^d$ be equipped with the Euclidean metric $\rho$.
Assume the distribution $\mu$ of $X$ has a density with respect to $\lambda$ and
let 
$\D=(\bX,\bY)$
%$\D'=(\bX',\bY')\subseteq\X\times\Y$
be a finite labeled sample where the samples in $\bX$ are independently drawn according to $\mu$.
%consisting of at least two instance-label pairs with different labels, and let
Let $\D'=(\bX',\bY')$ be any subset of $\D$.
In the case that $\D'$ consists of at least two instance-label pairs with different labels, let
%\begin{align}
%\tD=\D'\,\setminus\,&\big\{(X',Y'_{X'})\in \D':	
%\label{def:tD} \\ 
%&\forall Q'\in \NNm_{X'}(\bX'), Y'_{X'}=Y'_{Q'}\big\}
%,
%\nonumber
%\end{align}
\begin{align}
\tD=\D'\,\setminus\,&\big\{(X',Y')\in \D':	
\label{def:tD} 
\\ &\qquad \nonumber
\forall Q'\in 
%\NNm_{X'}(\bX')
\NNm(X';\bX')
, Y'(Q')=Y'\big\}
,
\end{align}
and else, let $\tD=\{(X'_1,Y'_1)\}$.
Then, with probability one over $\D$,
for $\lambda$-almost all $x\in\X$,
\begin{align}
\label{equal labels}
\Yinn{1}(x;\tD) = \Yinn{1}(x;\D')
%,\qquad \text{for $\lambda$-almost all }x\in\X
.
\end{align}
\end{theorem}
%\paragraph{Computational considerations.}
%\textcolor{violet}{[In response to the reviewer comment \#8:] In the supplementary material [ADD REFERENCE] we discuss how Theorem \ref{thm:comp_agree Lebesgue} may fail in the non-Euclidean space.}
As for feasibility, in principal, 
$\NNm(\cdot;\bX')$
%\textcolor{violet}{the Voronoi partition}
can be computed for all prototypes in $\bX'$ simultaneously by computing the corresponding Voronoi diagram.
Several algorithms have been proposed for this task for the  Euclidean space and other well-behaved metric spaces, including, for example, the gift-wrapping algorithm, Seidel's shelling algorithm, and a careful application of the simplex method for linear programming; see \citet{dwyer1991higher},  \citet{fortune1995voronoi} and references therein.
%  and a variant of the gift-wrapping algorithm \citep{dwyer1991higher}.

In the worst case, the Voronoi diagram can have up to 
$n^{\lfloor{\frac{d}{2}}\rfloor}$ cells \citep{chazelle1993optimal}, leading to impractical runtime.
Those cases however are degenerate and correspond to the case where $\bX'$ is not in general position (for example, when some $d+2$ points in $\bX'$ all lie on the surface of some ball).
When $\mu$ has a density satisfying the 
SDC,
%strong density condition, 
%the number of cells is $\Theta(n)$ with high probability,
$\bX'$ is in general position with high probability \citep{dwyer1991higher}.
%This is reflected 
In that case, 
%$\NNm$ can be computed from the Delaunay triangulation of $\bX_m'$ and 
the iterative Watson-Bowyer algorithm \citep{watson1981computing,bowyer1981computing} for computing the dual of the Voronoi diagram (a.k.a.\ the Delaunay triangulation) recovers $\NNm(\cdot;\bX')$ in time $O(|\bX'|^2)$.
A variant of the gift-wrapping algorithm has been shown to have expected runtime of $\Theta(|\bX'|)$ when $\bX'$ drawn uniformly from $\Ball{\boldsymbol 0}{1}$  \citep{dwyer1991higher}.
%In particular, the number of neighboring cells of typical cell depends only on $d$.

While the above algorithms for computing $\NNm$ give the exact set of neighboring cells, they are complex to implement and do not readily generalize to general metric spaces.
We thus also consider a natural approximation for $\NNm$
%AK
%which
that
uses the instances in $\D_n$ to efficiently approximate $\NNm$.
The heuristic,
termed \textbf{\compheu{}},
 designates a prototype $Q'\in\bX'$ as a neighbor of $X'\in\bX'$ if $Q'$ is the second nearest neighbor of any \emph{sample} from $\bX_n$ that fell into $X'$'s cell; formally,
\begin{align}
\label{eq:comp_heuristic}
& \widetilde\NNm(X';\bX',\bX_n)  = 
\\&\quad\nonumber
\big\{Q'\in \bX' : 
%\\ & 
\exists X\in\bX_n \cap V_{X'}(\bX'), 
%\text{ s.t. } 
Q' = \Xinn{2}(X; \bX') \big\}.
\end{align}
Note that $\widetilde\NNm$ can be applied in any metric space and can be used in the iterative algorithm above. Its runtime is $O(|\bX'| |\bX_n|)$ for a single query $\widetilde\NNm(X';\bX',\bX_n)$.
Pseudocode for \compalg{} and \compheu{} is given in the supplementary material (Procedure \ref{alg:protocomp}). 

\subsection{Further compression applied to 
\optinet{}
%\variant{}
}
%\subsection{Further Compression \textcolor{violet}{Study \sout{ Applied to \textcolor{violet}{\variant{}\ \sout{\optinet{}}}}}}
%\paragraph{Further compression for \optinet{}.}
%%\todo{The added sentence is a response to entry \#1 reviewer comment.} The compression scheme of Theorem \ref{thm:comp_agree Lebesgue} can be naively applied to any prototype-based rule and yield different results.
%For example, applying it to the $k$-NN rule may obviously lead to unsatisfying results. Applying it to the 1-NN rule, to \variant{}, or to \optinet, on the other hand, does not significantly change the classifier, as shown in Theorem \ref{thm:comp_agree Lebesgue}.
We now apply the compression scheme of Theorem \ref{thm:comp_agree Lebesgue} to 
\optinet{}
%\variant{} 
in \eqref{gdec}.
Unfortunately, so far we were unable to establish compression rates under the probabilistic Tsybakov margin condition \eqref{eq:Tsybakov}.
%rate achieved by the compression scheme in Theorem \ref{} as applied to OptiNet, 
Instead, we derive compression rates under a stronger geometric condition that bounds the noise far from the decision boundary.
This condition has been introduced by \citet{blaschzyk2018improved} for the binary classification setting, where it was shown that in conjunction with an additional margin condition that we do not consider here, slightly faster minimax error rates are achieved. 
%\citep{blaschzyk2018improved}. 
Here we study it in the context of sample compression in the multiclass setting.
%This condition was first
%\todo{check}
%introduced by \citet{blaschzyk2018improved} for the binary case and here we consider its multiclass generalization.

Recall the multiclass noise margin $\eta(x)=P_{(1)}(x)-P_{(2)}(x)$ in \eqref{eq:reg}.
Let $\dtb:\X\to\R^+$ be the distance function from the decision boundary,
\begin{align}
\delta(x)=\inf\limits_{x'\in\X:\ \reg(x')=0}\rho(x,x').
\label{def:delta}
\end{align}
For any $t\geq0$, define the \emph{$t$-envelope} around the decision boundary by 
%$\boundary{t} &=\{x\in\X:\delta(x)\leq t\}.$
\begin{align*}
\boundary{t} &=\{x\in\X:\delta(x)\leq t\}.
\end{align*}

%\todoi{\sout{Use the $\eta$ definition (eq. \eqref{eq:reg}) and also the $\delta$ definition.}}

\begin{definition}
\label{def:geom_margin}
%Let $\eta$ be as in \eqref{eq:reg}.
The geometric margin condition \textup{(\textbf{GMC})} means that there exist $\gm\geq0$ and $c_1>0$ such that for $\mu$-almost all $x\in\X$,
%$\eta(x)\geq \min\left\{c_1\dtb(x)^\gm,1\right\}$.
\begin{align*}
%\label{cond:Blas}
\eta(x)\geq 
%\delta(x)^\gm/c_1
\min\left\{c_1\dtb(x)^\gm,1\right\}
.
\end{align*}
%\todoi{Don't we need $\min\{\delta(x)^\gm/c_1,1\}$ on the RHS?
%\\
%On the one hand Blaschzyk presented this property without the minimum.
%On the other hand, taking the minimum doesn't cost too much, and it enables to include more distributions. For example: The distribution of $\X=\R$, $P_1(x)=1$ for all $x\geq 1$, $P_2(x)=1$, for all $x\leq -1$, and in $(-1,1)$ we set $P_1,P_2$ to be H\"older.
%}
%\todo{IMPORTANT: $\gm$ and $\beta$ are not independent. Need to figure out how they restrict each other. Need to see what \citet{blaschzyk2018improved} did.}
\end{definition}

Smaller $\gm$ means a sharper decision boundary.
Note that \textup{\textbf{GMC}} implies $\beta\leq\gm$.

Now let $\bXg(\gamma)$ be a $\gamma$-net of $\bX'_m$ and let $\bYg(\gamma)$ be the corresponding labels as computed by \optinet{},
%\variant{},
%Algorithm \ref{???},
%\todo{ref}
stacked into $\D(\gamma) = (\bXg(\gamma),\bYg(\gamma))$.
%For $X' \in \bXg(\gamma)$ abbreviate 
%$$\NNma(X') = \NNm(X';\bXg(\gamma)).$$
By Theorem \ref{thm:comp_agree Lebesgue}, the further-compressed dataset of $\Dg(\gamma)$,
\begin{align}
\label{eq:comp_set}
\tD(\gamma) = \D(\gamma) \setminus \big\{(&X',Y')\in\D(\gamma):
\\&\nonumber  \quad
\forall Q' \in 
\NNm(X';\bXg(\gamma))
%\NNma(X')
, Y'(Q') = Y'\big\},
\end{align}
induces the same classifier as $\D(\gamma)$ for $\lambda$-almost all $x$, and so has the same error. 

\begin{theorem}
\label{thm:comp_rate}
Assume the marginal distribution $\mu$ of $X$ has a density $f$ that satisfies the strong density condition
%{minimal mass condition} and is bounded away from zero 
\textup{(\textbf{SDC})} with $\gamma_0>0$ and that the regression functions $(P_j)_{j=1}^M$ are continuous.
%and that $\supp(\mu)$ is a convex set.
%\todo{Where in the proof did you use the convexity of the support?}
%the tail condition.
%$f$ is bounded away from zero.
%the distribution function $H_x(\cdot)$ is continuous for each $x$.
If 
%the margin condition is satisfied with $0<\alpha \leq 1$, 
the geometric margin condition \textup{(\textbf{GMC})} is satisfied with $\gm\geq0$,
%and the H\"older continuity condition is met with $0<\beta\leq 1$,
then there are $c,C,C'>0$ such that for any $0<\gamma\leq\gamma_0$ and $0<t\leq c_1^{-\xi}$,
the further-compressed dataset $\tD(\gamma)$ in \eqref{eq:comp_set} satisfies
\begin{align}
\label{eq:deco_thm}
\EXP\{|\tD(\gamma)|\}
&\leq
\CN_\gamma(\boundary{t+c\gamma})
\\ \nonumber & \quad 
+ \CN_\gamma(\boundary{t+c\gamma}) 
%\cdot
\CN_\gamma(\supp(\mu)) 
%\cdot 
e^{-C m \gamma^d}
%\\&\quad
\\ \nonumber & \quad 
+ C'
%\cdot
\CN_\gamma(\supp(\mu))^2 
%\cdot 
e^{- Cn  t^{2\gm}\gamma^d}
.
\end{align}
%\begin{align*}
%\EXP\left[|\tD(\gamma)|\right]
%&\leq
%\CN_\gamma(\boundary{t+c\gamma})
%\left( 1 + \CN_\gamma(\supp(\mu)) e^{-C m \gamma^d} 
%\right)
%\\&\quad
% + \CN_\gamma(\supp(\mu))^2 e^{- Cn  t^{2\gm}\gamma^d}
%.
%\end{align*}
%for any $0<t\leq 1$.
\end{theorem}

To interpret the result in Theorem \ref{thm:comp_rate}, consider for example the case $\gm=\beta=1$, which are compatible with $\alpha=1$ (a concrete example is given in the suplemantery material).
%, \ref{sec:simulations}). 
%\todo{}
First note that setting $m=m_n$ as it was set to obtain optimal error rates
in \eqref{eq:gamma_n and m_n}, and using the bound 
%Recall that under the SDC, the maximal size of a $\gamma$-net of $\supp(\mu)$ satisfies 
$\CN_\gamma(\supp(\mu))\leq(C/\gamma)^d$ 
%on $\CN_\gamma$ 
in \eqref{eq:cn_bound}, the second term in \eqref{eq:deco_thm} is $\CN_\gamma(\boundary{t+c\gamma}) \cdot O(1)$.
Setting 
%$t=t_n=\gamma_n^{1-\varepsilon_n}$ with $\varepsilon_n= \tfrac{(d+2)\log\log (n^{\frac{d+1}{C(2+d)}})}{2\log n} \xrightarrow[n\to\infty]{}0$
\begin{align*}
t&=t_n=\gamma_n^{1-\varepsilon_n}
\end{align*}
with
\begin{align*}
%\qquad\text{with}\qquad
%\\
\varepsilon_n &= \tfrac{(d+2)\log\log n^{\frac{d+1}{C(2+d)}}}{2\log n}
\xrightarrow[n\to\infty]{}0,
\end{align*}
and $\gamma_n=n^{-\frac{1}{2+d}}$ as in \eqref{eq:gamma_n and m_n},
the third term in \eqref{eq:deco_thm} satisfies
\begin{align*}
%&
 C'\CN_{\gamma_n}(\supp(\mu))^2 e^{- Cn  t_n^{2}\gamma_n^d} 
%\\
&\leq
C' \gamma_n^{-2d} e^{- Cn \gamma_n^{2+d-2\varepsilon_n}}
%\\& = 
%C' \gamma_n^{-2d} e^{- C\gamma_n^{-2\varepsilon_n}}
%\\& 
%= 
%C' n^{\frac{2d}{2+d}} e^{- C n^{-\frac{2\varepsilon_n}{2+d}}}
\\& 
= C' n^{\frac{d-1}{2+d}} 
\\&
= O(\gamma_n^{-(d-1)}).
\end{align*}
Lastly, the first term in \eqref{eq:deco_thm} corresponds to the size of a $\gamma_n$-net over a $(t_n+c\gamma_n)$-envelope of the decision boundary.
Assuming the decision boundary is a smooth manifold $\surf$ of dimension $d-1$ (such as the surface of a ball),
%When the decision boundary corresponds to a smooth manifold of , 
one expects that
%\todo{Fix ???}
\begin{align*}
\CN_{\gamma_n}(\boundary{t_n+c\gamma_n})
& \approx {\CN_{\gamma_n}(\surf)}
\cdot\frac{(t_n+c\gamma_n)}{\gamma_n} 
\\&
= \CN_{\gamma_n}(\surf)\cdot O(\gamma_n^{-\varepsilon_n})
\\&
= O(\gamma_n^{-(d-1)-\varepsilon_n}).
\end{align*}
%\todo{Omer: Why is the approximation term true?}
Putting all terms together,
$$\EXP\{|\tD(\gamma_n)|\}  = O(\gamma_n^{-(d-1)-\varepsilon_n}) = O(n^{\frac{d-1}{2+d}}\log n),$$
%\begin{align*}
%\EXP\{|\tD(\gamma_n)|\}  = O(\gamma_n^{-(d-1)-\varepsilon_n}) = O(n^{\frac{d-1}{2+d}}\log n),
%\end{align*}
leading to compression rate 
%$\EXP\{|\tD(\gamma_n)|\}/n$ 
of order $n^{-\frac{3}{2+d}}\log n$.
Hence, a factor of order $n^{-\frac{1}{2+d}}\log n = \widetilde O(\gamma_n)$ is gained in the compression rate by further using 
\compalg{}
%the compression scheme 
of Theorem \ref{thm:comp_agree Lebesgue} as compared to the 
%compression 
rate $n^{-\frac{2}{2+d}}$ in \eqref{eq:cn_bound} obtained for \optinet{}.
% \variant{}.
%\todo{Omer: Why is it true?}
This holds while still enjoying near-minimax optimal error rate.

\section{Experimental study}

\begin{figure}[t]
\centering
\includegraphics[width=0.75\columnwidth, 
%height=0.22\textheight
]{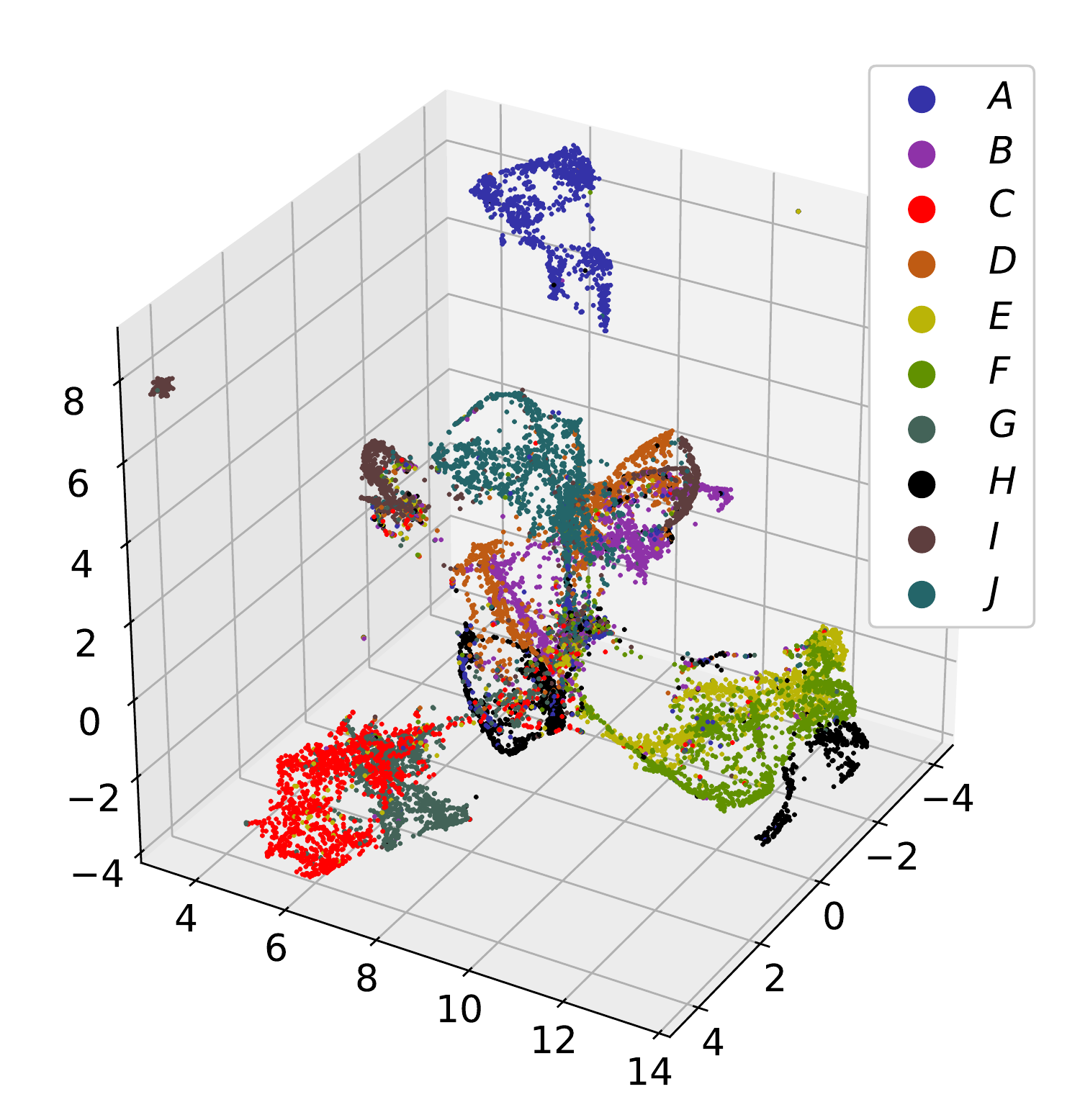}
\includegraphics[width=0.75\columnwidth, 
%height=0.22\textheight
]{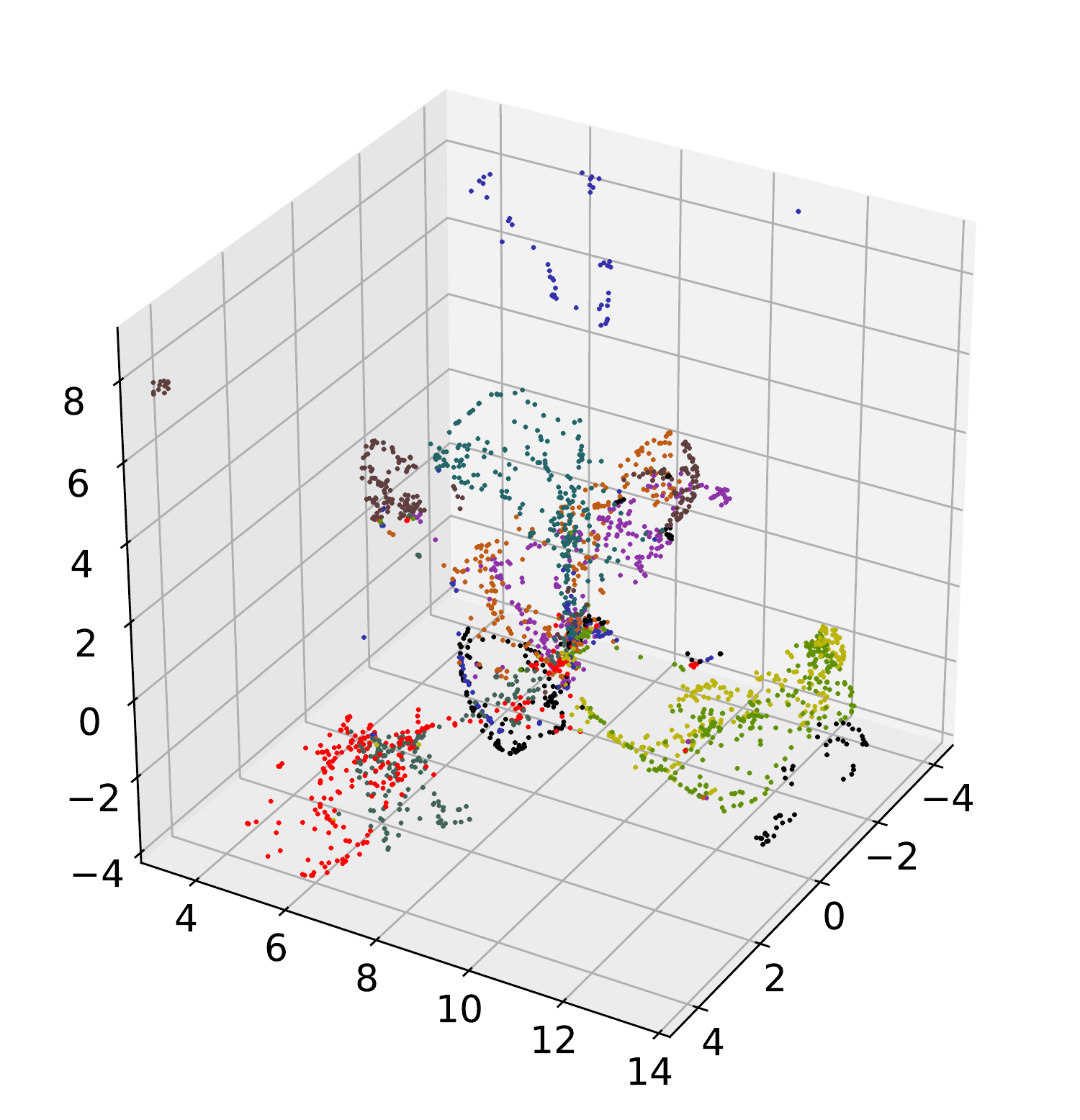}
\caption{The embedding of notMNIST by UMAP (top) and the prototype set as computed by \hybrid{} + \compalg{} (bottom).}
\label{fig:scatter}
\end{figure}

%\section{ SIMULATION RESULTS}
\label{sec:simulations}
We demonstrate the performance of the various algorithms discussed in this paper on the notMNIST dataset (Yaroslav Bulatov 2011), consisting of $\approx 19$k different font glyphs of the letters A-J (10 classes), each of dimension $28\times 28$.
To facilitate the experiments on a standard computer, we first applied Uniform Manifold Approximation and Projection dimensionality reduction (UMAP) of \citet{mcinnes2018umap}, reducing the dimension from $28\times28$ to $d=3$. The resulting embedding is shown in Figure \ref{fig:scatter} (top).
%Additional simulation results, where \compheu{} is also applied, are given in the supplementary material.

%\begin{figure}%
%\centering
%\hspace{5pt}
%%$\vcenter{\hbox{
%\includegraphics[width=0.35\columnwidth]{scatter}
%%}}$%
%\hspace{50pt}
%%$\vcenter{\hbox{
%\includegraphics[width=0.35\columnwidth]{legends}
%%}}$%
%\caption{The data embedding by UMAP (left) and the prototype learning rules considered (right).}%
%\label{fig:scatter}%
%\end{figure}

\begin{figure}[t]
\centering
\includegraphics[width=0.85\columnwidth, height=0.24\textheight]{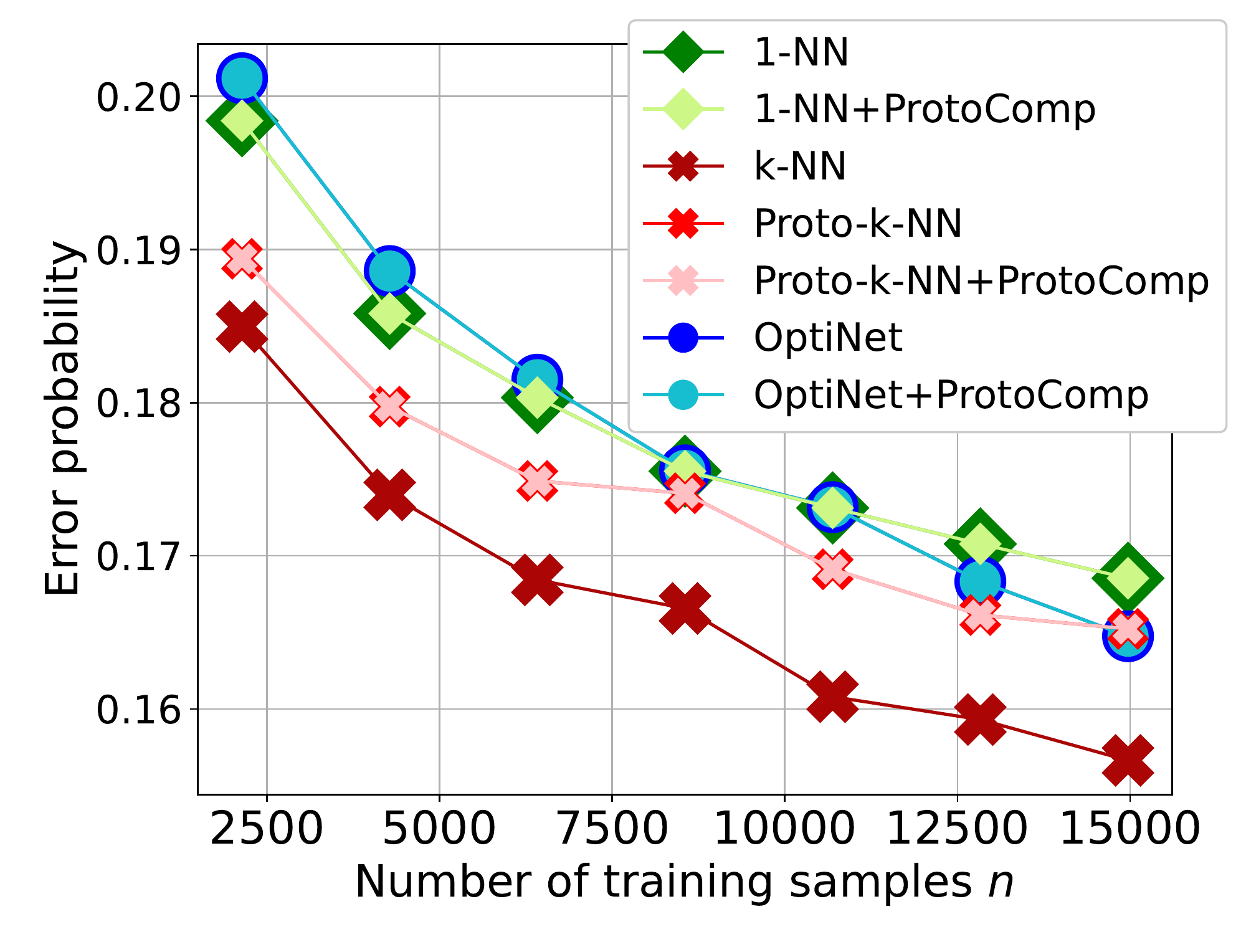}
\includegraphics[width=0.85\columnwidth, height=0.24\textheight]{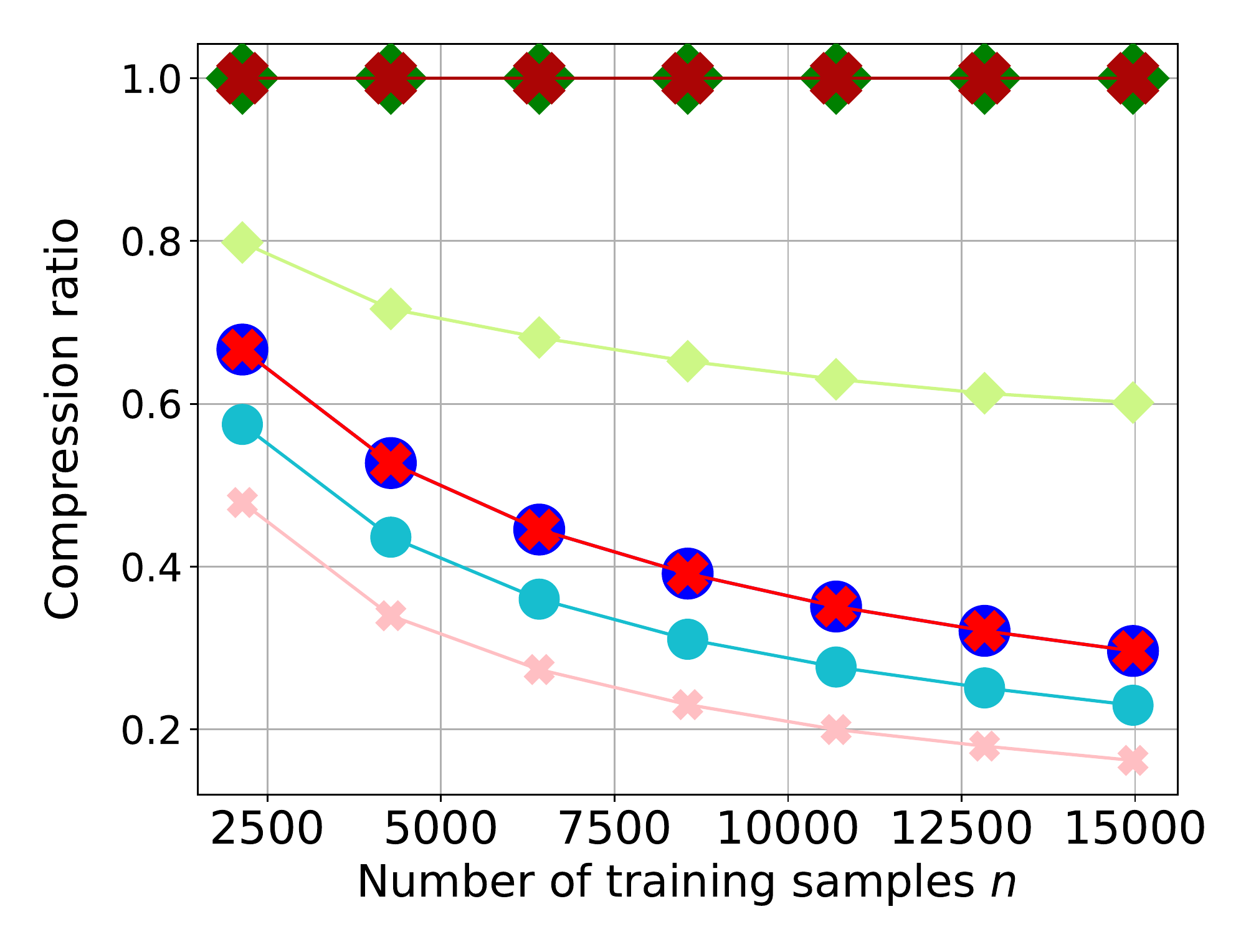}
\caption{Error and compression rates for the prototype rules.}
\label{fig:rates}
\end{figure}

We consider the methods listed in Figure \ref{fig:rates} (top).
%\todo{Remove titles and enlarge the font sizes of x and y axes labels and ticks and the legend. They should be approximately the size of the font in main text. Make grid finer.}
The dataset was split into training ($80\%$) and testing ($20\%$) sets.
%We randomly splited the dataset into $80\%$ training and $20\%$ for testing.
In Figure\ \ref{fig:rates} (top) we show 
the error obtained on the test set, over 5 realizations of the random splitting.
%on a logarithmic scale the excess errors (top) as estimated on a large test dataset, and 
The compression ratios achieved are shown on the bottom. 
The runtime for construction and evaluation are given in Figure \ref{fig:runtime} in the supplementary material.
%as a function of sample size 
%for $d=2$.
%\todo{fill}
The 
parameters 
$\gamma=0.11$ for \optinet{} and $k=10$ for $k$-NN were chosen by a validation procedure. The same $k$ is used for \hybrid{}, while its compression sizes $m_n$ were \emph{matched} to that of \optinet{}.
%the same of $m_n=n/k_n$.
%The same $m_n$s were used by \protonn{}.
%\todo{rephrase}
% so they use the same compression size.
%
%In Fig.\ \ref{fig:point_set} we show an example of the prototypes retained by \optinet{} (left) and those retained after further applying \compalg{} to it (right).

%for the methods were computed as follows.
%For \optinet{}, $\gamma_n$ for $n=10,\!000$ was first determined by a validation procedure on a holdout dataset. For other values of $n$, $\gamma_n$ was determined by interpolating according to the theoretical scaling in \eqref{eq:gamma_n and m_n} (here \optinet{} computes the $\gamma_n$-net directly over $\bX_n$).
%The value of $k_n$ for the $k$-NN rule was determined similarly.
%For \hybrid{}, $k_n$ was set as for $k_n$-NN and
%% $m_n$ to 
%$m_n=cn/k_n$ for $c=???$
%\todo{fill}
%tuned manually. This $m_n$ was then used for \protonn{}.

%\todo{In this paragraph we use write ``OptiNet's" as in the ``\_\_\_ of OptiNet". Maybe rephrase them because it is like the name ``OptiNet'".}
The results highlight the following:
(i) $k$-NN achieves the smallest error, but does not compress the data. $1$-NN's error lags behind and further compressing its prototype set (the latter being the whole training dataset) using \compalg{} gives non-trivial compression rates without changing the error;
(ii) \optinet{} and \hybrid{} achieves slightly better error than $1$-NN, but not as good as $k$-NN.
Further compressing \optinet{} and \hybrid{} using \compalg{} gives highly compressed prototype sets without changing the errors, with an advantage to \hybrid{}. 
The final prototype set of \compalg{} as applied on \hybrid{}  is shown in the bottom of Figure \ref{fig:scatter}.

\section{Conclusion}
\label{sec:conclusions}
%\section{CONCLUSION}
%
%\begin{remark}
%Note that in general metric spaces, $X'\in\NNm(Q',\bX')$ does not necessarily imply that $Q'\in\NNm(X',\bX')$. For example, consider the case where $\X=(0,3)\uplus(6,8)$ and $\bX'=\{X_1=1, X_2=2, X_3=7\}$. In that case $\NNm(X_3,\bX')=\{X_2\}$ but $\NNm(X_2,\bX')=\{X_1\}\not\ni X_3$.
%\todo{A sufficient condition is...}
%\end{remark}

%We believe our techniques for establishing compression rates can also be used to establish such rates for \protonn{} and the more advanced  learning rules mentioned above.
%The extension of our results to more general metric instance spaces is also of interest.
%In  particular, in doubling metric spaces, \citet{GoKoNi14} proposed a fast hierarchical compression heuristic for further compressing the sample without altering the resulting classifier. However so far no compression guarantees have been established for this rule.
%We leave these and other related problems for future research.

%Discuss open problems:
%\\ - Aryeh's paper on average margin

%A key aspect
%%as demonstrated by the inconsistency of 1-NN and the optimal rates of compressed prototype rules, 
%demonstrated in all of the above works is that 
%%in the non-parametric setting, 
%compression serves a dual purpose: it reduces computational complexity, and further \emph{improves} accuracy by acting as a regularizer so as to avoid statistical overfitting. 
%We believe that better understanding this interplay in the non-parametric setting in terms of prototype rules is of much interest.

In this paper we study jointly-achievable error and compression rates  for 
\optinet{}
%\variant{} 
in a common non-parametric classification setting.
% and other prototype learning rules.
% under the non-parametric setting.
%study several aspect of the error-compression interplay for .
We believe our techniques 
%for establishing compression rates 
can be extended to derive such rates for 
\protonn{}, \hybrid{},
and the more advanced adaptive rules mentioned in the Introduction,
% \ref{chapter:intro}, 
as well as for the fast hierarchical compression heuristic proposed by \citet{GoKoNi14}. 
The latter is particularly important, since
%the statistical performance and 
the computational feasibility of the prototype rules studied here rapidly deteriorates as the dimension of the instance space increases.
Studying compression rates in terms of the \emph{average margin} of \citet{ashlagi2021functions},  and extending the results to metric losses \citep{cohen2022learning}, are also compelling.

%The latter two are in particular important, since
%the statistical performance and the computational feasibility of the prototype rules studied here rapidly deteriorate as the dimension of the instance space increases. As such, our study is more of theoretical interest than practical. 

More fundamentally, given the universal consistency of \optinet{} and \protonn{} in any separable metric space,
the extension of our results beyond the Euclidean space is of interest.
In particular, Theorem \ref{thm:comp_agree Lebesgue} shows that in $(\R^d,{\lVert\cdot\rVert}_2)$ one can remove all spurious prototypes simultaneously, essentially without altering the classifier.
%to yield a classifier that is consistent with the original classifier for $\lambda$-almost all $x$.
We conjecture that this
%Theorem \ref{thm:comp_agree Lebesgue} 
holds also 
%when $\rho$ is the 
for the $\ell_p$-norm for any $p\in(1,\infty)$.
However, in more general metric spaces, Theorem \ref{thm:comp_agree Lebesgue} can fail, in the sense that removing all spurious prototypes simultaneously might lead to a classifier that is not consistent with the original one;
see the supplementary material for a concrete example.
%\todo{Add the example to the supp material}
In that case, one can use the iterative version of the compression (see the Further Compression section)
%(Section \ref{sec:further_compression}) 
while computing the neighboring cells using the heuristic \compheu{} in \eqref{eq:comp_heuristic}.
% \compheu{} discussed at the end of Section \ref{sec:}, 
However, the theoretical properties of this lossy compression scheme are currently unknown.
We leave these and related problems to future research.

\bibliography{refs}

\begin{thebibliography}{50}
\providecommand{\natexlab}[1]{#1}

\bibitem[{Alon et~al.(2021)Alon, Hanneke, Holzman, and Moran}]{alon2021theory}
Alon, N.; Hanneke, S.; Holzman, R.; and Moran, S. 2021.
\newblock A theory of {PAC} learnability of partial concept classes.
\newblock \emph{arXiv preprint arXiv:2107.08444}.

\bibitem[{Ariew(1976)}]{ariew1976ockham}
Ariew, R. 1976.
\newblock Ockham's Razor: A historical and philosophical analysis of Ockham's
  principle of parsimony, University of Illinois, Champaign-Urbana.

\bibitem[{Ashlagi, Gottlieb, and Kontorovich(2021)}]{ashlagi2021functions}
Ashlagi, Y.; Gottlieb, L.-A.; and Kontorovich, A. 2021.
\newblock Functions with average smoothness: structure, algorithms, and
  learning.
\newblock In \emph{Conference on Learning Theory}, 186--236. PMLR.

\bibitem[{Audibert and Tsybakov(2007)}]{AuTs05}
Audibert, J.-Y.; and Tsybakov, A.~B. 2007.
\newblock Fast learning rates for plug-in classifiers.
\newblock \emph{The Annals of statistics}, 35(2): 608--633.

\bibitem[{Biau and Devroye(2010)}]{biau2010layered}
Biau, G.; and Devroye, L. 2010.
\newblock On the layered nearest neighbour estimate, the bagged nearest
  neighbour estimate and the random forest method in regression and
  classification.
\newblock \emph{Journal of Multivariate Analysis}, 101(10): 2499--2518.

\bibitem[{Biau, Devroye, and Lugosi(2008)}]{biau2008consistency}
Biau, G.; Devroye, L.; and Lugosi, G. 2008.
\newblock Consistency of random forests and other averaging classifiers.
\newblock \emph{Journal of Machine Learning Research}, 9(9).

\bibitem[{Binev et~al.(2014)Binev, Cohen, Dahmen, and
  DeVore}]{binev2014classification}
Binev, P.; Cohen, A.; Dahmen, W.; and DeVore, R. 2014.
\newblock Classification algorithms using adaptive partitioning.
\newblock \emph{The Annals of Statistics}, 42(6): 2141--2163.

\bibitem[{Blanchard et~al.(2007)Blanchard, Sch{\"a}fer, Rozenholc, and
  M{\"u}ller}]{blanchard2007optimal}
Blanchard, G.; Sch{\"a}fer, C.; Rozenholc, Y.; and M{\"u}ller, K.-R. 2007.
\newblock Optimal dyadic decision trees.
\newblock \emph{Machine Learning}, 66(2-3): 209--241.

\bibitem[{Blaschzyk and Steinwart(2018)}]{blaschzyk2018improved}
Blaschzyk, I.; and Steinwart, I. 2018.
\newblock Improved classification rates under refined margin conditions.
\newblock \emph{Electronic Journal of Statistics}, 12(1): 793--823.

\bibitem[{Bousquet et~al.(2020)Bousquet, Hanneke, Moran, and
  Zhivotovskiy}]{bousquet2020proper}
Bousquet, O.; Hanneke, S.; Moran, S.; and Zhivotovskiy, N. 2020.
\newblock Proper learning, Helly number, and an optimal SVM bound.
\newblock In \emph{Conference on Learning Theory}, 582--609. PMLR.

\bibitem[{Bowyer(1981)}]{bowyer1981computing}
Bowyer, A. 1981.
\newblock Computing dirichlet tessellations.
\newblock \emph{The computer journal}, 24(2): 162--166.

\bibitem[{C{\'e}rou and Guyader(2006)}]{cerou2006nearest}
C{\'e}rou, F.; and Guyader, A. 2006.
\newblock Nearest neighbor classification in infinite dimension.
\newblock \emph{ESAIM: Probability and Statistics}, 10: 340--355.

\bibitem[{Chaudhuri and Dasgupta(2014)}]{ChDa14}
Chaudhuri, K.; and Dasgupta, S. 2014.
\newblock Rates of convergence for nearest neighbor classification.
\newblock In \emph{Advances in Neural Information Processing Systems},
  3437--3445.

\bibitem[{Chazelle(1993)}]{chazelle1993optimal}
Chazelle, B. 1993.
\newblock An optimal convex hull algorithm in any fixed dimension.
\newblock \emph{Discrete \& Computational Geometry}, 10(4): 377--409.

\bibitem[{Chitnis(2022)}]{chitnis2022refined}
Chitnis, R. 2022.
\newblock Refined Lower Bounds for Nearest Neighbor Condensation.
\newblock In \emph{International Conference on Algorithmic Learning Theory},
  262--281. PMLR.

\bibitem[{Cohen and Kontorovich(2022)}]{cohen2022learning}
Cohen, D.~T.; and Kontorovich, A. 2022.
\newblock Learning with metric losses.
\newblock In \emph{Conference on Learning Theory}, 662--700. PMLR.

\bibitem[{Cover(1999)}]{cover1999elements}
Cover, T.~M. 1999.
\newblock \emph{Elements of information theory}.
\newblock John Wiley \& Sons.

\bibitem[{David, Moran, and Yehudayoff(2016)}]{david2016supervised}
David, O.; Moran, S.; and Yehudayoff, A. 2016.
\newblock Supervised learning through the lens of compression.
\newblock \emph{Advances in Neural Information Processing Systems}, 29:
  2784--2792.

\bibitem[{Devroye and Gy{\"o}rfi(1985)}]{DeGy85}
Devroye, L.; and Gy{\"o}rfi, L. 1985.
\newblock \emph{Nonparametric density estimation: the $L{_{1}}$ view}.
\newblock Wiley Series in Probability and Mathematical Statistics: Tracts on
  Probability and Statistics. John Wiley \& Sons, Inc., New York.
\newblock ISBN 0-471-81646-9.

\bibitem[{Devroye, Gy{\"o}rfi, and Lugosi(1996)}]{DeGyLu96}
Devroye, L.; Gy{\"o}rfi, L.; and Lugosi, G. 1996.
\newblock \emph{A probabilistic theory of pattern recognition}.
\newblock Springer-Verlag New York, Inc.

\bibitem[{D{\"o}ring, Gy{\"o}rfi, and Walk(2017)}]{doring2017rate}
D{\"o}ring, M.; Gy{\"o}rfi, L.; and Walk, H. 2017.
\newblock Rate of convergence of k-nearest-neighbor classification rule.
\newblock \emph{The Journal of Machine Learning Research}, 18(1): 8485--8500.

\bibitem[{Dwyer(1991)}]{dwyer1991higher}
Dwyer, R.~A. 1991.
\newblock Higher-dimensional Voronoi diagrams in linear expected time.
\newblock \emph{Discrete \& Computational Geometry}, 6(3): 343--367.

\bibitem[{Floyd and Warmuth(1995)}]{floyd1995sample}
Floyd, S.; and Warmuth, M. 1995.
\newblock Sample compression, learnability, and the {V}apnik-{C}hervonenkis
  dimension.
\newblock \emph{Machine learning}, 21(3): 269--304.

\bibitem[{Fortune(1995)}]{fortune1995voronoi}
Fortune, S. 1995.
\newblock Voronoi diagrams and Delaunay triangulations.
\newblock \emph{Computing in Euclidean geometry}, 225--265.

\bibitem[{Gadat, Klein, and Marteau(2016)}]{GaKlMa16}
Gadat, S.; Klein, T.; and Marteau, C. 2016.
\newblock Classification in general finite dimensional spaces with the
  $k$-nearest neighbor rule.
\newblock \emph{Ann. Statist.}, 44(3): 982--1009.

\bibitem[{Gottlieb, Kontorovich, and Nisnevitch(2014)}]{GoKoNi14}
Gottlieb, L.-A.; Kontorovich, A.; and Nisnevitch, P. 2014.
\newblock Near-optimal sample compression for nearest neighbors.
\newblock In \emph{Neural Information Processing Systems (NIPS)}.

\bibitem[{Graepel, Herbrich, and Shawe-Taylor(2005)}]{graepel2005pac}
Graepel, T.; Herbrich, R.; and Shawe-Taylor, J. 2005.
\newblock {PAC}-{B}ayesian compression bounds on the prediction error of
  learning algorithms for classification.
\newblock \emph{Machine Learning}, 59(1): 55--76.

\bibitem[{Gy{\"o}rfi and Weiss(2021)}]{GyWe21}
Gy{\"o}rfi, L.; and Weiss, R. 2021.
\newblock Universal consistency and rates of convergence of multiclass
  prototype algorithms in metric spaces.
\newblock \emph{Journal of Machine Learning Research}, 22(151): 1--25.

\bibitem[{Hanneke and Kontorovich(2019)}]{hanneke2019sharp}
Hanneke, S.; and Kontorovich, A. 2019.
\newblock A sharp lower bound for agnostic learning with sample compression
  schemes.
\newblock In \emph{Algorithmic Learning Theory}, 489--505. PMLR.

\bibitem[{Hanneke and Kontorovich(2021)}]{hanneke2021stable}
Hanneke, S.; and Kontorovich, A. 2021.
\newblock Stable Sample Compression Schemes: New Applications and an Optimal
  {SVM} Margin Bound.
\newblock In \emph{Algorithmic Learning Theory}, 697--721. PMLR.

\bibitem[{Hanneke et~al.(2021)Hanneke, Kontorovich, Sabato, and
  Weiss}]{HaKoSaWe20}
Hanneke, S.; Kontorovich, A.; Sabato, S.; and Weiss, R. 2021.
\newblock Universal {B}ayes consistency in metric spaces.
\newblock \emph{Ann. Statist.}, 49(4): 2129--2150.

\bibitem[{Hanneke, Kontorovich, and Sadigurschi(2019)}]{hanneke2019sample}
Hanneke, S.; Kontorovich, A.; and Sadigurschi, M. 2019.
\newblock Sample compression for real-valued learners.
\newblock In \emph{Algorithmic Learning Theory}, 466--488. PMLR.

\bibitem[{Kontorovich, Sabato, and Urner(2016)}]{kontorovichsabatourner16}
Kontorovich, A.; Sabato, S.; and Urner, R. 2016.
\newblock Active nearest-neighbor learning in metric spaces.
\newblock In \emph{Advances in Neural Information Processing Systems},
  856--864.

\bibitem[{Kontorovich, Sabato, and Weiss(2017)}]{KoSaWe17}
Kontorovich, A.; Sabato, S.; and Weiss, R. 2017.
\newblock Nearest-neighbor sample compression: Efficiency, consistency,
  infinite dimensions.
\newblock In \emph{Advances in Neural Information Processing Systems},
  1573--1583.

\bibitem[{Kpotufe and Dasgupta(2012)}]{kpotufe2012tree}
Kpotufe, S.; and Dasgupta, S. 2012.
\newblock A tree-based regressor that adapts to intrinsic dimension.
\newblock \emph{Journal of Computer and System Sciences}, 78(5): 1496--1515.

\bibitem[{Krauthgamer and Lee(2004)}]{KL04}
Krauthgamer, R.; and Lee, J.~R. 2004.
\newblock Navigating nets: {S}imple algorithms for proximity search.
\newblock In \emph{15th Annual ACM-SIAM Symposium on Discrete Algorithms},
  791--801.

\bibitem[{Kusner et~al.(2014)Kusner, Tyree, Weinberger, and
  Agrawal}]{kusner2014stochastic}
Kusner, M.; Tyree, S.; Weinberger, K.; and Agrawal, K. 2014.
\newblock Stochastic neighbor compression.
\newblock In \emph{International Conference on Machine Learning}, 622--630.
  PMLR.

\bibitem[{Li, Vit{\'a}nyi et~al.(2008)}]{li2008introduction}
Li, M.; Vit{\'a}nyi, P.; et~al. 2008.
\newblock \emph{An introduction to Kolmogorov complexity and its applications},
  volume~3.
\newblock Springer.

\bibitem[{Lin and Jeon(2006)}]{lin2006random}
Lin, Y.; and Jeon, Y. 2006.
\newblock Random forests and adaptive nearest neighbors.
\newblock \emph{Journal of the American Statistical Association}, 101(474):
  578--590.

\bibitem[{Littlestone and Warmuth(1986)}]{warmuth86}
Littlestone, N.; and Warmuth, M.~K. 1986.
\newblock Relating Data Compression and Learnability.
\newblock Unpublished.

\bibitem[{MacQueen et~al.(1967)}]{macqueen1967some}
MacQueen, J.; et~al. 1967.
\newblock Some methods for classification and analysis of multivariate
  observations.
\newblock In \emph{Proceedings of the fifth Berkeley symposium on mathematical
  statistics and probability}, volume~1, 281--297. Oakland, CA, USA.

\bibitem[{Mammen and Tsybakov(1999)}]{MaTs99}
Mammen, E.; and Tsybakov, A.~B. 1999.
\newblock Smooth discrimination analysis.
\newblock \emph{The Annals of Statistics}, 27(6): 1808--1829.

\bibitem[{McInnes, Healy, and Melville(2018)}]{mcinnes2018umap}
McInnes, L.; Healy, J.; and Melville, J. 2018.
\newblock Umap: Uniform manifold approximation and projection for dimension
  reduction.
\newblock \emph{arXiv preprint arXiv:1802.03426}.

\bibitem[{Puchkin and Spokoiny(2020)}]{PuSp20}
Puchkin, N.; and Spokoiny, V. 2020.
\newblock An adaptive multiclass nearest neighbor classifier.
\newblock \emph{ESAIM: Probability and Statistics}, 24: 69--99.

\bibitem[{Scott and Nowak(2006)}]{scott2006minimax}
Scott, C.; and Nowak, R.~D. 2006.
\newblock Minimax-optimal classification with dyadic decision trees.
\newblock \emph{IEEE transactions on information theory}, 52(4): 1335--1353.

\bibitem[{Snell, Swersky, and Zemel(2017)}]{snell2017prototypical}
Snell, J.; Swersky, K.; and Zemel, R.~S. 2017.
\newblock Prototypical networks for few-shot learning.
\newblock \emph{arXiv preprint arXiv:1703.05175}.

\bibitem[{Tsybakov(2004)}]{Tsy04}
Tsybakov, A.~B. 2004.
\newblock Optimal aggregation of classifiers in statistical learning.
\newblock \emph{The Annals of Statistics}, 32(1): 135--166.

\bibitem[{Vapnik(2013)}]{vapnik2013nature}
Vapnik, V. 2013.
\newblock \emph{The nature of statistical learning theory}.
\newblock Springer science \& business media.

\bibitem[{Watson(1981)}]{watson1981computing}
Watson, D.~F. 1981.
\newblock Computing the n-dimensional Delaunay tessellation with application to
  Voronoi polytopes.
\newblock \emph{The computer journal}, 24(2): 167--172.

\bibitem[{Xue and Kpotufe(2018)}]{xue2018achieving}
Xue, L.; and Kpotufe, S. 2018.
\newblock Achieving the time of $1$-{NN}, but the accuracy of $k$-{NN}.
\newblock In \emph{International Conference on Artificial Intelligence and
  Statistics}, 1628--1636. PMLR.

\end{thebibliography}

\newpage
\onecolumn
\part*{Supplementary Material}
\begin{figure}[h]%
\centering
\includegraphics[width=0.4\columnwidth]{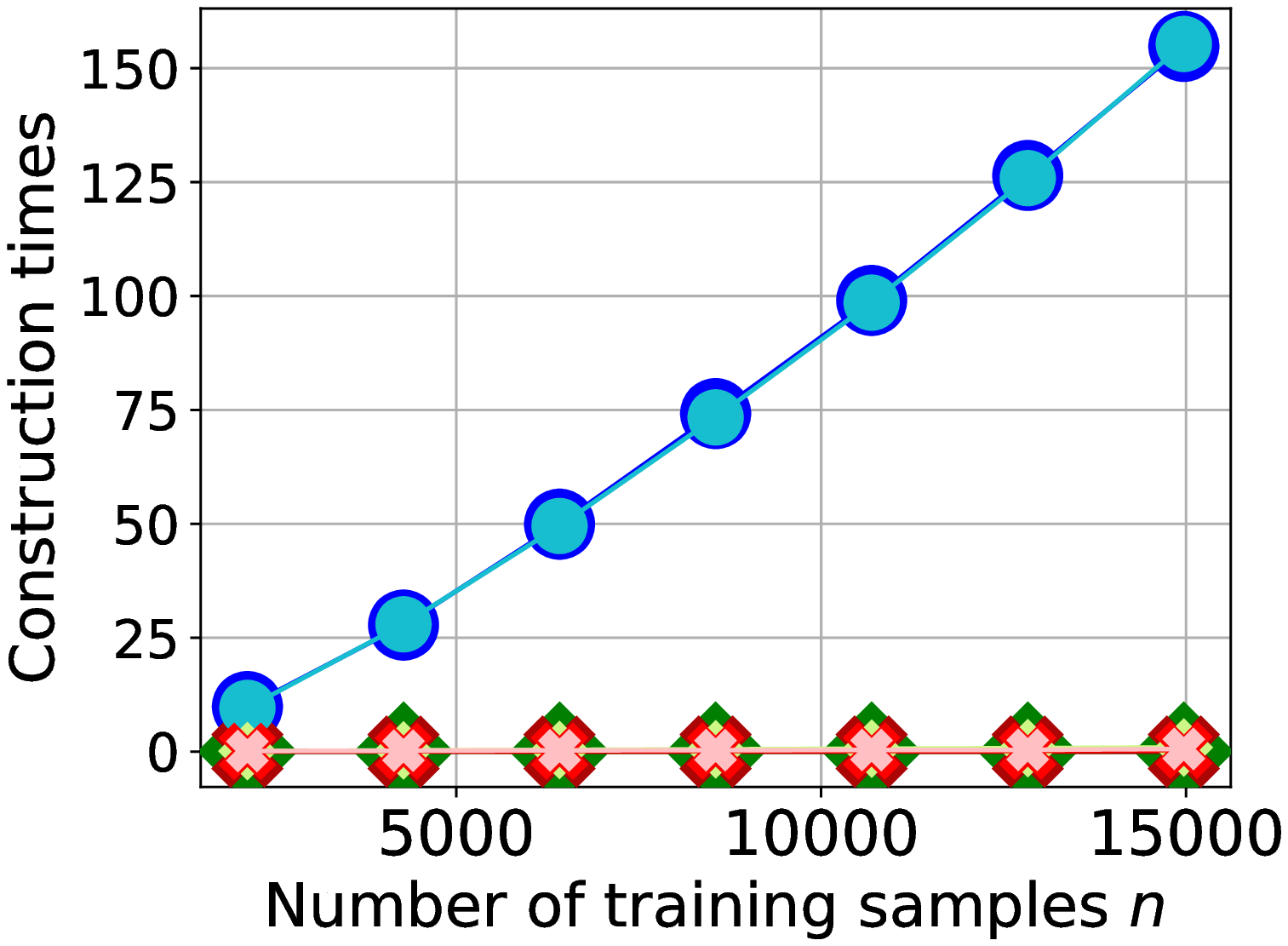}%
\hspace{30pt}
\includegraphics[width=0.4\columnwidth]{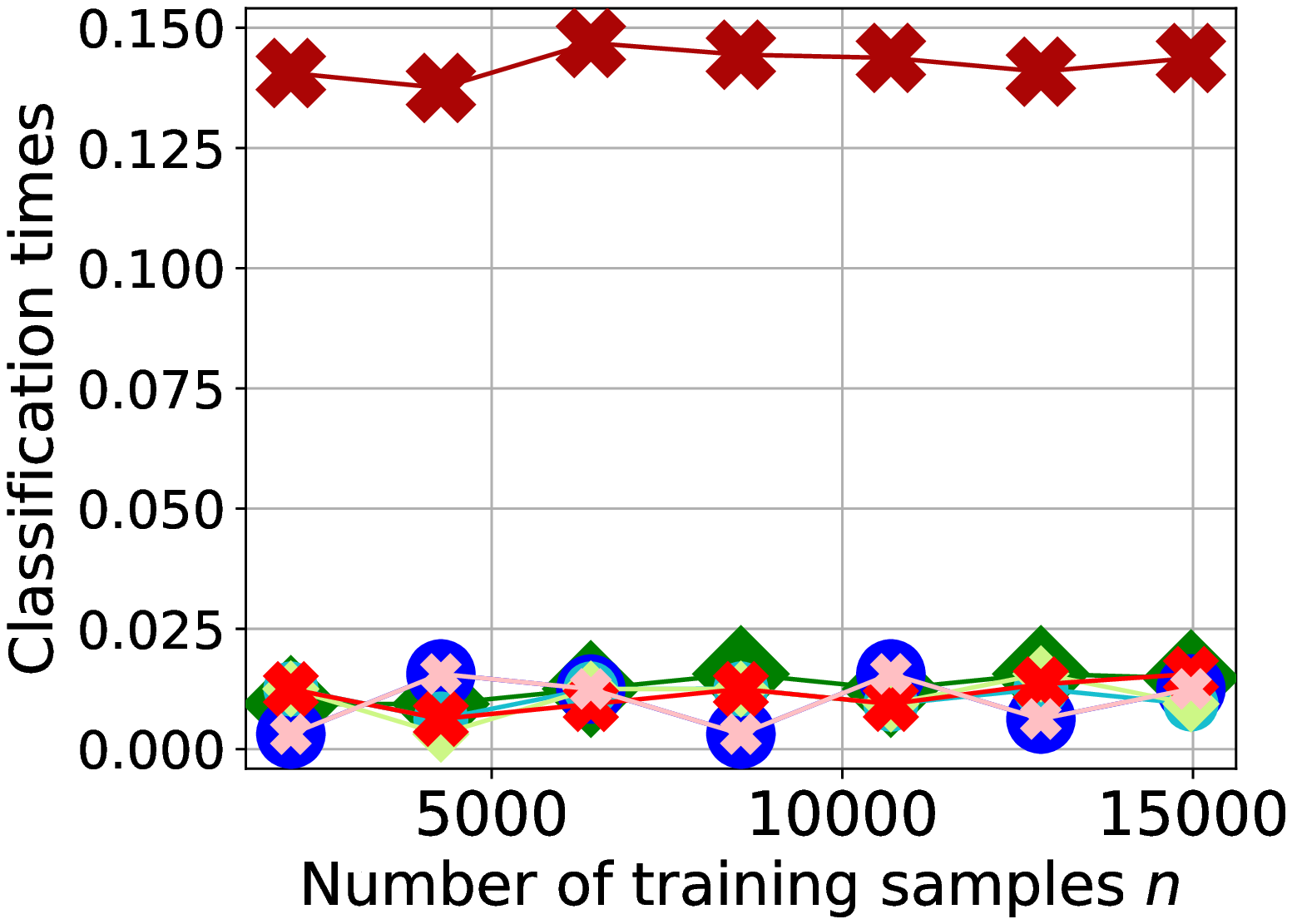}%
\\
\includegraphics[width=0.3\columnwidth]{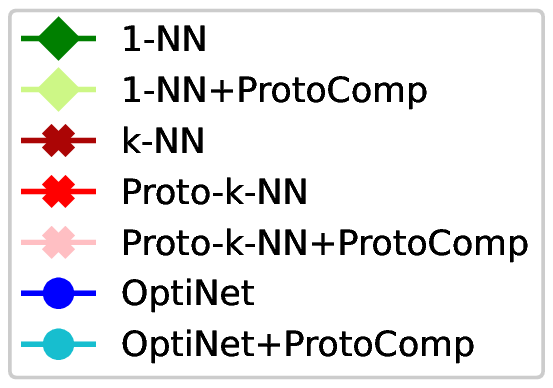}%
\caption{Construction and evaluation times
for the algorithms studied in the Experimental study section.
%Section \ref{sec:simulations}.
}%
\label{fig:runtime}%
\end{figure}

\iffalse
\textcolor{red}{Table \ref{table:notation} at the end of the manuscript summarizes the notation used throughout the paper.
\listoffigures
\listoftables
The following Table \ref{table:rates} summarizes the error and compression rates available for various prototype learning rules.}
\fi
Table \ref{table:rates} summarizes the error and compression rates available for various prototype learning rules.
Table \ref{table:notation}
%at the end of the manuscript
summarizes the notation used throughout the paper.
In Figure \ref{fig:runtime} we show the
construction and evaluation runtimes
for the algorithms studied in the Experimental study section.
Procedure \ref{alg:protocomp} is a pseudocode for \compalg{} and \compheu{}.
In this part we also provide proofs for the theorems that are provided in the paper
and
discuss Theorem \ref{thm:comp_agree Lebesgue}.
\part*{Proofs}
%\textcolor{red}{In this part we provide the proofs of the theorems that are provided in the paper
%% \ref{part:prel_res}
%%(in Sections \ref{sup-sec1}-\ref{sec:comp_rate_proof}) 
%and discuss Theorem \ref{thm:comp_agree Lebesgue}.
%% (in Section \ref{sup-sec4}).
%%Simulation results are given in Section \ref{sec:additional_simulations}.
%}

%In addition to the paper's notation that are summarized in Table \ref{table:notation}, additional

%\st{In this document we provide the proofs of the theorems in the main paper.}
%In this document we provide the proofs of all claims in the main paper and give some additional simulation results. 
%\inform{Omer: Rephrased this sentence.}

Additional
notation we use are
\[
\barBall{x}{r}
:=\{x'\in\X: \rho(x,x')\leq r\}
\]
to denote the closed sphere
%\des{In most place in the main paper we used the term ball, change the spheres to balls?}
around a given point $x\in\X$ with a given radius $r\geq 0$,
and
\begin{align*}
%\label{def:Seg}
\Seg(x,x')
:=
\{(1-u)x+ux':0\leq u\leq 1\}
\end{align*}
to denote the segment between two given points $x,x'\in\X$.
For any $n\in\mathbb N$ we let $[n]:=\{1,2,\dots,n\}$.

%\section{PROOFS}

\sectionmark{Proof of Theorem 1}{}
\section{Proof of Theorem \protect{\ref{thm}}}
\sectionmark{Proof of Theorem 1}{}
%\section{ PROOF OF THEOREM \protect{\ref{thm}}}
\label{sup-sec1}

For any measurable decision function $g:\X\to\Y$, 
%\todo{refer to \cite{gyorfi2020universal} instead. Omer: Understand ref}
\begin{align*}
\PROB\{g(X)\ne Y\mid X\}
&=
1-\PROB\{g(X)= Y\mid X\}\\
&=
1-\sum_{j=1}^M\PROB\{g(X)= Y=j\mid X\}\\
&=
1-\sum_{j=1}^M\IND{g(X)=j} P_j(X)\\
&=
1-P_{g(X)}(X)
.
\end{align*}
This implies
\begin{align}
\nonumber
\EXP\{L(g_{n})\}-L^*
&=
\EXP\{P_{g^*(X)}(X)-P_{g_n(X)}(X)\}\\
\nonumber
&=
\EXP\left\{\int (P_{g^*(x)}(x)-P_{g_n(x)}(x))
%\IND_{\{g^*(x)\ne  g_n(x)\}}
\mu(dx) \right\}
\\
\label{eq:ms_decomp}
%\nonumber
& \leq
\EXP\left\{\ms_{\gamma}\right\} + 
\EXP\left\{\int_{\UB{\gamma}} (P_{g^*(x)}(x)-P_{g_n(x)}(x))
%\IND_{\{g^*(x)\ne  g_n(x)\}}
\mu(dx) \right\},
%\\
%&=
%\EXP\left\{\int I_n(x)\mu(dx)\right\}
%\\
%&=
%\EXP\left\{\int \EXP\{I_n(x) \mid \bX'_n\}\mu(dx)\right\},
\end{align}
where $\UB{\gamma}$ is 
the 
%$\UB{\gamma}$ is the
$\gamma$-envelope around 
%the unlabeled samples 
$\Xul$,
% defined by
%\todo{Take close ball?}
%\todo{Need to change it to $\bigcup_{X' \in \Xg(\gamma)} S_{X', 2\gamma}$?}
\begin{align}
\label{eq:UB}
\UB{\gamma} 
%= \UB{\gamma}(\Xul) 
=
%\bigcup_{X'_i \in \Xul} 
\bigcup_{i = 1}^m 
\Ball{X'_i}{\gamma},
%S_{X'_i, \gamma}
\end{align}
where 
$\Ball{x}{\gamma}= \{x'\in\X : \rho(x,x')<\gamma\}$
is the open ball around $x$ with radius $\gamma$,
and
%as in \eqref{eq:UB}, 
\begin{align*}
\ms_{\gamma} := \ms_{\gamma}(\Xul) := \mu(\X\setminus\UB{\gamma})
\end{align*}
%$\ms_{\gamma_n}(\Xul)$ be 
is the $\gamma$-missing-mass of $\Xul$, and $g_n:=g_{n,m,\gamma}$.
By the law of total expectation,
\begin{align*}
%\EXP\{L(g_{n})\}-L^*
%&=
%\EXP\{P_{g^*(X)}(X)-P_{g_n(X)}(X)\}
& \EXP\left\{\int_{\UB{\gamma}} (P_{g^*(x)}(x)-P_{g_n(x)}(x))
%\IND_{\{g^*(x)\ne  g_n(x)\}}
\mu(dx) \right\}
\\
&=
\EXP\left\{\int_{\UB{\gamma}}  \EXP\left\{(P_{g^*(x)}(x)-P_{g_n(x)}(x))\IND{g^*(x)\ne  g_n(x)}\mu(dx) \mid \Xul\right\} \right\}\\
%&=
%\EXP\left\{\int I_n(x)\mu(dx)\right\}
%\\
&=
\EXP\left\{\int_{\UB{\gamma}}  \EXP\{I_n(x) \mid \Xul\}\mu(dx)\right\},
\end{align*}
where
\begin{align*}
I_n(x)
&=
(P_{g^*(x)}(x)-P_{g_n(x)}(x))\IND{P_{g^*(x)}(x)> P_{g_n(x)}(x)}\IND{ P_{n,g_n(x)}(x) \geq P_{n,g^*(x)}(x)}.
\end{align*}
For all $x\in\X$ abbreviate $V_x:=V_x(\bXg(\gamma))$ as the Voronoi cell containing $x$ and let
\begin{align*}
P_{n,j}(x)
:=\frac{\frac{1}{n}\sum_{X_i\in \Xl} \IND{Y_i=j,X_i\in V_x}}{\mu(V_x) }.
\end{align*}
(Note that in the main text, $P_{n,\ell,j}$ is defined differently for convenience, but one can easily verify that it corresponds to the same classifier.)
The relation
\begin{align*}
&\{ P_{n,g_n(x)}(x)- P_{n,g^*(x)}(x)\geq 0\}\\
&=
\{ P_{n,g_n(x)}(x)-P_{g_n(x)}(x)+P_{g_n(x)}(x)-P_{g^*(x)}(x)+P_{g^*(x)}(x)-P_{n,g^*(x)}(x) \geq 0\}\\
&\subseteq
\left\{ \sum_{j=1}^M|P_{n,j}(x)-P_{j}(x)| \geq P_{g^*(x)}(x)- P_{g_n(x)}(x)\right\}
\end{align*}
yields
\begin{align*}
& \quad\,\,\IND{P_{g^*(x)}(x)> P_{g_n(x)}(x)}\IND{ P_{n,g_n(x)}(x)\geq P_{n,g^*(x)}(x)}
\\
&\le
\IND{ \sum_{j=1}^M|P_{n,j}(x)-P_{j}(x)| \geq P_{g^*(x)}(x)-P_{g_n(x)}(x)}.
\end{align*}
Thus,
\begin{align*}
&I_n(x)\\
&\le
\sum_{\ell =1}^M(P_{g^*(x)}(x)-P_{\ell}(x))\IND{ \sum_{j=1}^M|P_{n,j}(x)-P_{j}(x)|\geq P_{g^*(x)}(x)-P_{\ell}(x)}\IND{ g_n(x)=\ell \ne g^*(x)}\\
&\le
\sum_{\ell =1}^M(P_{g^*(x)}(x)-P_{\ell}(x))\IND{ \sum_{j=1}^M|P_{n,j}(x)-P_{j}(x)|\geq P_{g^*(x)}(x)-P_{\ell}(x)}\IND{\ell \ne g^*(x)}\\
&\le
\sum_{j=1}^M\sum_{\ell =1}^M(P_{g^*(x)}(x)-P_{\ell}(x))\IND{\ell \ne g^*(x)}\IND{ |P_{n,j}(x)-P_{j}(x)|\geq (P_{g^*(x)}(x)-P_{\ell}(x))/M}.
\end{align*}
%Denote the instances obtained from the labeled set $\Dl$ by $\bX_n = \{X_1,\dots,X_n\}$ and 
Given $\Xul$, for every $j\in\Y$ put
%For any $x\in \X$ put
\begin{align}
\nonumber
\bar P_{n,j}(x)
&:=
\frac{\frac{1}{n}\EXP\{\sum_{X_i\in \bX_n } \IND{Y_i=j,X_i\in V_x }\mid \Xul\}}{\mu(V_x) }
\\
\label{bgest}
&=
\frac{\int_{V_x} P_j(z)\mu(dz)}{\mu(V_x) }.
\end{align}
% $V_x$ instead of $V$
Noting that
\begin{align*}
|P_j(x)- P_{n,j}(x)|
&\le
|P_j(x)- \bar P_{n,j}(x)|
+
|\bar P_{n,j}(x)- P_{n,j}(x)|
\end{align*}
%Recalling the notation $\bar P_{n,j,\gamma_n}$ in (\ref{bgest}) and 
and denoting
%we have that
%\begin{align*}
%\EXP\{L(g_{n})\mid \bX_n (\gamma_n)\}-L^*
%&\le
%\sum_{j=1}^M\sum_{\ell =1}^MJ_{n,1,j,\ell}+\sum_{j=1}^M\sum_{\ell =1}^MJ_{n,2,j,\ell},
%\end{align*}
%where
%\begin{align*}
%J_{n,1,j,\ell}
%&=\int (P_{g^*(x)}(x)-P_{\ell}(x))\IND_{\{\ell \ne g^*(x)\}}\\
%&\quad\cdot \PROB\{|P_{n,j,\gamma_n}(x)-\bar P_{n,j,\gamma_n}(x)|> (P_{g^*(x)}(x)-P_{\ell}(x))/M\mid \bX_n (\gamma_n)\}\mu(dx)
%\end{align*}
%and
%\begin{align*}
%J_{n,2,j,\ell}
%&=\int (P_{g^*(x)}(x)-P_{\ell}(x))\IND_{\{\ell \ne g^*(x)\}}\\
%&\quad\cdot \IND_{\{|\bar P_{n,j,\gamma_n}(x)-P_{j}(x)|> (P_{g^*(x)}(x)-P_{\ell}(x))/M \}}\mu(dx).
%\end{align*}
%Denoting
\[
D_{\ell}^*(x) := (P_{g^*(x)}(x)-P_{\ell}(x))\IND{\ell \ne g^*(x)},
\]
we have that
\begin{align*}
I_n(x)
&\leq
\sum_{j,\ell=1}^M D_{\ell}^*(x)\IND{ |P_{n,j}(x)-P_{j}(x)|\geq D_{\ell}^*(x)/M}
\\
&\leq
\sum_{j,\ell=1}^M D_{\ell}^*(x)\IND{ |P_{n,j}(x)- \bar P_{n,j}(x)|\geq D_{\ell}^*(x)/2M}
\\
& \quad + \sum_{j,\ell=1}^M D_{\ell}^*(x)\IND{ |\bar P_{n,j}(x)-P_{j}(x)|\geq D_{\ell}^*(x)/2M}.
\end{align*}
Thus,
\begin{align*}
%\EXP\{I_n(\textcolor{violet}{X}) \mid \textcolor{blue}{X\in\UB'_{\gamma_n}},\bX'_n  \}
%\EXP\{L(g_{n})\}-L^*
\int_{\UB{\gamma}}  \EXP\{I_n(x) \mid \Xul\}\mu(dx) 
&\le
\sum_{j,\ell=1}^M J_{n,1,j,\ell}+\sum_{j,\ell=1}^M J_{n,2,j,\ell},
\end{align*}
where
\begin{align}
\label{eq:J1}
J_{n,1,j,\ell}
&:=\int_{\UB{\gamma}} D_{\ell}^*(x)
%\IND_{\{\ell \ne g^*(x)\}}
%\\&\quad\cdot 
\PROB\{|P_{n,j}(x)-\bar P_{n,j}(x)|\geq D_{\ell}^*(x)/2M\mid \Xul \}\mu(dx)
\end{align}
and
\begin{align*}
%\label{eq:J2}
J_{n,2,j,\ell}
&:=\int_{\UB{\gamma}} D_{\ell}^*(x)
%\IND_{\{\ell \ne g^*(x)\}}
%\\&\quad\cdot 
\IND{|\bar P_{n,j}(x)-P_{j}(x)|\geq D_{\ell}^*(x)/2M }\mu(dx).
\end{align*}

Concerning the estimation error \eqref{eq:J1},
$J_{n,1,j,\ell}$, 
%\textcolor{violet}{delete: [recall that
%% instead of "let"
%$V_x$ is the cell in the partition $\Vor_{\ul,\gamma}$ into which $x$ falls.]}
for all $j\in[M]$, $i \in [n]$, and $x\in\X$, we define the
random variables
\[
K_{n,i,j}(x)=\IND{ Y_{i}=j,X_{i}\in V_x
%V_{\gamma_{n}}(x)
 }.
\]
%$(K_{n,i,j}\left(x\right)\right)_{i=1}^{n}$ 
Given $\Xul$,
%\textcolor{violet}{delete: [these random variables] put instead [for all $j\in\Y$, $(K_{n,i,j})_{i=1}^n$]}
for all $j\in\Y$, $(K_{n,i,j})_{i=1}^n$ are
%independent and identically distributed.
i.i.d.
Hence, the Bernstein
inequality yields
\begin{align*}
& \PROB\left\{\left|P_{n,j}(x)-\bar{P}_{n,j}(x)\right|
\geq D_\ell^*(x)/2M \mid \Xul\right\} 
\\
%& =\PROB\left\{\left|\frac{1}{n}\sum_{i=1}^{n}\left(\IND_{\left\{ Y_{i}=j,X_{i}\in V_{\gamma_{n}}(x)\right\} }-\EXP\{\IND_{\left\{ Y_{i}=j,X_{i}\in V_{\gamma_{n}}(x)\right\} } \mid \bX'_{n}\} \right)\right|
%\geq \mu(V_{\gamma_n}(x)) D_\ell^*(x)/(2M)\mid \bX'_{n}\right\} 
%\\
& =\PROB\left\{\left|\frac{1}{n}\sum_{i=1}^{n}\left(K_{n,i,j}(x)-\EXP\left\{ K_{n,i,j}(x) \mid \Xul\right\}\right)\right|
\geq \mu(V_x) D_\ell^*(x)/2M
\mid \Xul\right\} 
\\
& \leq 2\exp\left(-\frac{n}{2}\frac{(\mu(V_x) D_\ell^*(x))^{2}}{4M^{2}}\frac{1}{\Var\{ K_{n,1,j}(x) \mid \Xul\} +\frac{\mu(V_x) D_\ell^*(x)}{3\cdot 2 M}}\right)\\
 & =:\left(*\right)
 .
\end{align*}
Considering the variance
\begin{align*}
& \Var\!\left\{ K_{n,1,j}(x) \mid \Xul \right\}  
\\
& =\PROB\left\{Y_{1}=j, X_{1}\in V_x \mid \Xul\right\} \cdot \left(1-\PROB\{ Y_{1}=j,X_{1}\in V_x \mid \Xul\} \right)
\\
& \leq
\PROB\left\{Y_{1}=j, X_{1}\in V_x \mid \Xul\right\}
%\P\left\{ Y_{1}'=j,X_{1}'\in V|\Xs_{n}\right\} 
\\&
\leq \PROB\left\{X_{1}\in V_x \mid \Xul\right\}
\\&
=\mu(V_x).
\end{align*}
In addition, $D_\ell^*(x) \leq 1$.
Consequently,
\begin{align}
\left(*\right) & \leq2\exp\left(-\frac{n(\mu(V_x) D_\ell^*(x))^{2}}{8M^{2}}\cdot\frac{1}{\mu(V_x)+\frac{\mu(V_x)}{6M}}\right)\nonumber
\\
 & \leq 2\exp\left(-\frac{3n \mu(V_x)D_\ell^*(x)^{2}}
%\left(P_{g^{*}\left(x\right)}\left(x\right)-P_{l}\left(x\right)\right)^{2}\m\left(V\right)}
{28M^{2}}\right).
%\label{eq:estim_bound}
\nonumber
\end{align}
Towards applying the margin condition, we first lower bound $\mu(V_x)$.
% for $x\in\UB{\gamma}$.
By the packing and covering properties of $\gamma$-nets, for any nucleus $\nuc\in\bXg(\gamma)$,
\begin{align}
\label{SinV}
\Ball{\nuc}{\gamma/2} \subseteq V_{\nuc}.
\end{align}
Indeed, if some $x$
%$x\in\UB{\gamma}$ 
has $\rho(\nuc,x)<\gamma/2$ but $x\notin V_{\nuc}$, then $x$ belongs to a Voronoi cell whose nucleus $\nuc'\neq \nuc$ is such that
\[
\rho(x,\nuc')\leq \rho(x,\nuc) 
<
\gamma/2.
\]
This however implies
\[
\rho(\nuc,\nuc')\leq \rho(\nuc,x) + \rho(x,\nuc')<
\gamma/2 + \gamma/2 = \gamma,
\]
in contradiction to the requirement that $\nuc$ and $\nuc'$ must have interdistance larger or equal to $\gamma$.

Denote by $\nuc_x$ the nucleus in $\Xul$ corresponding to $V_x$. Using (\ref{SinV}), the MMC in
\eqref{MMA} and the BAZ assumption in (\ref{BAFZ}), we have that
for $\gamma\leq 2\gamma_0$,
\begin{align*} 
%\label{eq:minimal_mass_of_cell}
\mu(V_x) \geq \mu(\Ball{
\nuc_x
}{\gamma/2})  \geq  
\kappa \dlb(\gamma/2)^d.
% = c_d \gamma^d.
%\lambda(\Ball{x}{\gamma_n/2})
 %C_d f(X'_n(x)) \gamma_n^d.
\end{align*}
Thus, assuming $\gamma\leq 2\gamma_0$,
\begin{align*}
\PROB\left\{\left|P_{n,j}(x)-\bar{P}_{n,j}(x)\right|
\geq D_\ell^*(x)/2M \mid \Xul\right\} 
&\leq 
2\exp\left(- 
\frac{3\kappa \dlb \cdot n \gamma^d D_\ell^*(x)^{2}}{28\cdot 2^d M^2}
%\frac{ n \gamma^d D_\ell^*(x)^{2}}{28 \cdot 3^d M^2}  
%\frac{3n \mu(V_x)(D_\ell^*(x))^{2}}{28M^{2}}
\right)
\\
& = 
2\exp\left(-c_d n \gamma^d  D_\ell^*(x)^{2}\right).
\end{align*}
%\textcolor{violet}{Does $c_d$ absorb $\frac 3 {28}$?}\\
Therefore,
\begin{align*}
J_{n,1,j,l}&\leq 2\int_{\UB{\gamma}} D_\ell^*(x)\exp\left(- c_d n \gamma^d  D_\ell^*(x)^{2}\right) \mu(dx)\nonumber
\\
&\leq 2\int D_\ell^*(x)\exp\left(- c_d n \gamma^d  D_\ell^*(x)^{2}\right) \mu(dx).
\end{align*}
The margin condition with parameter $\alpha$ means that
for $0\leq t\leq1$,
\begin{align*}
G(t) & :=\PROB\left\{ D_\ell^*(X) \leq t  \right\} 
\leq
 \PROB\left\{ P_{(1)}(X)-P_{(2)}(X)\leq t\right\} \leq c^{*}\cdot t^{\alpha}.
% ,\qquad 0\leq t\leq1.
\end{align*}
Thus, applying integration by parts as in \cite[Lemma 2]{doring2017rate},
%\begin{align*}
%J_{n,1,j,l} & \leq2\int_{0}^{1}s\exp\left(-c_d n \gamma^{d} s^{2}\right)G(ds).
%\nonumber
%\end{align*}
%Applying integration by parts,
%Denoting $s_*=1/\sqrt{2c_d n \gamma^{d}}$, the integrand $s\cdot\exp\left(-c_d n \gamma^{d} s^{2}\right)$ is monotonic increasing for $0\leq s \leq s^*$ and monotonic decreasing for $s\geq s^*$.
%Thus, under the margin constraint,
%the distribution $G^*(s)$ that maximizes the integral corresponds to an atom at $s_*$ with probability mass $\min\{c^* s_*^\alpha,1\}$ and the remaining mass as a density that is proportional to $s^{\alpha-1}$ whose support consists only $s\geq s_*$. Hence, 
%assuming $c^* s_*^\alpha < 1$,
\begin{align*}
J_{n,1,j,l} & \leq 2\int_{0}^{1}s\exp\left(-c_d n \gamma^{d} s^{2}\right)G(ds)\nonumber
% \\
% &= 2\int_{0}^{\frac{1}{\sqrt{2c_d n \gamma^{d}}}}s\exp\left(-c_d n \gamma^{d} s^{2}\right)G(ds) + 
%2\int_{\frac{1}{\sqrt{2c_d n \gamma^{d}}}}^{1}s\exp\left(-c_d n \gamma^{d} s^{2}\right)G(ds)
 \\
 & =
 2e^{-c_d n \gamma^{d}} -
% \frac{}{\sqrt{2c_d n \gamma^{d}}}
% \int_{0}^{\frac{1}{\sqrt{2c_d n \gamma^{d}}}}G(ds)
 2\int_{0}^{1} \exp\left(-c_d n \gamma^{d} s^{2}\right)
  \left(1- 2c_d n \gamma^{d} s^{2}\right)G(s)\, ds
  \\
  &\leq 
2e^{-c_d n \gamma^{d}} +
% \frac{}{\sqrt{2c_d n \gamma^{d}}}
% \int_{0}^{\frac{1}{\sqrt{2c_d n \gamma^{d}}}}G(ds)
  2c^* c_d n \gamma^{d}\int_{0}^{1} \exp\left(-c_d n \gamma^{d} s^{2}\right)
   s^{2+\alpha}\, ds
   \\
   \nonumber
   & \leq 2e^{-c_d n \gamma^{d}} +
% \frac{}{\sqrt{2c_d n \gamma^{d}}}
% \int_{0}^{\frac{1}{\sqrt{2c_d n \gamma^{d}}}}G(ds)
  2c^* (c_d n \gamma^{d})^{-(1+\alpha)/2}\int_{0}^{\infty} \exp\left( -u^{2}/8\right)
   u^{2+\alpha}\, du
%\end{align*}
%where $c^{**}>0$ is an appropriate constant.
%Thus,
%\begin{align*}  
%J_{n,1,j,l} & \leq
%2c^{*}\left(c_d n\gamma^{d}\right)^{-(\alpha+1)/2}
%+ c^{**}\left(c_d n\gamma^{d}\right)^{-(\alpha+1)/2}
% \int_{0}^{\infty}u^{\alpha}\exp\left(-u^{2}\right)du
%  %\label{eq:DU1}
  \nonumber
  \\
 & =O\left(\left(n\gamma^{d}\right)^{-(\alpha+1)/2}\right).
 \nonumber 
\end{align*}
%}%
%\begin{align*}
%J_{n,1,j,l} & \leq2\int_{0}^{1}s\exp\left(-c_d n \gamma^{d} s^{2}\right)G(ds)\nonumber \\
% & \leq c^{*}\alpha\int_{0}^{1}s \exp\left(-c_d n \gamma^{d} s^{2}\right)
%  s^{\alpha-1}ds
%  \nonumber \\
% & = c^{*}\alpha\left(c_d n\gamma^{d}\right)^{-(\alpha+1)/2}
% \int_{0}^{\infty}u^{\alpha}\exp\left(-u^{2}\right)du
%  %\label{eq:DU1}
%  \nonumber
%  \\
% & =O\left(\left(n\gamma^{d}\right)^{-(\alpha+1)/2}\right).
% \nonumber 
%\end{align*}

To bound the
approximation error
%$J_{n,2,j,\ell}$ in 
(\ref{eq:J2}),
%we decompose
%\begin{align*}
%%\label{eq:J2}
%J_{n,2,j,\ell}
%&=
%\int D_{\ell}^*(x)\IND_{\{|\bar P_{n,j}(x)-P_{j}(x)|\geq D_{\ell}^*(x)/(2M) \}}\mu(dx)
%\\
%&\leq
%\ms_{\gamma_n}(\Xul) + 
%\int_{\UB{\gamma_n}(\Xul)} D_{\ell}^*(x)\IND_{\{|\bar P_{n,j}(x)-P_{j}(x)|\geq D_{\ell}^*(x)/(2M) \}}\mu(dx),
%\end{align*}
%where $\ms_{\gamma_n}(\Xul)$ is the $\gamma_n$-missing-mass,
%\begin{align*}
%\ms_{\gamma_n}(\Xul) = \mu(\X\setminus\UB{\gamma_n}(\Xul)),
%\end{align*}
%and $\UB{\gamma_n}(\Xul)$ is as in \eqref{eq:UB}.
the H\"older continuity assumption implies that for all $x\in\UB{\gamma}$,
%that for $x\in\UB'_{\gamma_n}$,
\begin{align*}
& \left|\bar{P}_{n,j}(x)-P_{j}(x)\right| 
\\& =\left|\frac{1}{\mu(V_x)}\int_{V_x}P_{j}(z)\mu(dz)-\frac{1}{\mu(V_x)}\int_{V_x}P_{j}(x)\mu(dz)\right|\nonumber \\
 & \leq\frac{1}{\mu(V_x)}\int_{V_x}\left|P_{j}(z)-P_{j}(x)\right|\mu(dz)\nonumber 
\\
&= \frac{1}{\mu(V_x)}\int_{V_x\setminus \UB{\gamma} }\left|P_{j}(z)-P_{j}(x)\right|\mu(dz) + \frac{1}{\mu(V_x)}\int_{V_x\cap\UB{\gamma}}\left|P_{j}(z)-P_{j}(x)\right|\mu(dz)
\nonumber 
\\
& \leq
\frac{\mu(V_x\setminus\UB{\gamma})}{\mu(V_x)}+
\frac{1}{\mu(V_x)}\int_{V_x\cap\UB{\gamma}}C\rho(x,z)^\beta\mu(dz)
%\frac{1}{\mu(V_x\cap\UB{\gamma})}\int_{V_x\cap\UB{\gamma}}\min\{C\rho(x,z)^\beta,1\}\mu(dz)
.%\\
%& \leq
%C\rho(x,X'_n(x)) + \frac{C}{\mu(V_x)}\int_{V_x}\rho(z,X'_n(z))\mu(dz),
\nonumber
\end{align*}
%For $x\in\UB{\gamma}$, 
For the second term, since both $x$ and $z$ belong to $V_x\cap\UB{\gamma}$ they both share the same prototype in $\bXg(\gamma)$, and since
$\bXg(\gamma)$ is a $\gamma$-net of $\Xul$ it holds that $\Xinn{1}(x;\bXg(\gamma)) \leq 2\gamma$ and $\Xinn{1}(z;\bXg(\gamma)) \leq 2\gamma$.
%\begin{align*}
%\rho(x,z) \leq \rho(x,\nuc_x) + \rho(z; \Xinn{1}(z;\Xul)) \leq 2\gamma.
%\end{align*}
%they share the same nucleus $\nuc_z = \nuc_x$ and
%$\Xinn{1}(z;\Xul)=\nuc_x$, so
%\begin{align*}
%\rho(x,z) \leq \rho(x,\nuc_x) + \rho(z; \Xinn{1}(z;\Xul)) \leq 2\gamma.
%\end{align*}
Thus, for $x\in\UB{\gamma}$,
\begin{align*}
\rho(x,z) \leq \Xinn{1}(x;\bXg(\gamma)) +\Xinn{1}(z;\bXg(\gamma)) \leq 2\gamma + 2\gamma =4\gamma.
\end{align*}
Hence, for $x\in\UB{\gamma}$,
\begin{align*}
\left|\bar{P}_{n,j}(x)-P_{j}(x)\right|  
\leq \frac{\mu(V_x\setminus\UB{\gamma})}{\mu(V_x)}+  C(4\gamma)^\beta.
\end{align*}
Hence,
\begin{align}
\nonumber
J_{n,2,j,\ell}
& =\int_{\UB{\gamma}} D_{\ell}^*(x)\IND{|\bar P_{n,j}(x)-P_{j}(x)|\geq D_{\ell}^*(x)/2M }\mu(dx)
\\
\nonumber
& \leq
\int_{\UB{\gamma}} D_{\ell}^*(x)\IND{\frac{\mu(V_x\setminus\UB{\gamma})}{\mu(V_x)} + C(4\gamma)^\beta \geq D_{\ell}^*(x)/2M }\mu(dx)
\\
\nonumber
& \leq
\int_{\UB{\gamma}} D_{\ell}^*(x)\IND{4M\frac{\mu(V_x\setminus\UB{\gamma})}{\mu(V_x)} \geq D_{\ell}^*(x)}\mu(dx)
+
\int_{\UB{\gamma}} D_{\ell}^*(x)\IND{ 4M C(4\gamma)^\beta \geq D_{\ell}^*(x) }\mu(dx)
\\
& \leq
4M\mu(\X\setminus \UB{\gamma})
+
\int D_{\ell}^*(x)\IND{2CM(4\gamma)^\beta \geq D_{\ell}^*(x) }\mu(dx).
\label{eq:J2}
\end{align}
For the second term in \eqref{eq:J2}, the margin condition yields
\begin{align}
 \int 
D_{\ell}^*(x) \IND{2CM(4\gamma)^\beta  \geq D_{\ell}^*(x) }\mu(dx)
&=
\int  s \cdot \IND{s \leq 2CM(4\gamma)^\beta }G(ds)\nonumber
\\
& \leq
2CM(4\gamma)^\beta \int_0^{2CM(4\gamma)^\beta }  G(ds) \nonumber
\\
%&\leq 2CM(4\gamma)^\beta  G(ds)
%\nonumber
%\\
\nonumber
& \leq c^* \alpha\left (2CM(4\gamma)^\beta \right)^{\alpha+1}
\\
&= O(\gamma^{\beta(\alpha+1)}).	\nonumber %\label{eq:approx_bound}
%\\
%&=
%O( h(1/M_n(\gamma_n) )^{\alpha+1})+O(1/n ).
\end{align}
The first term in \eqref{eq:J2} is proportional to $\ms_\gamma$, as the term in \eqref{eq:ms_decomp}, whose expectation is bounded by
%Lastly,
\begin{align}
\EXP\{\ms_{\gamma}\} &= 
\EXP\left\{\int \IND{x\notin \UB{\gamma}} \mu(dx)\right\}\nonumber
\\
& = 
\int \EXP\left\{\IND{x\notin \UB{\gamma}}\right\} \mu(dx)\nonumber
\\
& = 
\int \PROB\left\{\rho(x, \Xinn{1}(x;\bX'_m)) \geq \gamma \right\} \mu(dx)\nonumber
\\
& = 
\int  \left(1 - \mu(\Ball{x}{\gamma})\right)^\ul \mu(dx)\nonumber
\\
& \leq 
\int \exp\left(-\ul \mu(\Ball{x}{\gamma})\right) \mu(dx).	\nonumber %\label{eq:EL_n_bound}
\end{align}
Applying the MMC in \eqref{MMA}, followed by the BAZ assumption \eqref{BAFZ},
we have that for $x\in\supp(\mu)$,
\begin{align*}
\exp\left(-\ul \mu(\Ball{x}{\gamma})\right) \leq \exp\left(- \kappa \ul \gamma^d f(x)\right)
\leq \exp\left(- \kappa \ul \gamma^d \dlb\right).
\end{align*}
Hence,
\begin{align*}
\EXP\{\ms_{\gamma}\} & \leq \exp\left(- \kappa\dlb \ul \gamma^d \right) = \exp\left(-\Omega(\ul \gamma^d) \right),
%\label{eq:missing_mass_bound}
\end{align*}
concluding the proof.
\qed
 
\sectionmark{Proof of Theorem 2}{}
\section{Proof of Theorem \ref{thm:comp_agree Lebesgue}}
\sectionmark{Proof of Theorem 2}{}
%\section{ PROOF OF THEOREM \ref{thm:comp_agree Lebesgue}}
\label{sup-sec2}
%\todo{Rewrite proof. Omer: Done.}
%Assume that $\mu$ has a density, $\supp(\mu)$ is a convex subset of $\R^d$ and that $\rho$ is induced by the $\ell_2$-norm.
%In order to prove Lemma \ref{lem:comp_agree Lebesgue} we use the following lemmas:

%\textit{Proof of Lemma \ref{lem:comp_agree Lebesgue}.}

\label{sec:comp_ident}
Under the event that all the instance-label pairs have the same label, \eqref{equal labels} trivially holds and we are done.
Henceforth, we assume the complementary.
Since $\mu$ has a density, the event
\begin{align}
\{X'_i\neq X'_j,\quad  \text{for any distinct $i,j\in[m]$}\}
\label{ev:i j distinct}
\end{align}
occurs with probability one, and so, we assume the event in \eqref{ev:i j distinct} as well.
Denote $(\tbX,\tbY)=\tD$.
As reasoned by the following lemma whose proof is given below, we also assume that $\tD\neq\emptyset$. This implies that $\Xinn{1}(x;\tbX)$ exists for any $x\in\X$.

%Lemma \ref{lem:comp_agree Lebesgue} will follow  from the following series of lemmas.
%\todo{Moved Lemma \ref{lem:tD is not empty} to here since it doesn't involve $\widehat F$. If ok, swap the order of the proofs below as well.}
%\des{Number theorems, lemmas and claims in the same counter or each of them on its own counter? - Continue counter}
\begin{lemma}
\label{lem:tD is not empty}
Under the notation and assumptions of Theorem \ref{thm:comp_agree Lebesgue},
the event $\{\tD\neq\emptyset\}$
occurs with probability one.
\end{lemma}

As will become clear below, the following lemma shows that the set on which \eqref{equal labels} fails has zero Lebesgue measure.
%\todo{Not the best place to state}

\begin{lemma}
\label{lem:Lebesgue F zero}
%Assume that $\supp(\mu)$ is convex and that
Assume that $\rho$ is the Euclidean metric.
Let $\bX'=\{X'_1,\dots,X'_m\}\subseteq \X=\R^d$ be a set of $m$ distinct examples.
For all $i,j,l\in[m]$, let
\begin{align}
\label{def:Fijk}
F_{i,j,l}:=\{x'\in\X: \rho(x',X'_i)=\rho(x',X'_j)=\rho(x',X'_l)\}
\end{align}
and
\begin{align}
\widehat F_{i,j,l}:=\{x'\in\X: x' \text{ is a linear combination of elements from }F_{i,j,l}\cup\{X'_i\}\}
,
\label{def:widehat Fijk}
\end{align}
and define
\begin{align}
\label{def:widehat F}
\widehat F:=\bigcup_{\substack{\text{distinct}\\i,j,l\in[m]}}\widehat F_{i,j,l}
.
\end{align}
Then,
\begin{align*}
\lambda(\widehat F)=0
.
\end{align*}
\end{lemma}
Let $\widehat F$ be as in \eqref{def:widehat F}.
By the assumption that each nucleus in $\bX'$ is unique (that is, the assumption on the event in \eqref{ev:i j distinct}), and by Lemma \ref{lem:Lebesgue F zero}, it suffices to show that \eqref{equal labels} holds for all $x\in\X\setminus\widehat F$. So, fix $x\in\X\setminus\widehat F$.
%\todo{If Lemma \ref{lem:tD is not empty} doesn't involve $\widehat F$ then move it above before Lemma \ref{}}
%\begin{lemma}
%\label{lem:tD is not empty}
%Under the notation and assumptions of Theorem \ref{thm:comp_agree Lebesgue},
%the event $\{\tD\neq\emptyset\}$
%occurs with probability one.
%\end{lemma}
%So, assume also that $\tD\neq\emptyset$.
%By this assumption, $\Xinn{1}(x,\tD)$ exists.
%
%Firstly, in order to show that , we need to show that $\tD\neq\emptyset$.
%From now on in this proof, all the statements are to be true with probability one.
% and thus $\Xinn{1}(x,\tD)$ exists.
%
%Now, 

\begin{lemma}
%\todo{I think it is better to state the following result as a lemma}
\label{lem:forall i exists j}
Let $x\in \X\setminus\widehat F$, where $\widehat F$ is as in \eqref{def:widehat F}.
Under the assumptions of Theorem \ref{thm:comp_agree Lebesgue}, assume that $|\D'|=m$.
For all $k\in[m]$, denote 
\begin{align}
\label{def:nuc i}
\nuc_k:=\Xinn{k}(x;\bX').
\end{align}
%\todo{???for any such $x$,??? Omer: $x$ is fixed.}

%\todo{This result relies on $x\in\X\setminus\widehat F$? Omer: Yes.}
If the event in \eqref{ev:i j distinct} occurs, then
\begin{align}
\label{forall i exists j}
\forall\, 2\leq i\leq m,\  \exists\, 1\leq j<i
\qquad
\text{s.t.}
\qquad
\nuc_i\in\NNm(\nuc_j,\bX')
.
\end{align}
\end{lemma}

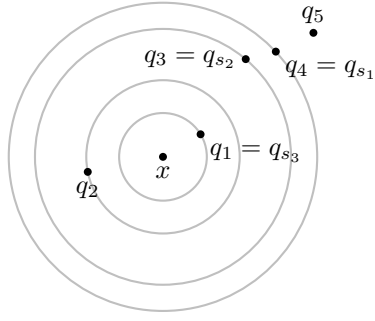
\begin{figure}
%\centering
%\ifthenelse{\equal\includefigs{1}}
%{
%\includestandalone[]{figs/fig1_neighbours_chain}%     without .tex extension
%or use
\begin{center}
\begin{tikzpicture}
\coordinate [] (x) at (0,0);
\coordinate [] (q1) at (0.5,0.3);
\coordinate [] (q2) at (-1,-0.2);
\coordinate [] (q3) at (1.1,1.3);
\coordinate [] (q4) at (1.5,1.4);
\coordinate [] (q5) at (2,1.65);

\foreach \point in {q1, q2, q3,q4}
\node [draw=lightgray, thick] () at (x) [circle through=(\point)] {};

\foreach \point in {x, q1, q2, q3,q4,q5}
\fill [black] (\point) circle (1.5pt);

\coordinate [draw, label=below:$x$] (xn) at (0,0);
\coordinate [label=below right:{$\nuc_1=\nuc_{s_3}$}] (q1n) at (0.5,0.3);
\coordinate [label=below:$\nuc_2$] (q2n) at (-1,-0.2);
\coordinate [label=left:{$\nuc_3=\nuc_{s_2}$}] (q3n) at (1.1,1.3);
\coordinate [label=below right:{${q_{4}}=\nuc_{s_1}$}] (q4n) at (1.5,1.4);
\coordinate [label=above:${q_{5}}$] (q5n) at (2,1.65);

\end{tikzpicture}
\iffalse
\caption{A demonstration of the inductive step with an example.
Assume that $\tD=\{\nuc_4,\nuc_5\}$.
Since $\nuc_4$ is the nearest neighbour of $x$ from among $\tbX$, then, $\nuc_4=\nuc_{s_1}$.
An index $j\in[4-1]$ that is guaranteed (in \eqref{forall i exists j}) to satisfy $\nuc_j\in\NNm(\nuc_4,\D')$ is, for example, $j=3$. So, $\nuc_3=\nuc_{s_2}$.
Since $\nuc_3\notin\tbX(\gamma)$, then, by the definition of $\tD$ in \eqref{def:tD}, $Y'(\nuc_3)=Y'(\nuc_4)$.
\\Repeating this argument, by \eqref{forall i exists j}, an index $j\in[3-1]$ such that $\nuc_j\in\NNm(\nuc_3,\D')$ is, for example, $j=1$. So, $\nuc_1=\nuc_{s_3}$. Since $\nuc_1\notin\tbX$, then, $Y'(\nuc_1)=Y'(\nuc_3)$.
So, $Y'(\nuc_4)$, the label of $x$ according to $\tD$, is equal to $Y'(\nuc_1)$, the label of $x$ according to $\D'$.
}
\fi
\end{center}
%}
%{\missingfigure{}}
\caption{A demonstration of the induction in the proof of Theorem \ref{thm:comp_agree Lebesgue} for a private example.
Assume that $\tD=\{\nuc_4,\nuc_5\}$.
Since $\nuc_4$ is the nearest neighbour of $x$ from among $\tbX$, then, $\nuc_{i_1}=\nuc_4$.
An index $j\in[4-1]$ that is guaranteed by Lemma \ref{lem:forall i exists j} to satisfy $\nuc_j\in\NNm(\nuc_4,\D')$ is, for example, $j=3$. So, $\nuc_{i_2}=\nuc_3$.
Since $\nuc_3\notin\tbX$, then, by the definition of $\tD$ in \eqref{def:tD}, $Y'(\nuc_3)=Y'(\nuc_4)$.
Repeating this argument, by Lemma \ref{lem:forall i exists j}, an index $j\in[3-1]$
such that
$\nuc_j\in\NNm(\nuc_3,\D')$ is, for example, $j=1$. So, $\nuc_{i_3}=\nuc_1$. Since $\nuc_1\notin\tbX$, then, $Y'(\nuc_1)=Y'(\nuc_3)$.
Finally, $Y'(\nuc_4)$, the label of $x$ according to $\tD$, is equal to $Y'(\nuc_1)$, the label of $x$ according to $\D'$, and so, $Y^{(1)}(x;\tD)=Y^{(1)}(x;\D')$.}
\label{fig:neighbours chain}
\end{figure}

The remaining of the proof is demonstrated in Figure
\ref{fig:neighbours chain}.
%\hyperref[fig:neighbours chain]{\getrefnumber{fig:neighbours chain}}

%\begin{figure}
%\fbox{Figure \ref{fig:neighbours chain}.}
%\caption{A demonstration of the inductive step with an example.
%Assume that $\tD=\{\nuc_4,\nuc_5\}$.
%Since $\nuc_4$ is the nearest neighbour of $x$ from among $\tbX$, then, $\nuc_4=\nuc_{s_1}$.
%An index $j\in[4-1]$ that is guaranteed (by Lemma \eqref{lem:forall i exists j}) to satisfy $\nuc_j\in\NNm(\nuc_4,\D')$ is, for example, $j=3$. So, $\nuc_3=\nuc_{s_2}$.
%Since $\nuc_3\notin\tbX(\gamma)$, then, by the definition of $\tD$ in \eqref{def:tD}, $Y'(\nuc_3)=Y'(\nuc_4)$.
%\\Repeating this argument, by \eqref{forall i exists j}, an index $j\in[3-1]$ such that $\nuc_j\in\NNm(\nuc_3,\D')$ is, for example, $j=1$. So, $\nuc_1=\nuc_{s_3}$. Since $\nuc_1\notin\tbX$, then, $Y'(\nuc_1)=Y'(\nuc_3)$.
%So, $Y'(\nuc_4)$, the label of $x$ according to $\tD$, is equal to $Y'(\nuc_1)$, the label of $x$ according to $\D'$.
%}
%\label{fig:neighbours chain}
%\end{figure}
Let $i_1\in[m]$ be the index for which $\nuc_{i_1}=\Xinn{1}(x;\tbX)$.
Proving by induction, suppose that we have already established a sequence $m\geq i_1>i_2>\dots>i_l\geq 1$ for some $1\leq l\leq m$ such that $\nuc_{i_{j}}\in\NNm(\nuc_{i_{j+1}},\bX')$ and $Y'(q_{i_{j+1}})=Y'(q_{i_{1}})$ for all $j=1,\dots,l-1$.
%\textcolor{violet}{delete: [If $i_l=1$, then 
%$$Y'(\Xinn{1}(x;\tbX))= Y'(q_{i_1})=  Y'(q_{i_l})=Y'(q_{1})=Y'(\Xinn{1}(x;\bX'))$$
%and the claim is proved.]}
If $i_l>1$, then, by Lemma \ref{lem:forall i exists j}, there exists $1\leq i_{l+1}<i_l$ such that $\nuc_{i_{l}}\in\NNm(\nuc_{i_{l+1}},\bX')$.
%, and suppose in contradiction that ${\nuc_{i_{l+1}}}\in \tbX$.
%\todo{$\nuc_{i_{l+1}}$?}
Recall that by the definition in \eqref{def:nuc i}, $\nuc_k=\Xinn{k}(x;\bX')$ for all $k\in[m]$. So, by $i_{l+1}<i_{l}\leq i_1$, $\nuc_{i_{l+1}}$ is closer\footnote{By ``closer" we mean firstly according to $\rho$, and in a case of tie, by the lower index.}
%\todo{The ordering depends on $x$}
to $x$ than $\nuc_{i_1}$.
Thus, we must have that ${\nuc_{i_{l+1}}}\notin\tbX$ (because otherwise $\nuc_{i_1}$ wouldn't have been equal to $\Xinn{1}(x;\tbX)$).
%\todo{No need for all the ``in contradiction'' argument. Just makes the text longer and less clear.}
In turn, by the definition of $\tD$ in \eqref{def:tD} and by $\nuc_{i_{l}}\in\NNm(\nuc_{i_{l+1}},\bX')$,
%\todo{above the other direction is considered}
 this implies that $Y'(\nuc_{i_{l+1}})=Y'(\nuc_{i_{l}})$. By the induction's assumption $Y'(\nuc_{i_{l}})=Y'(\nuc_{i_{1}})$. So, $Y'(\nuc_{i_{l+1}})=Y'(\nuc_{i_{1}})$.
%\inform{In the following put $l'$ instead of $l$ because they play different roles.}
Repeating the inductive step no more than $m-1$
%\inform{put $m-1$ instead of $m$}
iterations, this process must eventually produce the index $i_{l'}=1$ for some $l'\in[m]$, 
per \eqref{forall i exists j}.
%\inform{Why per \eqref{forall i exists j}? We use \eqref{forall i exists j} in the where is written "By Lemma 3".}
So, as guaranteed,
\begin{align*}
\Yinn{1}(x;\tD)
=Y'(\nuc_{i_1})
=Y'(\nuc_{i_{l'}})
=Y'(\nuc_{1})
=\Yinn{1}(x;\D')
.
\end{align*}
This concludes the proof.
\qed
%\ \\
%
%\noindent\textit{Proof of Lemma \ref{lem:tD is not empty}.}
%\subsubsection{Proof of Lemma \ref{lem:tD is not empty}
\subsection{Proof of Lemma \ref{lem:tD is not empty}}
Under the event that all the instance-label pairs have the same label, $\tD\neq \emptyset$ by its definition in Theorem \ref{thm:comp_agree Lebesgue}.
So, assume the complementary.
If $\D'$ contains $m\in\mathbb N$ instance-label pairs, assume also the event where $X'_{i'}\neq X'_{j'}$ for any distinct $i',j'\in[m]$.
Since $\mu$ has a density, this event occurs with probability one.

Let
%\todo{Consult with Roi - is it ok to let $i$ and $j$ instead of $i',j'$ in the argmin?}
\begin{align}
\{i,j\}
:=\argmin_{\{i',j'\}\subseteq [m]
:\ Y'(X'_{i'})\neq Y'(X'_{j'})} \rho(X'_{i'} ,X'_{j'})
%\\
%\{\nuc',\nuc ''\}:=\argmin_{\{\tilde\nuc',\tilde\nuc''\}\subseteq \Xul
%:\ Y'(\tilde\nuc')\neq Y'(\tilde\nuc'')} \rho(\tilde\nuc',\tilde \nuc'')
\label{def:nuc' nuc''}
\end{align}
%\des{Is it eligible to write "$\dots=\argmin$"? Should we change it to "$\dots\in\argmin$"?}
and
\begin{align}
\label{def:p0}
p_0:=(X'_{i}+X'_{j})/2
.
\end{align}
By $Y'(X'_{i})\neq Y'(X'_{j})$,
%we have $i\neq j$.
%So, by the aforementioned assumption,
$X'_{i}\neq X'_{j}$.
%\todo{Omer: Maybe add in a footnote in the definition of $Y'(X_i')$ that for completeness, the function $Y'(X'_i)$ is not $Y'(\cdot):\X\to\Y$, but $Y'(X'_{(\cdot)}):[m]\to\Y$ instead.}
Directly from the definition of $\NNm$ in \eqref{def:NNm}, $\Xinn{2}(p_0;\bX')\in\NNm(\Xinn{1}(p_0;\bX'),\bX')$.
We will show that
\begin{align}
\label{eq:NNs of p}
\{\Xinn{1}(p_0;\bX'),\Xinn{2}(p_0;\bX')\}=\{X'_{i},X'_{j}\}
.
\end{align}
Then, letting $(\tbX,\tbY)=\tD$, by the definition of $\tD$ in \eqref{def:tD} and by $Y'(X'_{i})\neq Y'(X'_{j})$, we will have that $\Xinn{1}(p_0;\bX')\in\tbX$. This will imply $\tD\neq \emptyset$.

So, towards \eqref{eq:NNs of p}, let $r_0:=\rho(p_0,X'_{i})=\rho(p_0,X'_{j})$, and suppose in a contradiction that there exists
$l\in[m]\setminus\{i,j\}$ such that
\[
X'_{l}\in  \barBall{p_0}{r_0}
.
\]
%where
%\begin{align*}
%\barBall{x}{r}:=\{x'\in\X:\rho(x,x')\leq r\}
%\end{align*}
%is the closed sphere of radius $r\geq0$ around $x\in\X$.
Since $Y(X'_{i})\neq Y'(X'_{j})$, then $Y'(X'_l)\neq Y'(X'_i)$ or $Y'(X'_l)\neq Y'(X'_j)$. W.L.O.G.\noindent\ assume the former.
By our assumption, $j\neq l$ implies $$X'_{j}\neq X'_l.$$
By the definition of $p_0$ in \eqref{def:p0}, $X'_i$ and $X'_j$ are antipodal points of $\barBall{p_0}{r_0}$.
Thus, from the geometrical properties of $\R^d$, the fact that $\rho$ is the Euclidean metric and  $X'_l\in\barBall{p_0}{r_0}\setminus\{X'_j\}$, we have that
$\rho(X'_i,X'_l)<\rho(X_i',X'_j)$. This is a contradiction to the definition of $i$ and $j$ in \eqref{def:nuc' nuc''}.
Hence, every nucleus $X'_l$, $l\in[m]\setminus\{i,j\}$, is strictly farther from $p_0$ than $X'_i$ and $X'_j$. This implies \eqref{eq:NNs of p}.
%is equidistant nor closer to $p_0$ \footnote{By ``closer", here and in the sequel, we mean first according to $\rho$, and in a case of a tie, the preferable according to the tie-breaking function. We will use the term ``$\rho$-closer" when we would like to talk about the $\rho$-distance only.} to $p_0$ than $\nuc'$ and $\nuc''$ do. So, \eqref{eq:NNs of p} is achieved, $\tD\neq \emptyset$.
\qed
%
%
%\ \\
%
%\noindent\textit{Proof of Lemma \ref{lem:Lebesgue F zero}.}
\subsection{Proof of Lemma \ref{lem:Lebesgue F zero}}
We say that $F\subseteq \R^d$ is a \textit{$k$-flat} if it is a $k$-dimensional affine subspace of $\mathbb R^d$, $1\leq k\leq d$.
A \textit{$0$-flat} is a singleton in $\mathbb R^d$ and a \textit{$(-1)$-flat} is the empty set. A $(d-1)$-flat is called a \textit{hyperplane}.
For all $i,j\in[m]$ define 
\begin{align*}
F_{i,j}:=\{x\in\X:\rho(x,X'_i)=\rho(x,X'_j)\},
\end{align*}
and note that given also $l\in[m]$, $F_{i,j,l}=F_{i,j}\cap F_{i,l}$.

Let distinct $i,j,l\in[m]$. Using $X'_j\neq X'_i\neq X'_l$ and the fact that $\rho$ is the Euclidean metric, one can easily show that $F_{i,j}$ and $F_{i,l}$ are hyperplanes.
Suppose in a contradiction that $F_{i,j}= F_{i,l}$.
Let $x_j:=(X'_i+X'_j)/2$ and $x_l:=(X'_i+X'_l)/2$.
%\textcolor{violet}{delete: [Since $\rho$ is induced by the $\ell_2$-norm,
%\begin{align*}
%&\rho(X'_i,x_j)
%=\rho(X'_i,(X'_i+X'_j)/2)
%=\Vert X'_i/2-X'_j/2 \Vert_{\ell_2}
%\qquad\text{and}
%\\
%&\rho(X'_j,x_j)
%=\rho(X'_j,(X'_i+X'_j)/2)
%=\Vert X'_j/2-X'_i/2 \Vert_{\ell_2}
%.
%\end{align*}
%Thus, $x_j\in F_{i,j}$. In the same way we can show that $x_l\in F_{i,l}$.]}
Since $\rho$ is the Euclidean metric, $x_j\in F_{i,j}$ and $x_l\in F_{i,l}$.
By $F_{i,j}=F_{i,l}$, $x_j,x_l\in F_{i,j,l}$.
Now, let $r_j:=\rho(x_j,X'_i)$ and $r_l:=\rho(x_l,X'_i)$.
%Also, for any point $x\in\X$ and for any radius $r\geq 0$ define the closed sphere
%\begin{align*}
%\barBall{x}{r}:=\{x'\in \X:\rho(x,x')\leq r\}
%.
%\end{align*}

By $x_j\in F_{i,j,l}$, $\rho(x_j,X'_l)=r_j$. So, $X_l'\in  \barBall{x_j}{r_j}\setminus\{X'_j\}$.
In addition, note that $X'_i$ and $X'_j$ are antipodal points in the close sphere $\barBall{x_j}{r_j}$.
By these two observations, by the geometry of $\R^d$ and due to the fact that $\rho$ is the Euclidean metric, we must have that $\rho(X'_i,X'_l)<\rho(X'_i,X'_j)$.
On the other hand, exchanging the roles of the indices $j$ and $l$ in this paragraph yields $\rho(X'_i,X'_j)<\rho(X'_i,X'_l)$.
This is a contradiction.
%by $x_l\in F_{i,j,l}$, we have $\rho(x_l,X'_j)=r_l$, so that $X'_j\in \barsp{x_l,r_l}$. As before, $X'_i$ and $X'_l$ are antipodal points in $\barsp{x_l,r_l}$. Theses two obse
%
%note that $\nuc_i$ and $\nuc_l$ are antipodal points in $\barsp{x_l,\rho(\nuc_l,x_l)}$. By $\nuc_j\in\barsp{x_l,\rho(\nuc_l,x_l)}\setminus\{\nuc_l\}$, $\rho(\nuc_i,\nuc_j)<\rho(\nuc_i,\nuc_l)$. This is a contradiction.
Thus, $F_{i,j}\neq F_{i,l}$.

It can be shown that the intersection of two distinct hyperplanes is a $k$-flat for some $k\in\{d-2,-1\}$.
So, $F_{i,j,l}=F_{i,j}\cap F_{i,l}$ is a $k_{i,j,l}$-flat for a certain $k_{i,j,l}\in\{d-2,-1\}$.
%Define $\widehat F_{i,j,l}$ to be the minimal $k'$-flat that contains $F_{i,j,l}\cup\{X'_i\}$, for some $-1\leq k'\leq d$.
Since $F_{i,j,l}$ is a $k_{i,j,l}$-flat and $X'_i\notin F_{i,j,l}$, one can verify that that $\widehat F_{i,j,l}$ is a $(k_{i,j,l}+1)$-flat.
Using elementary tools of measure theory, it can be proved that the Lebesgue measure of any $k'$-flat is zero, for all $k'<d$. Thus, $\lambda (\widehat F_{i,j,l})=0$.
By the generality of the $i,j$ and $l$,
\begin{align*}
\lambda( \widehat F)
\leq 
 \sum_{\substack{\text{distinct}\\ i,j,l\in[m]}}\lambda (\widehat F_{i,j,l})
 =0
.
\end{align*}
\qed
%where
%\begin{align*}
%\widehat F:=
%\bigcup_{\substack{\text{distinct}\\ i,j,l\in[m]}}\widehat F_{i,j,l}
%.
%\end{align*}
%So, by Lemma \ref{lem:comp_agree}, and due to the fact that $\mu$ is absolutely continuous with relation to $\lambda$, concludes the proof.
%
%
%\ \\
%
%\textit{Proof of Lemma \ref{lem:forall i exists j}.}
\subsection{Proof of Lemma \ref{lem:forall i exists j}}
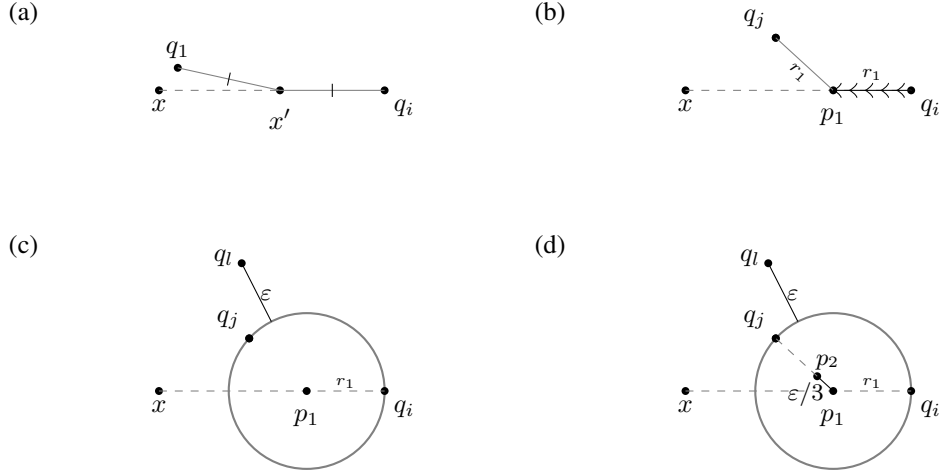
\begin{figure}
\centering
%\ifthenelse{\equal\includefigs{1}}
%{
   \begin{tikzpicture}[]
%]
\newcommand{\subFigs}{4}
\newcommand{\figN}{0}
\newcommand{\Xs}{0}
\newcommand{\Ys}{0}
\newcommand{\XlabelPos}{0}
\newcommand{\YlabelPosOT}{0}
\newcommand{\YlabelPosTF}{0}

   \foreach \i [count=\n]in {1,...,\subFigs}
{
\renewcommand{\figN}{\the\numexpr\subFigs -\i+1\relax}

\ifnum \figN=1
\renewcommand{\Xs}{0}
\renewcommand{\Ys}{0}
\fi
\ifnum \figN=2
\renewcommand{\Xs}{7}
\renewcommand{\Ys}{0}
\fi
\ifnum \figN=3
\renewcommand{\Xs}{0}
\renewcommand{\Ys}{-4}
\fi
\ifnum \figN=4
\renewcommand{\Xs}{7}
\renewcommand{\Ys}{-4}
\fi

%\begin{scope}[yshift = \the\numexpr(\i-1)*4\relax cm]
\begin{scope}[yshift = \the\numexpr\Ys\relax cm,xshift = \the\numexpr\Xs\relax cm]
\renewcommand{\XlabelPos}{-1.8}
\renewcommand{\YlabelPosOT}{1.3}
\renewcommand{\YlabelPosTF}{2.2}
\ifnum \figN=1
\coordinate [draw, label=below: (\symbol{\the\numexpr96+\figN\relax})] (sub) at (\XlabelPos,\YlabelPosOT);
\fi
\ifnum \figN=2
\coordinate [draw, label=below: (\symbol{\the\numexpr96+\figN\relax})] (sub) at (\XlabelPos,\YlabelPosOT);
\fi
\ifnum \figN=3
\coordinate [draw, label=below: (\symbol{\the\numexpr96+\figN\relax})] (sub) at (\XlabelPos,\YlabelPosTF);
\fi
\ifnum \figN=4
\coordinate [draw, label=below: (\symbol{\the\numexpr96+\figN\relax})] (sub) at (\XlabelPos,\YlabelPosTF);
\fi
	\coordinate [draw, label=below:$x$] (x) at (0,0);
	\coordinate [draw, label=below right:$q_i$] (qi) at (3,0);

	\foreach \point in {x, qi}
	\fill [black] (\point) circle (1.5pt);
	
	\ifnum \figN=1
	
	\coordinate [draw, label=above:$q_1$] (q1) at (0.25,0.3);
%	\coordinate [draw, label=left:$q_1$] (q1) at (-0.9,0.55);
	
	\foreach \point in { q1}
	\fill [black] (\point) circle (1.5pt);
	
	\path (q1) -- (qi) coordinate[midway] (qiPq1);
	\path [name path=qiPq1T]  ($(qiPq1)!1cm!270:(q1)$) -- ($(qiPq1)!1cm!90:(q1)$);
	\path [name path= qiPx] (x) -- (qi);
    \path [name intersections={of=qiPx and qiPq1T,by=x''}];
    \node[label={below:$x'$}] at (x'') {};
    \fill[] (x'') circle (1.5pt);
    
    \path [name path = xPx'', color=gray, dashed, draw] (x) -- (x'') coordinate[midway] (xPx'');
    \path [color=gray, thin, draw] (q1) -- (x'') coordinate[midway] (q1Px'');
    \path [name path = qiPx'', color=gray,thin, draw] (qi) -- (x'') coordinate[midway] (qiPx'');
    
    \path [name path=q1Px''T, draw]  ($(q1Px'')!2.5pt!270:(q1)$) -- ($(q1Px'')!2.5pt!90:(q1)$);
    \path [name path=qiPx''T,  draw]  ($(qiPx'')!2.5pt!270:(qi)$) -- ($(qiPx'')!2.5pt!90:(qi)$);
    \fi
    
	\ifnum \figN>1
	
	\coordinate [draw, label=above left:$q_j$] (qj) at (1.2,0.7);
    \fill[] (qj) circle (1.5pt);
    
	\path [name path = xPqi, color=gray, dashed, draw] (x) -- (qi) coordinate[midway] (xPqi);
	\path (qj) -- (qi) coordinate[midway] (qiPqj);
	\path [name path=qiPqjT]  ($(qiPqj)!1cm!270:(qj)$) -- ($(qiPqj)!1cm!90:(qj)$);
	\path [name path= qiPx] (x) -- (qi);
    \path [name intersections={of=qiPx and qiPqjT,by=p1}];
    \node[label={below:$p_1$}] at (p1) {};

	\foreach \point in { qj, p1}
	\fill [black] (\point) circle (1.5pt);

	\fi
	
	\ifnum \figN=2
	
 	 \node[yshift=-3pt,label={above:\scriptsize{$r_1$}}] at ($(p1)!0.5!(qi)$) {};

 	 \path [color=gray, thin, draw, sloped] (qj) edge node [color=black,below,xshift=0.5pt, yshift=1.25pt] {\scriptsize{$r_1$}} (p1);
 	 
 	 \path [decoration={
markings, mark=between positions 0 and 1 step 
%2.5mm with {\arrowreversed{Stealth[]}}}]
%2.5mm with {\arrowreversed{Straight Barb[length=5pt,width=5pt]}}}]
 2.1mm with {\arrowreversed{Classical TikZ Rightarrow[open, width=2mm,length=1.1mm]}}},postaction={decorate}, draw] (p1) -- (qi) coordinate[midway] (p1Pqi);
    
	\fi    
    
    \ifnum \figN>2
    
	\coordinate [] (ql) at (1.1,1.7);
    \fill[] (ql) circle (1.5pt);
    \node[xshift=4pt, yshift=2pt, label={left:$\nuc_l$}] at (ql) {};

 	 \node[yshift=-6pt,label={above:\tiny{$r_1$}}] at ($(p1)!0.5!(qi)$) {};
    
 \begin{scope}[on background layer]
 \node [name path=barS, draw=gray,  thick] (barS) at (p1) [circle through=(qi)] {};
    
    \path [name path= p1Pql] (p1) -- (ql);
    \path [name intersections={of=p1Pql and barS,by=z}];
    
    \end{scope}
    
    \path [draw] (ql) edge node [color=black,right,xshift=-2pt, yshift=-0.8pt] {\footnotesize{$\varepsilon$}} (z);

    \fi
    
    \ifnum\figN>3
    
	\path[name path=veCirc]  let \p1 = ($ (ql) - (z) $), \n1 = {veclen(\x1,\y1)/3}
    in circle [at = (p1), radius =\n1] ;

    \path [name path= p1Pqj] (p1) -- (qj);
    \path [name intersections={of=p1Pqj and veCirc,by=p2}];
    \fill[] (p2) circle (1.5pt);
    \node[xshift=-7.5pt, yshift=-4.5pt, label={above right:\small{$p_2$}}] at (p2) {};

    \path [draw] (p1) edge node [color=black,left,xshift=3pt, yshift=-4pt] {\footnotesize{$\varepsilon/3$}} (p2);

%    \path [draw] (p1) edge node [color=black,below left,xshift=-2pt, yshift=1pt] {\scriptsize{$\varepsilon/3$}} (p2);
%    \path[->]
    
    \path [color=gray, dashed, draw] (p2) -- (qj);
    
    \fi
\end{scope}
    }
   \end{tikzpicture}

\iffalse
\caption{
a) We show that there \underline{exists} a point ($x''$) in $\Seg (x,\nuc_i)\setminus\{\nuc_i\}$ with an equidistance to $\nuc _1$ and $\nuc_i$.
\\b) Thus, starting from $\nuc_i$, we can ``walk" on the segment $\Seg(x,\nuc_i)\setminus\{\nuc_i\}$, and ``stop" on the first point $p_1$ which has an equidistance to $\nuc_i$ and $\nuc_j$ for some $j\in[i-1]$.
\\c) Since $p_1$ is the ``first" point with an equidistance to $\nuc_i$ and to $\Xul\setminus\{\nuc_i\}$, there is no nucleus in the open ball $S_{p_1,r_1}$. Since $x\notin F$, no nucleus $\nuc_l\in\Xul\setminus\{\nuc_i,\nuc_j\}$ resides in the sphere $\partial{S}_{p_1,r_1}$.
Hence, $\varepsilon$, the distance between $\Xul\setminus\{\nuc_i,\nuc_j\}$ and $\bar S_{p_1,r_1}$, is a positive quantity. (If that distance is greater than $r_1$, or that $\Xul\setminus\{\nuc_i,\nuc_j\}=\emptyset$, then $\varepsilon=r_1$).
\\d) Let $p_2$ the point with distance $\varepsilon/3$ from $p_1$ towards $\nuc_j$. (i) The distance of $p_2$ to $\nuc_l$ is at least $r_1+2\varepsilon/3$, for any $l\in[m]\setminus\{i,j\}$, and its distance to $\nuc_i$ is at most $r_1+\varepsilon/3$. (ii) The geometry of $\mathbb R^d$ with the Euclidean metric $\rho$   implies that $p_2$ is closer to $\nuc_j$ than to $\nuc_i$.
\\ By (i) and (ii), the first- and second-NN of $p_2$ are $\nuc_j$ and $\nuc_i$, respectively. Then, by definition, $\nuc_i\in\NNm(\nuc_j,\D')$.
   }
\fi
%}
%{\missingfigure{}}
\caption{
An illustration of of the proof of Lemma \ref{lem:forall i exists j}.
\textbf{(a)} We show that there exists a point ($x'$) in $\Seg (x,\nuc_i)\setminus\{\nuc_i\}$ with an equidistance to $\nuc _1$ and $\nuc_i$.
\textbf{(b)} Thus, starting from $\nuc_i$, we can ``walk" on the segment $\Seg(x,\nuc_i)\setminus\{\nuc_i\}$, and ``stop" on the first point $p_1$ which has an equidistance to $\nuc_i$ and $\nuc_j$ for some $j\in[i-1]$.
\textbf{(c)} Since $p_1$ is the ``first" point with an equidistance to $\nuc_i$ and to $\bX'\setminus\{\nuc_i\}$, there is no nucleus in the open ball $\Ball{p_1}{r_1}$. Since $x\notin \widehat F$, no nucleus $\nuc_l\in\bX'\setminus\{\nuc_i,\nuc_j\}$ resides in the closed sphere $\barBall{p_1}{r_1}$.
Hence, $\varepsilon$, the distance between $\bX'\setminus\{\nuc_i,\nuc_j\}$ and $\barBall{p_1}{r_1}$, is a positive quantity. (If that distance is greater than $r_1$, or that $\bX'\setminus\{\nuc_i,\nuc_j\}=\emptyset$, we take $\varepsilon=r_1$.)
\textbf{(d)} Let $p_2$ the point with distance $\varepsilon/3$ from $p_1$ towards $\nuc_j$. (i) The distance of $p_2$ to $\nuc_l$ is at least $r_1+2\varepsilon/3$, for any $l\in[m]\setminus\{i,j\}$, and its distance to $\nuc_i$ is at most $r_1+\varepsilon/3$. (ii) The geometry of $\mathbb R^d$ with the Euclidean metric $\rho$   implies that $p_2$ is closer to $\nuc_j$ than to $\nuc_i$.
By (i) and (ii), the first- and second-NN of $p_2$ are $\nuc_j$ and $\nuc_i$, respectively. Then, by definition, $\nuc_i\in\NNm(\nuc_j,\D')$.
}
\label{fig:forall i exists j}
\end{figure}

Fix some $2\leq i\leq m$.
%The structure of the proof is as follows:
%We will define a point $p_1\in\X$ and an index $j<i$.
%Then, we will define another point $p_2$ which is dependent on $p_1$ and on a certain quantity $\ve$.
%\todo{Up to here non-informative. As phrased, I don't think it is helpful.}
We will define a point $p_2$ and will show that its first- and second-NN's are $\nuc_j$ and $\nuc_i$, respectively, for a certain $1\leq j<i$.
By the definition of $\NNm$ in $\eqref{def:NNm}$, this will conclude the proof.
A demonstration of the proof is shown in Figure \ref{fig:forall i exists j}.

For any $a,b,c,d\in \X$
%, denote the segment
%\begin{align}
%\label{def:Seg}
%\Seg(a,b):=\{(1-t)a+tb:0\leq t\leq 1\},
%\end{align}
%\todo{Put $\Seg[a,b]$ instead of $\Seg(a,b)$. I think it is more elegant to write $\Seg(a,b]$ instead of $\Seg(a,b)\setminus\{a\}$.}
%and
define the following continuous function
%$\rho_{\Seg(a,b)}^{c,d}
%\footnote{Formally, the subscript ``$\Seg$" is notational and doesn't represent the function $\Seg$ because $\Seg(c,d)=\Seg(d,c)$ but $\rho^{a,b}_{\Seg(c,d)}$ doesn't necessarily equal to $ \rho^{a,b}_{\Seg(d,c)}$.}
%:[0,1]\to\R$
$\rho_{\Seg(a,b)}^{c,d}
:[0,1]\to\R$
(\footnote{Formally, the subscript ``$\Seg$" is notational and doesn't represent the function $\Seg$ because $\Seg(c,d)=\Seg(d,c)$ but $\rho^{a,b}_{\Seg(c,d)}$ doesn't necessarily equal to $ \rho^{a,b}_{\Seg(d,c)}$.})
according to the rule
\begin{align*}
%\label{def:rho Seg}
t\mapsto
\rho((1-t)a+tb,c)-\rho((1-t)a+tb,d)
.
\end{align*}
Since $i\neq 1$, (and by the assumption on the event in \eqref{ev:i j distinct}), $\nuc_1\neq \nuc_i$. And so, 
\begin{align*}
\rho_{\Seg(\nuc_i,x)}^{\nuc_i,\nuc_1}(0)
= \rho(\nuc_i,\nuc_i) - \rho(\nuc_i,\nuc_1)
<0.
\end{align*}
%\todo{add a figure illustrating the quantities in the proof of \eqref{forall i exists j}}
By the definition of $\nuc_k, k\in[m]$, in \eqref{def:nuc i},
\begin{align*}
\rho_{\Seg(\nuc_i,x)}^{\nuc_i,\nuc_1}(1) = \rho(x,\nuc_i) - \rho(x,\nuc_1) \geq 0.
\end{align*}
By the Intermediate Value Theorem, the function $\rho_{\Seg(\nuc_i,x)}^{\nuc_i,\nuc_1}$ has a root $0<t'\leq 1$. Then, $\nuc_1$ and $\nuc_i$ are equidistant to the point $x':=(1-t')\nuc_i+t'x\in\Seg(\nuc_i,x)\setminus\{\nuc_i\}$.
This, in turn, ensures us the existence of
\begin{align}
\label{def:p_1}
p_1 :=
\argmin_{\substack{p\in\Seg(\nuc_i,x):\ \exists  j<i,\rho(p,\nuc_j)=\rho(p,\nuc_i)}}
\rho(p,\nuc_i)
\in \Seg(\nuc_i,x)\setminus\{\nuc_i\}
.
\end{align}
Define $r_1:=\rho(p_1,\nuc_i)$, and fix an index $j\in[i-1]$ that satisfies $\rho(p_1,\nuc_j)=r_1$.
We will show that
%\todo{Awkward phrasing of the paragraph. First show the Eq that follows and then define $\ve$.}
\begin{align}
\rho(p_1,\nuc_l)>r_1,\qquad \forall l\in[m]\setminus\{i,j\}
.
\label{eq:1-2NN of p1 are nuc i nuc j}
\end{align}
For all $i',j',l'\in[m]$, let $F_{i',j',l'}$ and $\widehat F_{i',j',l'}$ as are defined in \eqref{def:Fijk} and in \eqref{def:widehat Fijk}.
Suppose in a contradiction that $\rho(p_1,\nuc_l)=r_1$ for some $l\in[m]\setminus\{i,j\}$.
For all $k\in[m]$,
%\todo{notation $t$ already used.}
 let $s_k\in [m]$ be the index for which $X'_{s_k}=\nuc_k$.
So, $p_1\in F_{s_i,s_{j},s_{l}}$.
%Recall that $\widehat F_{s_i,s_{j},s_{l}}$ is a $k$-flat containing $F_{s_i,s_{j},s_{l}}\cup\{X'_{s_i}\}$, $k<d$. Specifically,
Since $p_1\in\Seg(x,X'_{s_i})$, $x$ is a linear combination of $p_1\in F_{s_i,s_{j},s_{l}}$ with $X'_{s_i}$.
As such, by definition, $x\in \widehat F_{s_i,s_{j},s_{l}} \subseteq \widehat F$.
This is a contradiction to $x\notin \widehat F$.

Now, suppose in a contradiction that there exists $l\in[m]\setminus\{i,j\}$ s.t.\noindent\ 
$\rho(p_1,\nuc_l)<r_1=\rho(p_1,\nuc_i)$.
In order to see that $l<i$, note that
\begin{align*}
\rho(x,\nuc_l)
&\leq \rho(x,p_1)+\rho(p_1,\nuc_l)
\\&< \rho(x,p_1)+\rho(p_1,\nuc_i)
\\&=\rho(x,\nuc_i),
\end{align*}
where the last equality is by $p_1\in \Seg(x,\nuc_i)$ and due to the fact that $\rho$ is the Euclidean metric.
Now,
\begin{align*}
\rho_{\Seg(\nuc_i,p_1)}^{\nuc_i,\nuc_l}(0)
=\rho(\nuc_i,\nuc_i	)-\rho(\nuc_i,\nuc_l)
< 0
\qquad\text{and}\qquad
\rho_{\Seg(\nuc_i,p_1)}^{\nuc_i,\nuc_l}(1)
=\rho(p_1,\nuc_i	)-\rho(p_1,\nuc_l)
> 0
.
\end{align*}
By the Intermediate Value Theorem the  function $\rho_{\Seg(\nuc_i,p_1)}^{\nuc_i,\nuc_l}$ has a root  $0<t''< 1$.
Let $$x'':=(1-t'')\cdot \nuc_i+{t''}\cdot p_1 \in\Seg(\nuc_i,p_1)\setminus\{p_1\}.$$
So,
$$\rho(\nuc_l,x'')=\rho(\nuc_i,x'')<\rho(\nuc_i, p_1).$$
From the above equation, and by $l<i$, $x''\in \Seg(\nuc_i,p_1)$ satisfies the property beneath the ``argmin" in the definition of $p_1$ in \eqref{def:p_1}, and it is also closer to $\nuc_i$ than $p_1$ does. This is a contradiction to the definition of $p_1$.
Consequently, \eqref{eq:1-2NN of p1 are nuc i nuc j} is satisfied. 

%\todo{In FIgure 2 bold the circles more, bold (a,b,c,d) and erase newlines.}

Now, define
\begin{align}
\label{def:ve}
\ve:&=\min(\{\rho(p_1,\nuc_l)-r_1
:l\in[m]\setminus\{i,j\}\}\ \cup\ \{r_1\})
%\\&=\begin{cases}
%\min\limits_{l\in[m]\setminus\{i,j\}}\rho(X'_l,\Ball{p_1}{r_1}),\quad &\text{if }[m]\setminus\{i,j\}\neq\emptyset
%\\
%r_1,&\text{else.}
%\end{cases}
\end{align}
(see Figure \ref{fig:forall i exists j}%
%{\getrefnumber{fig:forall i exists j}(c)}
).
Due to \eqref{eq:1-2NN of p1 are nuc i nuc j}, $0<\ve\leq r_1$.
Set
\begin{align*}
%\label{def:p2}
p_2:=\left(1-\frac{\ve/3}{r_1}\right)\cdot p_1+\frac{\ve/3}{r_1}\cdot \nuc_j
\in\Seg(\nuc_j,p_1)\setminus\{\nuc_j\}
.
\end{align*} 
Note that $p_2$ is the point in $\Seg(p_1,\nuc_j)$ with a distance $\ve/3$ from $p_1$.
After we will show that $\Xinn{1}(p_2;\bX')=\nuc_j$ and $\Xinn{2}(p_2;\bX')=\nuc_i$, we will immediately get that $\nuc_i\in\NNm(\nuc_j,\bX')$. This will conclude the proof.
To this end, it suffices to show both
\begin{align}
\rho(p_2,\nuc_i)<\rho(p_2,\nuc_l),
\qquad \forall l\in[m]\setminus\{i,j\}
\label{eq: 2NN of p2 is nuc i}
\end{align}
and
\begin{align}
\label{eq: 1-NN of p2 is nuc j}
\rho(p_2,\nuc_j)<\rho(p_2,\nuc_i)
.
\end{align}

To see \eqref{eq: 2NN of p2 is nuc i}, note that for any $l\in[m]\setminus\{i,j\}$,
\begin{align}
\rho(p_2,\nuc_i)	\nonumber
&\leq \rho(p_2,p_1)+\rho(p_1,\nuc_i)\nonumber
\\&=\ve/3+r_1	\nonumber
\\&<(r_1+\ve)-\ve/3	\nonumber
\\&\leq \rho(p_1,\nuc_l)-\rho(p_1,p_2)	\label{Tran:1}
\\&\leq \rho(p_2,\nuc_l)	\label{Tran:2}
.
\end{align}
\eqref{Tran:1} follows the definition of $\ve$ in \eqref{def:ve} and \eqref{Tran:2} is by the triangle inequality.
%In \eqref{tr:1} we used the definition of $p_2$ in \eqref{def:p2}. \eqref{tr:2} is by the definition of $\ve$ in \eqref{def:p2}
To see \eqref{eq: 1-NN of p2 is nuc j}, suppose towards a contradiction that $\rho(p_2,\nuc_i)\leq \rho(p_2,\nuc_j)$. Then,
\begin{align}
\rho(p_1,p_2)+\rho(p_2,\nuc_i)
&\leq \rho(p_1,p_2)+\rho(p_2,\nuc_j)\nonumber
\\&=\rho(p_1,\nuc_j)	\label{tr:3}
\\&=\rho(p_1,\nuc_i)\nonumber
\\&\leq \rho(p_1,p_2)+\rho(p_2,\nuc_i)\label{tr:4}
.
\end{align}
\eqref{tr:3} is explained by $p_2\in\Seg(p_1,\nuc_j)$, and by the fact that $\rho$ is the Euclidean metric.
As we both lower- and upper-bounded $\rho(p_1,\nuc_i)$ with $\rho(p_1,p_2)+\rho(p_2,\nuc_i)$ in \eqref{tr:4}, these two terms are equal. Due to the fact that $\rho$ is the Euclidean metric and by the geometry of $\R^d$, this implies
\begin{align}
\label{inc:1 2 i}
p_2\in\Seg(p_1,\nuc_i)
\setminus\{p_1,\nuc_i\}
.
\end{align}
Note that $p_2$ must be distinct from $\nuc_i$ because their distances to $p_1$ are different.

Let $\ell\subseteq\R^d$ the line determined by $p_1$ and $\nuc_i$. By $p_2\in\Seg(p_1,\nuc_i)$, $p_2\in\ell$. By $p_2\in\Seg(p_1,\nuc_j)$, $\nuc_j\in\ell$.
By the definition of $p_1$ in \eqref{def:p_1}, $\nuc_i$ and $\nuc_j$ are equidistant to $p_1$. Hence, and since $p_1, \nuc_i$ and $\nuc_j$ are co-linear, $p_1=(\nuc_i+\nuc_j)/2$.
So, 
\begin{align}
p_1\in\Seg(\nuc_i,\nuc_j)\setminus\{\nuc_i,\nuc_j\}
.
\label{inc: i 1 j}
\end{align}
Informally, from \eqref{inc:1 2 i} and \eqref{inc: i 1 j}, the discussed points must lie in $\ell$ ``in the undirected order"\footnote{To put it formally, we say that a $k$-tuple $(a_1,\dots,a_k)\in\X^k$, $k\in\mathbb N$, is in \textit{an undirected order} if $a_s\in\Seg(a_{s-1},a_{s+1})$, for all $2\leq s\leq k-1$.} \[(\nuc_i, p_2,p_1,\nuc_j),\] but this is a contradiction to $p_2\in \Seg(p_1,\nuc_j)\setminus\{p_1,\nuc_j\}$.
Thus, \eqref{eq: 1-NN of p2 is nuc j} is satisfied, and the proof is concluded.
\qed

\sectionmark{Proof of Theorem 3}{}
\section{Proof of Theorem \ref{thm:comp_rate}}
\label{sec:comp_rate_proof}
\sectionmark{Proof of Theorem 3}{}
%\section{ PROOF OF THEOREM \ref{thm:comp_rate}}
%\label{sec:comp_rate_proof}
\label{sup-sec3}
%\begin{theorem}
%Assume that the marginal distribution $\mu$ of $X$ has a density $f$ that satisfies the {minimal mass condition} and 
%is bounded away from zero (\textbf{SDC}) and that $\supp(\mu)$ is a convex set.
%%the tail condition.
%%$f$ is bounded away from zero.
%%the distribution function $H_x(\cdot)$ is continuous for each $x$.
%If the margin condition is satisfied with $0<\alpha \leq 1$, the geometric margin condition is satisfied with $\gm>0$, and the
%%generalized
%H\"older continuity condition is met with parameters $0<\beta\leq 1$ and $C>0$,
%then for any $0<\gamma\leq\gamma_0$,
%\begin{align*}
%\frac{\EXP\left[|\tD(\gamma)|\right]}{n} \leq \dots
%\end{align*}
%\end{theorem}
%\todoii{Should annotate $\tD=(\tbX,\tbY)$ or $\tD(\gamma)=(\tbX(\gamma),tbY(\gamma))$?}

\noindent
%\textit{Proof.}
%\todoii{change $s$ to be summand instead of multiple. Omer: done.}
%\textcolor{violet}{remove: [Let $0<\gamma\leq \gamma_0$ where $\gamma_0$ is as in Definition \ref{def:MMA}.]}
%\des{Maybe remove ``where $\gamma_0$ is as in Definition \ref{def:MMA}" because $\gamma_0$ is defined in the theorem.}
Let $t>0$
%\todo{If we use $t<1$ fix it in the statement, else, fix it in the supplementary.}
%and $ 8\gamma\leq s\leq 8\gamma_0$ that will be determined later.
and $s=16\gamma$.
%\inform{Roi: Currently $s$ is not determined later. I fixed $s=16\gamma/3$ and $c=8$.
%\\Omer: We need $8\gamma\leq s$ so I it changed to $s=16\gamma$.}
%\textcolor{violet}{remove: Let $(\tXg(\gamma),\tYg(\gamma))=\tD(\gamma)$.}
We decompose
%\todo{Why not $\boundary{t+5\gamma}$?}
\begin{align*}
\EXP\left\{|\tbXg(\gamma)|\right\} 
=
\EXP\left\{|\tbXg(\gamma) \cap \boundary{t+3s} |\right\}
 + \EXP\left\{|\tbXg(\gamma)\setminust  \boundary{t+3s}|\right\}
\leq
\CN_\gamma(\boundary{t+3s})
+ 
\EXP\left[|\tbXg(\gamma) \setminust \boundary{t+3s}|\right]
.
\end{align*}
We say that $S'\subseteq \X$ is an \textit{$s$-cover} of $S\subseteq\X$ if for any $x\in S$ there exists $x'\in S'$ such that $\rho(x,x')<s$.
For any $u\geq 0$, let
\[
\Bboundary{u}
=
\{x\in\X:\delta(x)=u\}
\]
the topological boundary of $\boundary{u}$.
%As in \eqref{eq:comp_set}, let $\bXg(\gamma)$ be a $\gamma$-net of $\bX'_m$ and let $\bYg(\gamma)$ be the corresponding labels as computed by OptiNet.
%\todo{Should we redefine $\bXg(\gamma)$ and $\CN_\gamma$?}
Define the event
\begin{align*}
A :&=\{\bXg(\gamma)\text{ is an $s$-cover of $\C$}\}
,
\end{align*}
where
\begin{align*}
\C:=\Bboundary{t}\cup\Bboundary{t+2s}
.
\end{align*}
Then
\begin{align*}
\EXP[|\tbXg(\gamma) \setminust \boundary{t+3s}|]
&\leq
\PROB\left[\neg A \right] 
\cdot
\CN_\gamma(\supp(\mu) \setminust \boundary{t+3s})
+ \EXP\left[ |\tbXg(\gamma) \setminust \boundary{t+3s}|\mid A\right]
.
\end{align*}
Define also the event
\begin{align*}
B &:=\{\forall X'\in \bXg(\gamma)\setminust \boundary{t+s}, Y'(X')=
Y^*(X')
%\sign(\eta(X'))
\}
,
\end{align*}
where 
\begin{align*}
Y^*(x) = \argmax_{j\in\Y} P_j(x)
\end{align*}
is the optimal label for $x\in\X$.
Then,
\begin{align*}
&\EXP\left[|\tbXg(\gamma) \setminust \boundary{t+3s}| \mid A\right]
\\&\leq
\PROB\left[\neg B \mid A \right] 
\cdot
\CN_\gamma(\supp(\mu) \setminust \boundary{t+3s})
+ \EXP\left[ |\tbXg(\gamma) \setminust \boundary{t+3s}|\mid A, B\right]
.
\end{align*}
We show below that
%\todo{removed extra $\gamma$ (also several times below)}
\begin{align}
\label{eq:zero_AB}
\EXP\left[ |\tbXg(\gamma) \setminust \boundary{t+3s}|\mid A, B\right] = 0.    
\end{align}
So
\begin{align*}
%&
\EXP[|\tbXg(\gamma)|] 
&\leq
\CN_\gamma(\boundary{t+3s}) +
\CN_\gamma(\supp(\mu) \setminust \boundary{t+3s})
%\CN_\gamma(\supp(\mu))
\cdot\left(\PROB\left[\neg A\right] + \PROB\left[\neg B \mid A\right]\right)
.
\end{align*}

To bound $\PROB\left[\neg A\right]$ 
note that
\begin{align}
\neg A
&=\{\bXg(\gamma)\text{ is not an $s$-cover of }\C\}	\nonumber
	\\&=\{\exists x\in\C,\rho(x,\bXg(\gamma))\geq s \}	\nonumber
	\\&=\{\exists x\in \C,\Ball{x}{s }\cap\bXg(\gamma)=\emptyset\}	\nonumber	
	.
\end{align}
%\todo{make the following paragraph more elegant.Omer: Done}
Let
\begin{align*}
\Cp
:=\{x\in\X:\rho(x,\C)<s/4\}
\end{align*}
the open $s/4$-envelope around $\C$.
%Let $\Cp:=\boundary{t+2.5s}\setminust \clBound{t+ 1.5s}\cup \boundary{t+0.5s}\setminust \clBound{t- 0.5s}$.
%\inform{Assuming $0.5s\leq t$?
%\\
%Omer: I don't think it is necessary.}
Fix $\bG\subseteq \C$ to be an $s/4$-net of $\C$,
and let $\Vor_{\bG}$ be the Voronoi partition of $\Cp$ induced by the prototypes in $\bG$.
For any $x\in \Cp$, let $G_x$ be the Voronoi cell from $\Vor_{\bG}$ in which $x$ resides, and let $g_x\in \bG$ be the corresponding prototype.
%\inform{Corrected the calculations here.}
To see that $G_x\subseteq \Ball{x}{s}$ for all $x\in\C$, consider some $x_1\in G_x\subseteq \Cp$.
Let $x_2\in\C$ such that $\rho(x_1,x_2)<s/4$. Then,
\begin{align*}
\rho(x,x_1)
&\leq \rho(x,g_x)+\rho(x_1,g_x)
<s/4+\rho(x_1,g_{x_2})
\\&\leq s/4+\rho(x_1,x_2)+\rho(x_2,g_{x_2})
<3\cdot s/4
<s
.
\end{align*}
%Since $\bG$ is an $s/2$-net of $\C$, 
%for any $x\in\C$ and for any $x'\in G_x$,
%\begin{align*}
%\rho(x',x)
%\leq \rho(x',g_x)+\rho(g_x,x)<s/2+s/2=s.
%\end{align*}
%So, $G_x\subseteq \Ball{x}{s}$.
It follows that
\begin{align*}
	\nonumber
	\neg A
%	&=\{\bX'(\gamma)\text{ is not a $s\gamma$-cover of }\boundary{t+3s\gamma}\setminust \clBound{t}\}	\nonumber
%	\\&=\{\exists x\in \boundary{t+3s\gamma}\setminust \clBound{t},\rho(x,\bX'(\gamma))\geq s \gamma\}	\nonumber
%	\\&=\{\exists x\in \boundary{t+3s\gamma}\setminust \clBound{t},\Ball{x}{s \gamma}\cap\bX'(\gamma)=\emptyset\}	\nonumber
	&\subseteq 
	\{\exists g\in\bG ,G_g\cap\bXg(\gamma)=\emptyset\}	\nonumber	
%	\\&\subseteq \{\exists g\in\bG ,\Ball{g}{s\gamma/4}\cap\Xg(\gamma)=\emptyset\}	
%\label{t2}
	.
\end{align*}

In addition, $\Ball{g}{s/8}\subseteq G_g$ for all $g\in\bG$.
Indeed, if for some $x'\in \Ball{g}{s/8}$ it holds that $g' := \Xinn{1}(x';\bG)\neq g$, then it must hold that
\[
\rho(g,g') \leq \rho(g,x') + \rho(x',g') \leq  2\rho(x',g) < s/4,  
\]
which is in contradiction to the fact that any two prototypes in the $s/4$-net $\bG$ must have distance at least $s/4$.
% the packing property of $\bG$ as a $s\gamma/2$-net.
%for any 
%$x'\in \sp{g,s\gamma/4}$,
%$$
%\rho(x', \Xinn{1}(x',\bG))\leq\rho(x',g)<s\gamma/4,
%$$
%%so if $\Xinn{1}(x',\bG) \neq g$, 
%
%it must be that
%$$
%\rho(g,\Xinn{1}(x',\bG))\leq \rho(g,x')+\rho(x',\Xinn{1}(x',\bG))<\frac{s\gamma}4+\frac{s\gamma}4=\frac{s\gamma}2.
%$$
%Thus, by the definition of a $s\gamma/2$-net, $\Xinn{1}(x',\bG)=g$.
It follows that
\begin{align*}
\neg A
&\subseteq \{\exists g\in\bG ,\Ball{g}{s/8}\cap\bXg(\gamma)=\emptyset\}
.
\end{align*}
Recall that $\gamma= s/16$.
So, in order to see that
\begin{align*}
\neg A
&\subseteq \{\exists g\in\bG ,\Ball{g}{s/16}\cap\Xul=\emptyset\}
,
\end{align*}
%\begin{align*}
%\neg A
%&\subseteq \{\exists g\in\bG ,\Ball{g}{\gamma}\cap\Xul=\emptyset\}
%,
%\end{align*}
%\inform{Changed the radius of the above ball.}
note, that if the event $\{\exists g\in\bG ,\Ball{g}{s/8}\cap\bXg(\gamma)=\emptyset\}$ occurs, and $X'_i\in \Ball{g}{s/16}\cap \Xul$ for some $g\in\bG$, then, for all $j\in[| \bXg(\gamma)|]$,
\begin{align*}
\rho(X_j(\gamma),X'_i)
&\geq \rho(g,X_j(\gamma))-\rho(g,X'_i)
\\&>s/8-s/16
=\gamma
.
\end{align*}
Then, $\bXg(\gamma) \cap \Ball{X'_i}{\gamma}=\emptyset$, and thus, $\bXg(\gamma)\uplus\{X'_i\}\subseteq\Xul$ is a $\gamma$-net. This is a contradiction to the maximality of $\bXg(\gamma)$ as a $\gamma$-net.
Therefore, using a union bound,
\begin{align*}
\PROB\{\neg A\}
\leq \sum_{g\in\bG}\PROB\{\Ball{g}{s/16}\cap\Xul=\emptyset\}
.
\end{align*}

By the MMA condition (\ref{MMA}), by $s/16=\gamma\leq \gamma_0$, and by the BAZ condition (\ref{BAFZ}), we have that for all $g\in\bG$,
\begin{align*}
\mu(\Ball{g}{s/16})
\geq \kappa \dlb(s/16)^d 
.
\end{align*}
Thus,
%\todoii{Missing argument. Below is a bound on the probability that no sample from $\bX_m'$ fell into $\sp{g,\frac{s}4}$. But above we need to bound the probability that no point from $\bX(\gamma)$ fell into $\sp{g,\frac{s}4}$.
%If we take $s/4\geq 2\gamma$,
%%I think that for an appropriate $c>0$ (find such $c$),
%%$$
%%\{\sp{g,s/4}\cap\Xg(\gamma)=\emptyset\} \subseteq \{\sp{g,s/4
%%+\textcolor{blue}{c\gamma}
%%}\cap\bX_m'=\emptyset\}
%%$$
%$$
%\{\sp{g,s/4}\cap\Xg(\gamma)=\emptyset\} \subseteq \{\sp{g,s/8}\cap\bX_m'=\emptyset\}
%$$
%}
\begin{align*}
\PROB\{\neg A\}
&\leq |\bG|\cdot (1-\kappa\dlb(s/16)^d)^m
\leq \CN_{s/4}(\C)\cdot\exp(-m\cdot \kappa\dlb(s/16)^d).
\end{align*}

As for $\PROB[\neg B\mid A]$, by the law of total expectation, followed by a union bound,
\begin{align}
\nonumber
\PROB[\neg B \mid A] 
& =
\EXP[\PROB[\neg B \mid A, \bXg(\gamma)]]
\\
\label{eq:PnotB}
%\label{eq:PnotB}
& \leq 
\EXP\left\{
\sum_{X'\in \bXg(\gamma)\setminust \boundary{t+s}} \PROB[Y'(X') \neq 
Y^*(X')
%\sign(\eta(X')) 
\mid  A, \bXg(\gamma)]
\right\}
.
\end{align}
%To bound the summands in \eqref{eq:PnotB}, we first establish the following lemma, whose proof is given at the end of the section.
%
%\begin{lemma}
%\label{lem:unif_label}
%Under the event $A$,
%\todo{and the assumption that $\supp(\mu)$ is a convex set?}
%the geometric margin condition \eqref{cond:Blas} implies that 
%with probability one, for $0< t\leq {c_1}^{1/\gm}$
%\todo{why upper bound on $t$ needed?}
% and all $X'\in \Xg(\gamma)\setminust \boundary{t+s\gamma}$,
%\begin{align*}
%Y^*(x) = Y^*(X'), \qquad \forall x\in V_{X'} \cap \UB{\gamma},
%\end{align*}
%such that
%\begin{align*}
%\reg(x) \geq 
%%\EXP\left\{\reg(X)\right\}
%%\int_{V_{X'}\cap \UB{\gamma}} \reg(x) \mu(dx) \geq 
%\frac{t^{\gm}}{c_1} > 0
%, \qquad \forall x\in V_{X'} \cap \UB{\gamma}.
%\end{align*}
%%namely, there is no approximation error.
%\end{lemma}
For any $X'\in\bXg(\gamma)\setminust \boundary{t+s}$, abbreviating $Y^*=Y^*(X')$ and note that
\begin{align}
& \quad \,\, \PROB[Y'(X') \neq Y^* \mid  A, \bXg(\gamma)]	\nonumber
\\
& \leq \PROB\left\{
\exists j\in \Y\setminust\{Y^*\},
P_{n,j}(X') \geq P_{n,Y^*}(X') \mid  A, \bXg(\gamma)
\right\}	\label{eq:prob for different Y'}
%\\
%& \leq \sum_{j\in \Y\setminus\{Y^*(X')\}}
%\PROB\left\{P_{n,j}(X')-P_{n,Y^*(X')}(X') \geq 0 \mid  A, \Xg(\gamma)\right\}
.
%\\
%&\leq
%\EXP\left\{ \IND_{\{\argmax_{j \in \Y} P_{n,j}(X') \neq Y^*(X') \right\}\}}
%\\
%& = 
\end{align}
Similarly to \eqref{bgest} in the proof of Theorem \ref{thm:comp_rate}, for all $x\in\X$ abbreviate $V_x:=V_x(\bXg(\gamma))$ and for every $j\in\Y$ put 
$\bar P_{n,j}(x):=\frac{\int_{V_x} P_j(z)\mu(dz)}{\mu(V_x) }$.
%Thus, recalling the definition $\bar P_{n,j}(x)=\frac{\int_{V_x\cap \UB{\gamma}} P_j(z)\mu(dz)}{\mu(V_x\cap \UB{\gamma}) }$
%in \eqref{bgest},
We have that, for all $j\in[M]\setminus\{Y^*\}$,
%\todo{$j\in \Y\setminust\{Y^*(X')\}$?}
\begin{align}
&\quad\,\,\{P_{n,j}(X') - P_{n,Y^*}(X') \geq 0\}         \nonumber
\\& = \{P_{n,j}(X') - \bar P_{n,j}(X') + \bar P_{n,Y^*}(X') - P_{n,Y^*}(X')
\geq \bar P_{n,Y^*}(X') - \bar P_{n,j}(X')\} 		
\label{inc:pre gm}
.
\end{align}
We claim that given $A$, the geometric margin condition \textup{(\textbf{GMC})} 
%that is given in Definition \ref{def:geom_margin}
implies
\begin{align*}
%\label{gm usage}
\bar P_{n,Y^*}(X') - \bar P_{n,j}(X')
\geq c_1 t^\gm
,\qquad\forall j\in\Y\setminus\{Y^*\}
.
\end{align*}
To see this,
%\todoii{Better to reorder the proof here and instead of the last line above write something along the line: We claim that by the geometric margin condition, [*the equation below Eq.(51)*]. To see this, ...}
we first establish the following lemma, whose proof is given at the end of the section.
%\textcolor{violet}{For any $X'\in \Xg(\gamma)$, let $V_{X'}$ the cell containing $X'$ in the Voronoi partition induced by $\Xg(\gamma)$.}\des{Should we denote it $V_{X'}(\Xg(\gamma))$ instead of $V_{X'}$ as in the main paper?}
\begin{lemma}
\label{lem:VX}
Assume that $A$ occurs.
\begin{description}
\item[(a)]
For all $X'\in\bXg(\gamma)\setminus\boundary{t+s}$,
\begin{align*}
\delta(x)>t
\quad\text{and}\quad
Y^*(x)=Y^*(X')
,
\qquad\qquad\quad
 \forall x\in V_{X'}
.
\end{align*}
\item[(b)]
For all $X'\in\bXg(\gamma)\setminus\boundary{t+3s}$,
\begin{align*}
&\delta(Z')>t+s
\quad
\text{and}
\quad
Y^*(Z')=Y^*(X')
,
\qquad\qquad\quad
\forall
Z'\in\NNm(X',\bXg(\gamma))
.
\end{align*}
\end{description}
\end{lemma}

%\todo{any metric is continuous}
By Lemma 
\ref{lem:VX}
%{\getrefnumber{lem:VX}(a)}
and by the Geometric Margin Condition, 
%\closed{I changed the place of $c_1$ in the definition of GMC --- its definition in the main text didn't make sense. Omer: Fixed it here.}
%for any $j\in\Y\setminus\{Y^*\}$,
\begin{align*}
\bar P_{n,Y^*}(X') -\bar P_{n,j}(X')
\geq \cdot\min\{ c_1 t^\gm,1\}
\geq c_1 t^\gm
,\qquad\forall j\in\Y\setminus\{Y^*\}
.
\end{align*}
Thus, by \eqref{inc:pre gm}, under event $A$,
\begin{align*}
&\quad\,\,\{P_{n,j}(X') - P_{n,Y^*}(X') \geq 0\} 
\\
& \subseteq \{P_{n,j}(X') - \bar P_{n,j}(X') + \bar P_{n,Y^*}(X') - P_{n,Y^*}(X')
\geq c_1 t^{\gm}\} 
.
\end{align*}
So, assuming $A$,
\begin{align*}
&\quad\,\,\left\{
\exists j\in \Y\setminust\{Y^*\},
P_{n,j}(X') \geq P_{n,Y^*}(X')
%\mid  A, \Xg(\gamma)
\right\} 
\\&
\subseteq 
%\bigcup_{j\in\Y} 
\left\{
\sum_{j\in\Y}| P_{n,j}(X') - \bar P_{n,j}(X')| \geq c_1 t^{\gm}
%\mid  A, \Xg(\gamma)
\right\} 
\\&
\subseteq 
\bigcup_{j\in\Y} 
\left\{
| P_{n,j}(X') - \bar P_{n,j}(X')| \geq \frac{c_1 t^{\gm}}{M}		
%\mid  A, \Xg(\gamma)
\right\} 
.
\end{align*}
From \eqref{eq:prob for different Y'}, as in the proof of Theorem \ref{thm}, following a union bound and the Bernstein inequality,
\begin{align*}
& \quad \,\, \PROB[Y'(X') \neq Y^*(X') \mid  A, \bXg(\gamma)]
\\
& \leq
\sum_{j\in\Y} \PROB\left\{
| P_{n,j}(X') - \bar P_{n,j}(X')| \geq \frac{c_1t^{\gm}}{ M}
 \mid  A, \bXg(\gamma)\right\}
\\
&\leq
2 M \exp\left(- \frac{C n  t^{2\gm} \mu(V_{X'})}{M}  \right).
%\\
%&\leq
%2 M \exp\left(- \frac{C n  t^{2\gm} \gamma^d}{M}  \right),
\end{align*}
To lower bound $\mu(V_{X'})$, recall that $\bX(\gamma)$ is a $\gamma$-net, and so, as in \eqref{SinV}, $V_{X'}\supseteq \Ball{X'}{\gamma/2}$.
%\todo{Why here $\gamma/2$ but in \eqref{SinV} $\gamma/3$? Omer: Changed  \eqref{SinV} to $/2$.}
%Let $x\in \sp{X',\gamma/2}$. We have that $\rho(x,\Xinn{1}(x,\bX'(\gamma)))\leq\rho(x,X')<\gamma/2$. By the triangle inequality we then have $\rho(X',\Xinn{1}(x,\bX'(\gamma)))<\gamma$. So, it must have that $X'=\Xinn{1}(x,\bX'(\gamma))$. The above inclusion is than achieved,
By the \textup{\textbf{SDC}} and by $\gamma/2<\gamma_0$,
$\mu(V_{X'}) \geq \kappa \dlb(\gamma/2)^d$.
Thus,
\begin{align*}
\PROB[Y'(X') \neq Y^*(X') \mid  A, \bXg(\gamma)]
&\leq
2 M \exp\left(- C n  t^{2\gm} \gamma^d  \right).
%\\
%&\leq
%2 M \exp\left(- \frac{C n  t^{2\gm} \gamma^d}{M}  \right),
\end{align*}
Putting this in \eqref{eq:PnotB},
\begin{align*}
\PROB[\neg B \mid A] 
& \leq     2 M \cdot \CN_\gamma(\supp(\mu)\setminust\boundary{t+s}) \cdot\exp\left(-C n t^{2\gm} \gamma^d\right)
.
\end{align*}

We thus conclude,
\begin{align*}
\EXP[|\tbXg(\gamma)|]
&  \leq \CN_\gamma(\boundary{t+3s})
+\CN_\gamma(\supp(\mu)\setminus \boundary{t+3s  })\times
\\&\quad
\left (\CN_{4\gamma}(\Cp) e^{-m\cdot Cs^d}
+2M\cdot \CN_\gamma(\supp(\mu)\setminus\boundary{t+s}) e^{-n\cdot C t^{2\gm}\gamma^d}\right )
,
\end{align*}
%\gap{The covering number is with $4\gamma$, not with $\gamma$ as in the main file.}
%\EXP[|\tXg(\gamma)|] &  \leq \NNm_\gamma(\boundary{t+3s\gamma})
%\\&+\NNm_\gamma(\supp(\mu)\setminust\boundary{t+3s\gamma})
%\\&\qquad \cdot\left (	\NNm_\gamma(\boundary{t+3s\gamma}\setminust\boundary{t})e^{-m\cdot C(s\gamma/4-\gamma)^d}+2\NNm_\gamma(\supp(\mu)\setminust\boundary{t+s\gamma})\cdot e^{-n\cdot Ct^{2\gm}(\gamma/2)^d}	\right ).
%\end{align*}
%Setting $t=???$ and $s=5$,
%\begin{align*}
%\EXP[|\tbX'(\gamma)|] &  
%\leq ???
%,
%\end{align*}
concluding the proof of the Theorem.

We are left to prove \eqref{eq:zero_AB},
namely,
$$
\EXP\left[ |\tbXg(\gamma) \setminust \boundary{t+3s}|\mid A, B\right] = 0.   $$
%Assume that $A$ and $B$ occur. 
%By the definition of $\tD(\gamma)=(\tbX(\gamma),\tbY(\gamma))$
To show this, we show that under the events $A$ and $B$, 
%We need to show that 
for every $X'\in\bXg(\gamma)\setminust \boundary{t+3s}$,
\begin{align*}
%\label{eq:same labels of neighbours}
Y'(X')=Y'(Z'),\qquad \forall Z'\in\NNm(X',\bX(\gamma)).
\end{align*}
Then, 
by the definition of $\tD(\gamma)=(\tbX(\gamma),\tbY(\gamma))$ in \eqref{def:tD}, 
$X'\notin  \tbXg(\gamma)$ and \eqref{eq:zero_AB} follows.
%\todoii{Missing argument to why $Y'(X')=Y'(Z')$. Omer: done.}

%\textcolor{blue}{To show \eqref{eq:same labels of neighbours}, we first show that the event $A$ implies that $V(X',\Xg(\gamma))\subseteq \X\setminus\clBound{t+2s}$.
%This, together with the GMC and the continuity of $\eta$, will imply that $Y^*(x) = Y^*(X')$ for all $x \in V(X',\Xg(\gamma))$. We then similarly establish that $V(Z',\Xg(\gamma))\subseteq \X\setminus\clBound{t+s}$ and that $Y^*(z) = Y^*(Z') = Y^*(X')$ for all $z \in V(Z',\Xg(\gamma))$. The event $B$ will then imply \eqref{eq:same labels of neighbours}.}
%\inform{The structure of the proof changed so I think to remove this paragraph.}
%\inform{If we delete this then delete also the label of \eqref{eq:same labels of neighbours}.}

Indeed, assume that $A$ and $B$ occurred.
Let $X'\in\bXg(\gamma)\setminust \boundary{t+3s}$ and $Z'\in\NNm(X',\bX(\gamma))$.
By Lemma
\ref{lem:VX},
%{\getrefnumber{lem:VX}(b)}, 
$Y^*(X')=Y^*(Z')$ and $\delta(Z')\geq t+s$.
So, $X',Z'\in \bXg(\gamma)\setminus\clBound{t+s}$.
Hence, by the assumption on $B$,
\begin{align*}
Y'(X')=Y^*(X')=Y^*(Z')=Y'(Z')
.
\end{align*}
\qed

\subsection{Proof of Lemma \ref{lem:VX}}
To prove Lemma \ref{lem:VX}, we first state Claim \ref{clm:exist closer nuc} and Claim \ref{clm:dif labels} and prove them.
%In the proof of Claim \ref{clm:dif labels} we use Lemma \ref{lem:g continuous}, which its own proof is brought in the following subsection.

For any $x\in\X$ and $S\subseteq \X$ define $\rho(x,S):=\rho(S,x):=\inf_{x'\in S}\rho(x,x')$ where a infimum over the empty set is defined to be $\infty$.
\begin{claim}
\label{clm:exist closer nuc}
Assume $A$ occurs.
Let $k\in\{0, 2\}$.
For all $X'\in \bXg(\gamma)$ and $x\in\X$, if
\begin{align*}
\rho(X',\Bboundary{t+ks})\geq s
\quad\text{and}\quad
\Seg(X',x)\cap\Bboundary{t+ks}\neq\emptyset
\end{align*}
then 
$$
\exists Z'\in\bXg(\gamma)\setminus\{X'\}
\quad
\text{s.t.}
\quad
(
\rho(x,Z')<\rho(x,X')
\quad\text{and}\quad
\rho(Z',\C)<s
)
.
$$
\end{claim}
%\textit{Proof of Claim \ref{clm:exist closer nuc}.}
Let $x'\in\Seg(X',x)\cap\Bboundary{t+ks}$ and define $Z':=\Xinn{1}(x';\bXg(\gamma))$.
By $x'\in\C$ and by the assumption that $A$ occurs,
\begin{align*}
\rho(\C,Z')
\leq \rho(x',Z')
<s
.
\end{align*} 
So,
\begin{align}
\rho(x,Z')
&\leq \rho(x,x')+\rho(x', Z')	\nonumber
\\&< \rho(x,x')+s		\nonumber
\\&\leq \rho(x,x')+\rho(\Bboundary{t+ks},X')		\nonumber
\\&\leq\rho(x,x')+\rho(x',X')		\nonumber
\\&=\rho(x,X')	\label{eq:x' in Seg x X'}
,
\end{align}
where \eqref{eq:x' in Seg x X'} is due to the fact that $x'\in\Seg(X',x)$ and $\rho$ is the Euclidean metric.
This proves Claim \ref{clm:exist closer nuc}.

\begin{claim}
\label{clm:dif labels}
Let $x_1, x_2\in \X$.
If $Y^*(x_1)\neq Y^*(x_2)$, then, for any $0<r\leq \delta(x_1)$, there exists $x_3\in\Seg(x_1,x_2)$ such that $\delta(x_3)=r$.\footnote{Since $r$ is an arbitrary small positive number, then, by the continuity of $\delta$, standard calculus shows that Claim \ref{clm:dif labels} holds also for $r=0$.}
\end{claim}

%\textit{Proof of Claim \ref{clm:dif labels}.}
For the proof of this claim, we state the following lemma, whose proof is given at the end of the section:
\begin{lemma}
\label{lem:g continuous}
Assume that $\Y$ is a finite set of labels,
%\todo{any metric is continuous in the induced topological space generated by the open balls}
 and $P_j$ is continuous in $\X$, for all $j\in\Y$.
Assume that we are given some point $x_0\in\X$. Then, the function $\cf{x_0}:\X\to\R_{\geq0}$ that is defined according to the rule
\begin{align*}
%\label{def:cf X'}
x\mapsto \IND{Y^*(x)=Y^*(x_0)}\cdot \delta(x)
\end{align*}
is continuous in $\X$.
\end{lemma}
Adapting the notation of Lemma \ref{lem:g continuous},
since $Y^*(x_1)\neq Y^*(x_2)$,
\begin{align*}
\cf{x_1}(x_2)
\eqdef \delta(x_2)\cdot \IND{Y^*(x_1)= Y^*(x_2)}
=\delta(x_2)\cdot 0
\leq \delta(x_1)\cdot \IND{Y^*(x_1)=Y^*(x_1)}
\eqdef \cf{x_1}(x_1)
.
\end{align*}
By Lemma \ref{lem:g continuous} and by the Intermediate Value Theorem, there exists $x_3\in\Seg(x_1,x_2)$ with
\begin{align*}
0
<r
=\cf{x_1}(x_3)
=\delta(x_3)\cdot \IND{Y^*(x_3)=Y^*(x_1)}
.
\end{align*}
Hence, $\IND{Y^*(x_3)=Y^*(x_1)}=1$. Thus, $\delta(x_3)=r$, and Claim \ref{clm:dif labels} is proved.

%\des{Is this space here eligible?}
Now, for the proof of Lemma \ref{lem:VX}, let $k\in\{0,2\}$.
Let $X'\in\bXg(\gamma)\setminus\boundary{t+(k+1)s}$ and $x\in V_{X'}$.
Note that $\rho(X',\Bboundary{t})\geq (k+1)s$.
Suppose in contradiction that $\delta(x)\leq t+ks$.
Note that
\begin{align*}
\delta(x)
\leq t+ks
<t+(k+1)s
\leq \delta(X')
.
\end{align*}
By the continuity of $\delta$ and the Intermediate Value Theorem, there exists $x_4\in \Seg(x,X')\setminus\{X'\}$ with $\delta(x_4)=t+ks$.
That is, $\Seg(x,X')\cap\C\neq \emptyset$.
According to Claim \ref{clm:exist closer nuc}, there exists $P'\in\bXg(\gamma)\setminus\{X'\}$ with $\rho(x,P')<\rho(x,X')$.
This is a contradiction to $x\in V_{X'}$.
So,
\begin{align}
\label{eq:delta(x)>t+ks}
\delta(x)> t+ks
.
\end{align}

Suppose in contradiction that $Y^*(x)\neq Y^*(X')$.
According to Claim \ref{clm:dif labels}, since $\delta(X')\geq t+s$, there exists $x_5\in \Seg(x,X')$ such that $\delta(x_5)=t$.
That is, $\Seg(x,X')\cap\C\neq \emptyset$.
As in the previous paragraph, from Claim \ref{clm:exist closer nuc} there exists $Q'\in\bXg(\gamma)\setminus\{X'\}$ with $\rho(x,Q')<\rho(x,X')$, but this contradicts $x\in V_{X'}$.
So,
\begin{align}
\label{eq:Y* x=Y*X'}
Y^*(x)=Y^*(X')
.
\end{align}
Recall the generality of $k\in\{0,2\}$. By choosing
%putting
$k=0$, \eqref{eq:delta(x)>t+ks} and \eqref{eq:Y* x=Y*X'} prove Part \textbf{(a)} of Lemma \ref{lem:VX}.

As for Part \textbf{(b)}, let $X'\in\bXg(\gamma)\setminus\boundary{t+3s}$ and $Z'\in\NNm(X',\bXg(\gamma))$.
By the definition of $\NNm$ in \eqref{def:NNm}, there exists $x_6\in\X$ such that $X'$ and $Z'$ are the first- and second-NN of $x_6$ in $\bXg(\gamma)$, respectively.
The generality of $k$ and $x$ in \eqref{eq:delta(x)>t+ks} and \eqref{eq:Y* x=Y*X'} enables us to assign $k=2$ and $x=x_6$ and get
\begin{align*}
\delta(x_6)>t+2s
\quad
\text{and}
\quad
Y^*(x_6)=Y^*(X')
.
\end{align*}
In a similar manner as before, suppose in a contradiction that $\delta(Z')\leq t+s$.
Then,
\begin{align*}
\rho(Z',\Bboundary{t+2s})\geq s
.
\end{align*}
Note that
\begin{align*}
\delta(Z')
\leq t+s<t+2s<\delta(x_6)
.
\end{align*}
As previously done, using the continuity of $\delta$ in the Intermediate Value Theorem resulting in $\Seg(X', x_6)\cap \C\neq \emptyset$. Then, by Claim \ref{clm:exist closer nuc} there exists $W'\in\bXg(\gamma)\setminus\{Z'\}$ such that 
\begin{align}
\label{nuc:W'}
\rho(x_6,W')<\rho(x_6, Z')
\quad
\text{and}
\quad
\rho(W', \C)<s
.
\end{align}
Recall that $\delta(X')\geq t+3s$.
Then, trivially, $\rho(X', \C)\geq s$.
%So, by the continuity of $\rho$, $\rho(Z',\C)=\rho(Z',\clBound{t+2s})\geq s$.
Since $\rho(W',\C)<s$, $X'\neq W'$.
We found a nucleus $W'$ whose distinct from $X'=\Xinn{1}(x;\bXg(\gamma))$, and is also strictly closer to $x$ than $Z'$.
That is a contradiction to $Z'=\Xinn{2}(x_6;\bXg(\gamma))$.
Consequently, $\delta(Z')>t+s$.
%Now, as we did before in \eqref{eq:Y* x=Y*X'},
Again, suppose in a contrary that $Y^*(Z')\neq Y^*(x_6)$. Claim \ref{clm:dif labels} resulting in 
\[
\emptyset \neq \Seg(Z',x_6)\cap \Bboundary{t} \subseteq \Seg(Z',x_6)\cap\C
.
\]
By Claim \ref{clm:exist closer nuc} we get a nucleus $U'\in\bXg(\gamma)\setminus\{Z'\}$
with
\[
\rho(x_6,U')<\rho(x_6,Z')
\quad
\text{and}
\quad
\rho(U',\C)<s
,
\]
which, exactly as in \eqref{nuc:W'}, yields a contradiction.
%whose strictly closer to $x_6$ than $Z'$ does. 
%This is a contradiction to $Z'=\Xinn{2}(x_6,(\gamma))$.
So, $Y^*(Z')=Y^*(x_6)=Y^*(X')$.
\qed

\subsection{Proof of Lemma \ref{lem:g continuous}}
Let the function $\tcf{x_0}:\X\to\{0,1\}$ according to the rule
\[
x\mapsto\IND{Y^*(x)=Y^*(x_0)}
.
\]
Note that $\cf{x_0}=\tcf{x_0}\cdot \delta$.
Let $z\in\X$.
If $\tcf{x_0}$ is continuous in $z$, then, since the function $\delta$ is continuous in $\X$, $\cf{x_0}$ is continuous in $z$ and we are done.
Assume $z$ is an inconsistency point of $\tcf{x_0}$.
We will show that in that case, $\delta(z)=0$.
Indeed, since $\tcf{x_0}$ is not continuous in $z$, there exists a convergent sequence $(z''_i)_{i\in\mathbb N}\subseteq \X\setminus\{z\}$ to $z$, such that
\begin{align}
\label{eq:neq tcf}
\IND{Y^*(z''_i)=Y^*(x_0)}
=
\tcf{x_0}(z''_i)
\neq 
\tcf{x_0}(z)
=
\IND{Y^*(z)=Y^*(x_0)}
\end{align}
for infinitely many $i\in\N$.
Taking all the indices $i\in\mathbb N$ that satisfy \eqref{eq:neq tcf}, we get a subsequence
$(z'_i)_{i\in\mathbb N}\subseteq (z''_i)_{i\in\mathbb N}\subseteq \X\setminus\{z\}$ whose limit is $z$, such that
\begin{align*}
%z'_i \xrightarrow[i\to\infty]{}z \qquad \text{and}\qquad
Y^*(z'_i)\neq Y^*(z),\quad \forall i\in\N
.
\end{align*}

Recall that $\Y$ is a finite set. So, there must exists a label $j\in\Y\setminus\{Y^*(z)\}$ such that $Y^*(z'_i)=j$ for infinitely many indices $i\in\N$.
Taking all the indices $i\in\mathbb N$ with $Y^*(z'_i)=j$, there exists a subsequence $(z_i)_{i\in\N}\subseteq (z'_i)_{i\in\N}\subseteq\X\setminus\{z\}$ with a limit $z$, such that $Y^*(z_i)=j$, for all $i\in\N$.
By the continuity of the functions $P_{Y^*(z)}$ and $P_j$, and by $\lim\limits_{i\to\infty}z_i=z$, we have
\begin{align*}
0
&\leq (P_{Y^*(z)}-P_j)(z)
%\\ &
= (P_{Y^*(z)}-P_j)\left(\lim_{i\to\infty} z_i\right)
%\\ &
= \lim_{i\to\infty}(P_{Y^*(z)}-P_j)(z_i)
%\\ &
\le0
.
\end{align*}
Consequently,
\[
P_{Y^*(z)}(z)=P_j(z)
.
\]
Since $j\neq Y^*(z)$, we sure have \[P_{Y^*(z)}(z)=P_{(1)}(z)\geq P_{(2)}(z)\geq P_j(z)=P_{Y^*(z)}(z).\]
So $P_{(1)}(z)=P_{(2)}(z)$ and by the definition of $\delta$ in \eqref{def:delta}, $\delta(z)=0$.
%by the definition of $\clBound0$ in \eqref{def:clBound}, $z\in\clBound0$.
%So, $\delta(z)=0$.

Using the continuity definition in terms of $\ve_1$-$\ve_2$, let $\ve_1>0$.
By the continuity of $\delta$, there exists $\ve_2>0$ such that for any $z'\in \Ball{z}{\ve_2}$, $|\delta(z')-\delta(z)|<\ve_1$.
Then, for all $z'\in \Ball{z}{\ve_2}$,
\begin{align*}
&|\cf{x_0}(z')-\cf{x_0}(z)|
\\&= |\IND{Y^*(z')=Y^*(x_0)}\cdot \delta(z')- \IND{Y^*(z)=Y^*(x_0)}\cdot \delta(z)|
\\&= \IND{Y^*(z')=Y^*(x_0)}\cdot| \delta(z')|
\\&\leq |\delta(z')|
\\&=| \delta(z')- \delta(z)|
\\&<\ve_1.
\end{align*}
So, $\cf{x_0}$ is continuous in $z$ for this case as well.
\qed

\sectionmark{Proof of Theorem 3}{}
\part*{Failure of Theorem \ref{thm:comp_agree Lebesgue} in Non-Eucliden Spaces}
%\sectionmark{Failure of Theorem 2 in Non-Eucliden Spaces}{}

%\section{ FAILURE OF THEOREM \ref{thm:comp_agree Lebesgue} IN NON-EUCLIDEAN SPACE}
\label{sup-sec4}
As discussed in the
Conclusion section,
% (Section \ref{sec:conclusions}),
Theorem \ref{thm:comp_agree Lebesgue} can fail in non-Euclidean spaces, in the sense that removing all spurious prototypes simultaneously might lead to a classifier that is not consistent with the original one.
Consider for example the uniform distribution over $\X$, $\Y=\{0,1\}$, and $P_1(x)=\IND{x>0}$, the case of $\X=(-3,-1)\uplus(1,3)$ doesn't necessarily fulfil (\ref{equal labels}) with probability one (See Figure \ref{fig:unconnected space}):
\begin{figure}
\centering
%\ifthenelse{\equal\includefigs{1}}
%{
\begin{tikzpicture}
\coordinate [draw, label={below:\footnotesize{$-3$}}] (pm3) at (-3,0);
\coordinate [draw, label=below:\footnotesize{$-1$}] (pm1) at (-1,0);
\path[draw] (pm3) -- (pm1) coordinate[midway] (pm3Ppm1);

\coordinate [draw, label=below:\footnotesize{$3$}] (p3) at (3,0);
\coordinate [draw, label=below:\footnotesize{$1$}] (p1) at (1,0);
\path[draw] (p3) -- (p1) coordinate[midway] (p3Pp1);

\coordinate [draw, label=below:\footnotesize{$-2$}] (pm2) at (-2,0);
\coordinate [draw, label=below:\footnotesize{$2$}] (p2) at (2,0);

\path [draw, name path=qiPq1T]  ($(pm2)!2.5pt!270:(pm3)$) -- ($(pm2)!2.5pt!90:(pm3)$);
\path [draw, name path=qiPq1T]  ($(p2)!2.5pt!270:(p3)$) -- ($(p2)!2.5pt!90:(p3)$);

\path [name path = m1P1, color=gray, dashed, draw] (pm1) -- (p1) coordinate[midway] (m1P1);

\foreach \point in {pm1, pm3, p1, p3}
\draw [black, fill=white] (\point) circle (1.5pt);

\coordinate [draw, label={[align=center,yshift=3pt, xshift=-4pt]\footnotesize{$X'_1$}\\\scriptsize{label=$0$}}] (X1) at (-2.5,0);
\coordinate [draw, label={[align=center,yshift=3pt, xshift=4pt]\footnotesize{$X'_2$}\\\scriptsize{label=$0$}}] (X2) at (-1.5,0);
\coordinate [draw, label={[align=center,yshift=3pt]\footnotesize{$X'_3$}\\\scriptsize{label=$1$}}] (X3) at (2.5,0);

\foreach \point in {X1, X2, X3}
\fill [black] (\point) circle (2pt);

\end{tikzpicture}
%\caption{One can see that if $\X=(-3,-1)\uplus (1,3)$ and $\Xg(\gamma)=\{X'_1,X'_2,X'_3\}$ is as figured, then $\NNm(X'_1,\Xg(\gamma))=\{X'_2\}$, $\NNm(X'_2,\Xg(\gamma))=\{X'_1\}$ and $\NNm(X'_3,\Xg(\gamma))=\{X'_2\}$.
%Hence, $\tD=\{(X'_3,2)\}$, which induce a constant label $2$ on all $\X$.
%}
%}
%{\missingfigure{}}
\caption{Example of a failure of Theorem \ref{thm:comp_agree Lebesgue} in non-Euclidean space.
}
\label{fig:unconnected space}
\end{figure}
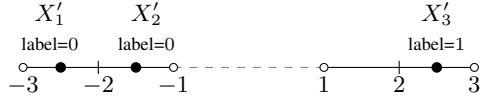
Suppose that $\D'=\{(X_i',Y_i'):i\in[3]\}$. The event $$\{-3<X_1'<-2<X_2'<-1\quad\text{ and }\quad 2<X_3'<3\}$$ occurs with a positive probability.
Since $\NNm(X_1',\D')=\{X_2\}$, $\NNm(X_2',\D')=\{X_1'\}$ and $\NNm(X_3',\D')=\{X_2'\}$, we have $\tD=\{(X'_3,Y_3')\}$. This implies $Y'_{(1)}(x,\tD)\ne Y'_{(1)}(x,\D')$ for all $x\in (-3,-1)$.

The last example also shows that in general, the ``neighbouring relation" is not necessarily symmetric, nor even with probability one. That is, there are cases where the event
\[
\{
X'_i\in\NNm(X'_j,\D')
\quad\text{and}\quad
X'_j\notin\NNm(X'_i,\D')
\}
,
\]
for some $i,j\in[n]$, occurs with a positive probability.

%\todo{Omer: Need to fix the alignment of the table in the last page.}

%\part*{Additional experimental results}

%\section{Runtime on notMNIST}
%Figure \ref{fig:runtime} shows the construction and evaluation runtime for the prototype rules studied in the Experimental study section.
%Section \ref{sec:simulations}.

% crnote OK: I removed the `[h]` from the table
\begin{table}[h]
%\begin{adjustwidth}{-2.15cm}{}
\centering
\begin{tabular}{l|l |l|l|l}
& Method  & Bayes consistency & Error rate & Compression rate
\\\hline 
%\\ \hline 1 & 1-NN & None &&
%\\ 2 & $k_n$-NN & None &&
%\\
1& SVM & linearly realizable settings & $O(d/n)$& $O(d/n)$
\\\hline 
2& \multicolumn{3}{l|}
{various hierarchical tree-based and compression-based rules}
& unknown
%\\&a stable compression scheme (???)
\\\hline 3& 1-NN &not guaranteed& N/A & $O(1)$ (no compression)
\\\hline 4&$k_n$-NN &universally; in $\R^d$ &minimax rate $O\left(n^{-\frac{\beta(1+\alpha)}{(2\beta+d)}}\right)$$^\dagger$& $O(1)$ (no compression)
\\\hline 5 &\thead[l]{\normalsize{\protonn}\\\normalsize{\ }\\\normalsize{\ }}  & \thead[l]{\footnotesize{universally; in any metric}\\\footnotesize{space admitting a universally}\\\footnotesize{consistent learning rule}}
% line break after metric and after universally
&\thead[l]{\normalsize{unknown}\\\normalsize{\ }\\\normalsize{\ }}&\thead[l]{\normalsize{unknown}\\\normalsize{\ }\\\normalsize{\ }}
\\\hline  6 &\protoknnn &universally; in $\R^d$ & minimax rate $O\left(n^{-\frac{\beta(1+\alpha)}{(2\beta+d)}}\right)$$^\dagger$& near-optimal; $O\left(n^{-\frac{2\beta}{2\beta+d}}\right)$$^\dagger$
%\\\hline  7&\thead[l]{\normalsize{\optinet{}}\\\normalsize{\ }\\\normalsize{\ }}  & \thead[l]{\footnotesize{universally; in any metric}\\\footnotesize{space admitting a universally}\\\footnotesize{consistent learning rule}}
%% line break after metric and after universally
%&\thead[l]{\normalsize{unknown}\\\normalsize{\ }\\\normalsize{\ }}&\thead[l]{\normalsize{unknown}\\\normalsize{\ }\\\normalsize{\ }}
\\\hline 
7&\thead[l]{\normalsize{\optinet{}}\\\normalsize{\ }\\\normalsize{\ }} & \thead[l]{\footnotesize{universally; in any metric}\\\footnotesize{space admitting a universally}\\\footnotesize{consistent learning rule}}
&\thead[l]{\normalsize{minimax rate $O\left(n^{-\frac{\beta(1+\alpha)}{(2\beta+d)}}\right)$$^\dagger$}\\\normalsize{\ }}& \thead[l]{\normalsize{near-optimal; $O\left(n^{-\frac{2\beta}{2\beta+d}}\right)$$^\dagger$}\\\normalsize{\ }}
\\\hline 
8 &\thead[l]{\normalsize{\optinet{}+}\\\normalsize{\compalg}}  & \thead[l]{\normalsize{as \optinet{}}\\\normalsize{\ }} &\thead[l]{\normalsize{as \optinet{}}\\\normalsize{\ }} & \thead[l]{\normalsize{further compression, see}\\\normalsize{Eq.\noindent\ \eqref{eq:deco_thm}$^\ddagger$}}
\\\hline 
9 & \thead[l]{\normalsize{\optinet{} +}\\\normalsize{\compheu}}& \thead[l]{\normalsize{unknown}\\\normalsize{\ }} & \thead[l]{\normalsize{N/A}\\\normalsize{\ }}&\thead[l]{\normalsize{unknown}\\\normalsize{\ }}
\\\multicolumn{5}{l}{}
\\ \multicolumn{5}{l}
{
\footnotesize{\quad $^\dagger$ Under the $\alpha$-Tsybakov margin condition, the $\beta$-H\"older assumption and the \textup{\textbf{SDC}}}}
\\ \multicolumn{5}{l}
{
\footnotesize{\quad $^\ddagger$ Under the \textup{\textbf{SDC}}, $(P_j)_{j=1}^M$ are continuous, and the $\xi$-\textup{\textbf{GMC}}.}}
\end{tabular}
\caption{Error and compression rates}
\label{table:rates}
%\end{adjustwidth}
\end{table}

\begin{algorithm}%[tb]
\caption{\textbf{\compalg{} (\compheu{})}}
\label{alg:protocomp}
\textbf{Input}: A finite labeled set $\D' = (\bX',\bY')$ where the instances in $\bX'$ are distinct
\\
\textbf{Require}: An oracle for $\NNm$ \eqref{def:NNm} ($\widetilde\NNm$ \eqref{eq:comp_heuristic})
\\
\textbf{Output}: A compressed consistent labeled dataset (in \compheu{}, heuristic)
\begin{algorithmic}[1] %[1] enables line numbers
\STATE $\tD \gets \D'$
\FOR{$(X',Y') \in \tD$}
\IF{$\tD$ is a singleton}
\STATE \textbf{break the for-loop}
\ENDIF
\IF{$Y'(Q') = Y'$ for all $Q'\in \NNm(X';\tbX)$ (for all $Q'\in \widetilde\NNm(X';\tbX)$)}
\STATE $\tD \leftarrow \tD \setminus \left\{\left(X',Y'\right)\right\}$
\ENDIF
\ENDFOR
\STATE \textbf{return} $\tD$
\end{algorithmic}
\end{algorithm}

\begin{table}[h]
\centering
\begin{tabular}{| l  l |}
\hline
$\X\times\Y$&$=\R^d\times\{1,\dots,M\}$
%\\$\Y$&$=\{1,\dots,M\}$
\\$\rho$& Euclidean metric
%in $\R^d$
\\
%$S_{x,\gamma}$&$= \{z\in\X : \rho(x,z)<\gamma\}$\\
$L(g)$&$=\PROB\{g(X)\ne Y\}$
\\$\nu$& the distribution over $X\times Y$
\\$P_j(x)$&$=\PROB\{Y=j\mid X=x\}$\\
%, $j=1,\dots ,M$\\
$g^*(x)$&
%$ = \argmax_{j\in\Y}  P_j(x)$ is the 
%\\&
Bayes-optimal classifier\\
$L^*$&$=\PROB\{g^*(X)\ne Y\}$ Bayes
%\\& 
error
% where $g^*$ is the Bayes-optimal classifier
\\
$\Dl$&
%$=\{(X_1,Y_1), \dots, (X_n,Y_n)\}$  
%\\& 
$n$ labeled sample\\
{$\widetilde O$} & $O$-notation with logarithmic factor\\
{$\IND{\cdot}$}& the characteristic function \\
$\Dul$&
%$=\left\{X'_1, \dots, X'_\ul\right\}$ 
$m$ unlabeled samples
% \\& sample
\\
%$P_{n,\ul,j}(x)$& empirical estimators of $P_j(x)$ based on $\Dul$ and $\Dl$
%%,\\& $j\in[M]$
%\\
%$g_{n,m}(x)$&$ = \argmax_{j\in\Y} P_{n,\ul,j}(x)$, a plug-in classification rule
%%\\&%\textcolor{violet}{(Note that is is defined with no usage - so keep it?).}
%\\
{$\tbX$}&an arbitrary subset $\{\tX_1, \dots,\tX_{\tm}\}\subseteq \bX_m'$\\
%\multicolumn{2}{|l|}
%{
%\textcolor{violet}{$\Xinn{i}(x;\tbX)$, $\Xinn{i}(x;\tD)$} 
%\qquad 
%the $i$th nearest neighbor of $x$ in $\tbX$
%}
%\\
{$\Xinn{i}(x;\tbX)$} & the $i$th nearest neighbor of $x$ in $\tbX$\\
%$X_{(1)}(x,\D)$ & $X_{(1)}(x,\bX)$ where $\D=(\bX,\bY)$\\
%$Y_{(1)}(x,\D)$ & label paired to $X_{(1)}(x,\D)$ in  $\D$\\
%$\rho(x,S)$&$=\begin{cases} \argmin\limits_{x'\in S}\rho(x,x'), & \text{if }S\neq\emptyset	,\\ \infty, &\text{if } S=\emptyset , \end{cases}$ $\quad S\subseteq \X$\\
%$S_{x,\gamma}$&$= \{z\in\X : \rho(x,z)<\gamma\}$\\
%$\supp(\mu)$ & $=\{x\in \X: \mu(S_{x,r})>0,\forall r>0\}$ is the support of $\mu$\\
{$V_\ell(\tbX)$}& the Voronoi cell induced by $\tbX$ that corresponds to $\tX_\ell$\\
{$\V(\tbX)$}&$=\{V_1(\tbX),\dots, V_{\tm}(\tbX)\}$, the Voronoi partition induced by $\tbX$\\
%\textcolor{violet}{$P_{n,\ell,j}$} & \textcolor{violet}{$=\frac 1 {|V_l(\tbX)|} \sum_{(X_i,Y_i)\in \Dl }\IND\{Y_i=j,X_i\in V_\ell(\tbX) \}$}\\
{$\tY_\ell$}&$ = \argmax_{j\in\Y} P_{n,\ell,j}$\\
$\tD$ &$=(\tbX,\tbY)=((\tX_1,\tY_1),\dots,(\tX_{\tm},\tY_{\tm}))$ (defined before used)\\ 
$\Xinn{i}(x;\tD)$ & $=\Xinn{i}(x;\tbX)$\\
$\Yinn{i}(x;\tD)$ & the label paired to $\Xinn{i}(x;\tbX)$ in $\tD$ \\
$g_n(x)$&   $=\Yinn{1}(x;\tD)$\\
$\bXg(\gamma)$& a $\gamma$-net $\{\Xg_1(\gamma),\dots, \Xg_{m(\gamma)}(\gamma)\}$ over an arbitrary $\bXg\subseteq \X$
\\$\Ball{x}{\gamma}$ & $=\{x'\in\X:\rho(x,x')<\gamma\}$
\\$\UB{\gamma} $&$=\bigcup_{i = 1}^m  \Ball{X'_i}{\gamma}$
\\ $P_{n,\ell,j}$ &$=\sum_{i=1}^n \IND{Y_i=j,X_i\in V_{\ell}(\gamma) \cap \UB{\gamma}}$
\\$\Yg_\ell(\gamma) $&$= \argmax_{j\in\Y} P_{n,\ell,j}$
\\ $\Dg(\gamma) $&$= (\bXg(\gamma),\bYg(\gamma))=((X_1(\gamma),Y_1(\gamma)),\dots,(X_{\tm(\gamma)}(\gamma),Y_{\tm(\gamma)}(\gamma)))$
\\ $\goptinet_{n,\ul,\gamma}(x)$&$=\Yinn{1}(x; \Dg(\gamma))$
\\$\mu$ (and $f$) & the marginal distribution over $X$ (and its density)
\\\thead[l]{\normalsize{$\alpha$, $\beta$, $\xi$}\\\normalsize{\ }} & \thead[l]{\normalsize{the parameters corresponding to Tsybakov margin condition,}\\\normalsize{to H\"older assumption and to \textup{\textbf{GMC}}, respectively}}
\\ \multicolumn{2}{|l|}{
$P_{(1)}(x)\ge \dots \ge P_{(M)}(x)$ \qquad the ordered values of the conditionals $P_{1}(x), \dots , P_{M}(x)$
}
\\ $\eta(x)$ & $=P_{(1)}-P_{(2)}(x)$
\\ \hline
$\D'$ & $= (\bX',\bY')$, a distinct finite labeled samples
\\ $Y'(X')$ & $X'$'s corresponding label in $\D'$
\\ $V_x(\bX')$ & the cell containing $x$ in the Voronoi partition induced by $\bX'$
\\ \thead[l]{\normalsize{$\NNm(X';\bX')$}\\\normalsize{\ }} & \thead[l]{\normalsize{$=\big\{Q'\in \bX' : \exists x\in V_{X'}(\bX')	
\text{\,\, s.t.\,\, } Q' = \Xinn{2}(x; \bX') \big\}$,}\\\normalsize{the neighbours of $X'$ in $\bX'$}}
\\ $\tD$ & a certain sample that is further compressed from $\D'$
\\ $\lambda$ & the Lebesgue measure on $\R^d$
\\ $\widetilde\NNm(X';\bX',\bX_n) $& $=\big\{Q'\in \bX' : \exists X\in\bX_n \cap V_{X'}(\bX'), Q' = \Xinn{2}(X; \bX') \big\}$
\\ $\delta(x)$ &$= \inf\limits_{x'\in\X:\ \reg(x')=0}\rho(x,x')$
\\ $\boundary{t}$ &$=\{x\in\X:\delta(x)\leq t\}$
\\ \hline
$\barBall{x}{r}$ &$=\{x'\in\X: \rho(x,x')\leq r\}$
\\ $\Seg(x,x')$ & $=\{(1-u)x+ux':0\leq u\leq 1\}$
\\\hline
\end{tabular}
\caption{Notation table}
\label{table:notation}
%\thisfloatpagestyle{}
\end{table}
%\hskip-4.0cm
%\newpage crnote OK: removed newpage 

\end{document}